\documentclass[iicol]{sn-jnl}

\usepackage{graphicx}%
\usepackage{multirow}%
\usepackage{amsmath,amssymb,amsfonts}%
\usepackage{amsthm}%
\usepackage{mathrsfs}%
\usepackage[title]{appendix}%
\usepackage{xcolor}%
\usepackage{textcomp}%
\usepackage{manyfoot}%
\usepackage{booktabs}%
\usepackage{algorithm}%
\usepackage{algorithmicx}%
\usepackage{algpseudocode}%
\usepackage{listings}%

\usepackage{acronym}
\usepackage{adjustbox}
\usepackage{makecell}   
\usepackage[backend=bibtex,style=numeric]{biblatex}
\newcommand{\repeatstring}[2]{ 
  \begingroup
  \count0=0
  \loop
    #1%
    \advance\count0 by 1
    \ifnum\count0<#2
      \repeat
  \endgroup
  }

\newcommand{\sphere}{\ensuremath{\mathbb{S}}}
\newcommand{\real}{\ensuremath{\mathbb{R}}}

\newcommand{\torus}{\ensuremath{\mathbb{T}}}

\newcommand{\ltwo}{\ensuremath{\mathbb{L}^2}}
\newcommand{\project}{\ensuremath{\mathbb{P}}} 
\newcommand{\object}{\alpha} 
\newcommand{\geomReflect}{{\cal O}^{\object}} 
\newcommand{\sampleSize}{n}

\newcommand{\soThreeSet}{\mathcal{S}}
\newcommand{\soThreePoint}{s} 
\newcommand{\imageSpace}{\mathcal{I}}
\newcommand{\imagePoint}{I}
\newcommand{\imageDimension}{D}

\newcommand{\latentManifold}{{\cal M}}
\newcommand{\latentPoint}{z}
\newcommand{\latentDimension}{d}
\newcommand{\latentMap}{\Phi} 
\newcommand{\permutationSet}{\Pi}
\newcommand{\permutation}{\sigma}
\newcommand{\rotationMatrix}{Q}
\newcommand{\lightSet}{\mathcal{L}} 
\newcommand{\lightPoint}{\ell} 
\newcommand{\etal}{{\it et~al.~}}
\newtheorem{defn}{Definition}
\newtheorem{prop}{Proposition}
\acrodef{CAD}[CAD]{Computer-Aided Design}
\acrodef{ML}[ML]{Machine Learning}
\acrodef{MDS}[MDS]{Multidimensional Scaling}
\acrodef{PCA}[PCA]{Principal Component Analysis}
\acrodef{RMSE}[RMSE]{Root Mean Square Error}
\acrodef{LLE}[LLE]{Locally Linear Embedding}

\begin{document}

\title{Investigating Image Manifolds of 3D Objects: Learning, Shape Analysis, and Comparisons}


\author*[1]{\fnm{Benjamin} \sur{Beaudett}}\email{bbeaudett@fsu.edu}
\equalcont{These authors contributed equally to this work.}

\author*[1]{\fnm{Shenyuan} \sur{Liang}}\email{sliang2@fsu.edu}
\equalcont{These authors contributed equally to this work.}

\author[1]{\fnm{Anuj} \sur{Srivastava}}\email{anuj@stat.fsu.edu}

\affil*[1]{\orgdiv{Statistics Department}, \orgname{Florida State University}, \orgaddress{\street{117 N. Woodward Ave}, \city{Tallahassee}, \postcode{32306}, \state{Florida}, \country{USA}}}



\abstract{
Despite high-dimensionality of images, the sets of images of 3D objects have long been hypothesized to form low-dimensional manifolds. What is the nature of such manifolds? How do they differ across objects and object classes? Answering these questions can provide key insights in explaining and advancing success of machine learning algorithms in computer vision. This paper investigates  dual tasks -- learning and analyzing shapes of image manifolds -- by revisiting a classical problem of manifold learning but from a novel geometrical perspective. It uses {\it geometry-preserving} transformations to map the pose image manifolds, sets of images formed by rotating 3D objects, to low-dimensional latent spaces. The pose manifolds of different objects in latent spaces are found to be nonlinear, smooth manifolds. The paper then compares {\it shapes} of these manifolds for different objects using Kendall's shape analysis, modulo rigid motions and global scaling, and clusters objects according to these shape metrics. Interestingly, pose manifolds for objects from the same classes are frequently clustered together.   
The geometries of image manifolds can be exploited to simplify vision and image processing tasks, to predict performances, and to provide insights into learning methods.
}



\maketitle

\section{Introduction}
\label{sec:introduction}

There has been a rapid growth in development of machine learning and AI techniques that carry out vision tasks using image data. Seeking explanation for this growth and advancement, one can attribute this success to the structured nature of image spaces. In the ambient space of all 2D arrays, the sets of images corresponding to real-world scenes are hypothesized to be relatively compact, low-dimensional, and smooth. The consequences of these structures are plenty. The knowledge of image manifolds can be exploited in improving AI tasks that are performed using high-dimensional image data. For example, we can restrict attention to these pertinent manifolds and reach efficient computational strategies for object recognition, despite high-dimensionality of image data. More ambitiously, characterizing these manifolds and their geometries can potentially form a basis of mathematical foundations of understanding learning and AI approaches. 

Image manifolds are loosely defined to be the sets of all images formed by 3D objects under different imaging conditions such as pose, camera distance, and illumination. In the late 1990s and early 2000s, many researchers (see next section) sought to identify and characterize these  image manifolds in order to validate the hypothesis of low-dimensionality. A number of {\it manifold-learning} methods were proposed and investigated but they only attained limited success. These methods were hampered by several factors including low computing power, limited datasets, and lack of differential geometrical tools. Consequently, they focused on relatively simple problems like flattening a Swiss roll. Moreover, these methods seldom considered analyzing geometries of image manifolds in a systematic manner. 

Spurred by recent gains in computing, data storage, data-driven machine learning, and advanced geometrical methods, researchers have resurrected the problem of learning and characterizing image manifolds. In this paper, we investigate manifold learning but we go a step further. In addition to learning manifolds, we systematically investigate geometries of these underlying manifolds using shape analysis methods. Image manifolds for 3D objects are associated with changes in such variables as pose, distance, and illumination. Although the combinatorial dimensions of such a manifold are rather high -- pose or 3D rotation is three-dimensional, a single source illumination can be two-dimensional, and so on -- they are still minor compared to the image dimensions. If we fix other imaging variables and change only the 3D pose of imaged objects, we obtain the {\it pose} image manifolds. Similarly, if we fix the pose and change illuminations, we reach {\it illumination} image manifolds, and so on. \\ 
 
\noindent In this context, we outline the central problems of interest: 
\begin{enumerate}
    \item {\bf Problem P1 --  Image Manifold Learning}: The starting point in this discussion is: How to estimate and characterize the pose, illumination, and other image manifolds associated with 3D objects? While it is popular to express these sets of images as point clouds, we prefer to view them as manifolds in order to capture their differential and global geometries. 
    Our problem differs from traditional manifold learning in that we already know the topology and neighborhood structures, having the goal of finding comprehensible low-dimensional representations of high-dimensional image data. One can phrase our problem as that of embedding graphs into low-dimensional Euclidean spaces while preserving pairwise nodal distances and maintaining neighborhood structures. 
    
     \item {\bf Problem P2 -- Shape Analysis of Image Manifolds}: What are the geometries or {\it shapes} of pose image manifolds? Are these manifolds approximately linear so that they are amenable to linear methods for vision tasks? Are these shapes predictable, interpretable, or consistent with our current understanding? What are the similarities and differences in pose image manifolds across different objects and object classes? Are pose manifolds shaped differently from illumination manifolds? If so, what are the salient differences? 

     \item {\bf Problem P3 -- Statistical Analysis on Image Manifolds}: How can one exploit the shapes of these manifolds to design efficient vision tasks? Can one restrict generative models, such as diffusion flows, to the manifolds to improve sampling, or impose statistical models on these manifolds to facilitate image-based inferences?
\end{enumerate}
In this paper, we focus on the problems {\bf P1} and {\bf P2} with a limited exploration of {\bf P3}. 
We give a brief discussion of {\bf P3} in section \ref{ssec:statAnalysis}.

In order to focus our discussion, we introduce some notation following~\cite{grenander-etal:2000}. Let $\object$ be a 3D object, such as a chair, airplane, or car, and let $\geomReflect$ denote its 3D geometry and reflectance model. 
Let $\soThreePoint \in SO(3)$ represent the 3D pose of $\geomReflect$ relative to the camera, $\lightSet$ be the set of possible illumination conditions, and $\project$ be the perspective projection of $\soThreePoint\geomReflect$ into the focal plane of the camera under the conditions $\lightPoint\in\lightSet$, resulting in a $\imageDimension\times\imageDimension$ image $\imagePoint=\project(\soThreePoint\geomReflect)$.
\begin{defn}
Fixing all imaging conditions other than illumination and rotational pose, the set
$\imageSpace^{\object} = \{ \project(\soThreePoint \geomReflect; \lightPoint)  \vert\, \soThreePoint \in SO(3), \lightPoint\in\lightSet\}$, 
a subset of $\real^{\imageDimension\times\imageDimension}$,
is called a {\it rotation-illumination} or {\it pose-illumination image manifold} of $\object$.
\end{defn}
\begin{defn}
Fixing all imaging conditions other than rotational pose, the set
$\imageSpace^{\object}_{SO(3)} = \{ \project(\soThreePoint \geomReflect) \vert\, \soThreePoint \in SO(3)\}$, 
assuming constant illumination conditions $\lightPoint$, 
is called a {\it rotation} or {\it pose image manifold} of $\object$.
\end{defn}
\begin{defn}
Fixing all imaging conditions other than illumination, the set
$\imageSpace^{\object}_{\lightSet} = \{ \project(\geomReflect; \lightPoint)  \vert\, \lightPoint\in\lightSet\}$,
where the rotational pose remains at a default position,
is called an {\it illumination image manifold} of $\object$.
\end{defn}

Using additional assumptions on the smoothness of $\project$ and non-symmetry of $\geomReflect$ with respect to the rotation group, $\imageSpace^{\object}_{SO(3)}$ shares the closed, boundary-free manifold topology of $SO(3)$. 
(This statement deserves additional consideration to be precise but we leave the details for a future paper.) It is thus three-dimensional and its nonlinear embedding is small (almost singular) in the ambient space $\real^{\imageDimension\times\imageDimension}$. 
While the set of all conceivable lighting conditions has infinitely many dimensions, we restrict variation to the location of a single point light source, and thus limiting the dimensionalities of $\imageSpace^\object$ and $\imageSpace^\object_\lightSet$.
With this notation, we elaborate on the nature of problems {\bf P1} and {\bf P2}.
\\

\noindent {\bf Problem P1 -- Dimensional Reduction and Manifold Learning}:
The biggest challenge faced in {\bf P1}, or manifold learning, is the high dimensionality of image data. Although the image manifolds are expected to be low-dimensional, they are embedded in ultra high-dimensional Euclidean spaces in unknown ways. Thus, a useful step in the pipeline for estimating these manifolds is {\it dimension reduction}, {\it i.e.}, map the high-dimensional ambient observation space to a manageable low-dimensional space and then estimate the embedded manifold in the smaller space. Our initial problem is to analyze $\imageSpace^{\alpha}$ from a set of training images $\{ I^{\alpha}_i \in \real^{\imageDimension\times\imageDimension}\}$ and their given neighborhood structures. 
The dimension reduction seeks a mapping $\Phi$ from the ambient space $\real^{\imageDimension\times\imageDimension}$ to a low-dimensional latent space $\real^\latentDimension$ ($ \latentDimension << \imageDimension^2$) so that $\imageSpace^{\alpha}$ maps to $\latentManifold^{\alpha} = \Phi(\imageSpace^{\alpha}) \subset \real^\latentDimension$, while the neighborhood structures are retained from the larger space. Thus, the problem becomes that of analyzing $\latentManifold^{\alpha}$ from the mapped data points $\latentPoint_i=\{\Phi(I^{\alpha}_i) \in \real^\latentDimension\}$.
If there is a convenient map $\Phi$, then the process  of manifold learning and geometric investigation of learned manifolds will greatly simplify. The concern here is to preserve the geometry of the embedded manifold during this mapping.

Furthermore, how should one formalize the notion of preservation of geometry under $\Phi$?
We must select some characteristics of the high-dimensional data to preserve in the low-dimensional embedding.
Many modern geometry-based dimensionality reduction techniques (see section~\ref{sec:litReview}) focus on local distances or similarities while de-emphasizing relations between more distant points. Others use global geodesic distances constructed using sequences of local values rather than chord lengths.
These are effective for unfolding manifolds and separating groups of points from different classes, which is often useful for low-dimensional visualization. However, this unfolding loses the very shape that we seek to understand.
Thus we will use {\it global isometry} of $\Phi$ to define and quantify its ability to preserve geometry of $\imageSpace^{\alpha}$. 
That is, given two images $I_i, I_j \in \imageSpace^{\alpha}$, we seek $\| \Phi(I_i) - \Phi(I_j)\|$ to be as close to $d_I(I_i, I_j)$ as possible, where $d_I$ is a chosen distance in the image space.
Simple Euclidean distance $d_I(I_i, I_j) = \| I_i - I_j\|$ is well-suited to our goals.

The next question is: How to find an optimal $\Phi$? 
While there are several deep-learning tools available for manifold estimation, including adaptations of GANs, VAEs, etc., we will rely on the older approach of multidimensional scaling (MDS). While MDS has limitations, it provides a straightforward and computationally efficient method of obtaining a distance-preserving $\latentManifold^{\alpha} = \Phi(\imageSpace^{\alpha})$ that is sufficient for our purpose of shape analysis. 
We remark that problem {\bf P3} requires the inverse of $\Phi$, especially to visualize images associated with arbitrary latent points. 
Deep-learning tools such as \cite{liang2024learning} can provide an invertible $\Phi$ that allows visualizing the results of analysis on $\latentManifold^{\alpha}$ as images. In this paper, we are restricting ourselves to shape analysis of $\latentManifold^{\alpha}$ and thus do not require an invertible $\Phi$. 
\\

\noindent {\bf Problem P2 -- Shape Analysis of Image Manifolds}: Shape analysis of one- and two-dimensional manifolds in Euclidean ambient spaces ({\it i.e.}, curves and surfaces) has been studied extensively, and there are several efficient tools available (see the literature review). The two main approaches for shape analysis are:
(1) Kendall's landmark-based approach where objects are represented by a fixed number of registered landmarks with a rigid structure, and 
(2) Elastic shape analysis where objects are represented by curves or surfaces along which sampled points can move during registration.
Since our manifolds are indexed by rigidly structured pose (or illumination) variables, 
we use the simpler Kendall's shape analysis for now.
\\

\noindent
The main contributions of this paper are: 
\begin{enumerate}
    
\item {\bf Latent Space Manifold Learning}: The paper utilizes a classical technique (MDS) to systematically learn pose and illumination manifolds of 3D objects from different classes in low-dimensional latent spaces. This manifold learning, using geometry-preserving dimension-reduction, is different from past learning methods that focused on unfolding low-dimensional manifolds (such as a Swiss roll). This paper uses simulated data to demonstrate superiority of global isometry in learning manifold shapes. It also studies the effects of object symmetries on the shapes of pose manifolds. Finally, it firmly establishes the nonlinearity of these manifolds that has consequences in downstream processing tasks. 

\item {\bf Kendall's Shape Analysis of Latent Space Manifolds}: The main contribution of this paper lies in characterizations of shapes of the image manifolds using formal techniques. The use of Kendall's shape analysis is motivated by the need for invariance to certain transformations that arises during mapping to low-dimensional latent spaces. 

\item {\bf Shape-Based Manifold Clustering}: Using tools from Kendall's shape analysis, we compare and cluster shapes of pose manifolds of multiple objects. The agreement of this clustering with the original object classes provides an interesting and important result. The shape tools can also be used to {\it deform} pose manifolds into one another and develop techniques for learning future manifolds in an efficient manner.  

\end{enumerate}

\section{Literature Review}
\label{sec:litReview}

There is an extensive literature on dimension reduction, manifold learning, and shape analysis. Here we summarize research most pertinent to our goals and point out strengths and weaknesses.  
\\

\noindent {\bf Dimensional Reduction and Manifold Learning Techniques}:
Traditional dimensionality reduction techniques build the foundation for modern representation learning. \ac{PCA}~\cite{pearson1901liii,wold1987principal}, while linear, established core principles by identifying orthogonal directions of maximum variance in data.
Beginning with \cite{torgerson1952multidimensional}, a variety of linear and nonlinear methods termed \ac{MDS} have worked by preserving global pairwise relations between points in the high-dimensional set \cite{mead1992review}. The ability to maintain global $\ltwo$ distances makes \ac{MDS} particularly useful for applications requiring preservation of the overall geometry of the data.
Later methods shift focus toward preserving local structures given the curse of dimensionality affecting Euclidean distances. For instance, \ac{LLE}~\cite{roweis-saul:2000} reconstructs each data point from its neighbors, preserving local geometry in the lower-dimensional space. 
Isomap~\cite{balasubramanian2002isomap} constructs a graph using local distances and applies MDS to the resulting geodesics.
Laplacian Eigenmaps~\cite{belkin2003laplacian} construct a graph representation of the data and preserve local neighborhoods through the Graph Laplacian. 
More recently, the t-SNE~\cite{van2008visualizing} method, modeling similarities between pairs of points as conditional probabilities, has gained widespread adoption due to its excelling results for visualization tasks.
Although these local preservation methods are successful in capturing fine-grained structure, they tend to distort global relations that are essential for characterizing shape. \\

\noindent {\bf Learning Image Manifolds}:
These traditional dimensionality reduction techniques have built the foundations for understanding image manifolds, which have become fundamental to computer vision and image analysis. Early research in image manifold learning observed that collections of images representing objects under smooth variations, such as changes in viewpoint, illumination, or pose, lie in low-dimensional manifolds embedded in a high-dimensional pixel space~\cite{belhumeur1997eigenfaces,tenenbaum2000global,donoho-grimes:2003,lee2005acquiring}. For instance, Belhumeur ~\etal~\cite{belhumeur1997eigenfaces} demonstrated that face images under varying lighting conditions could be represented in a subspace defined by only a few parameters. Similarly, Tenenbaum~\etal~\cite{tenenbaum2000global} and Donoho~\etal~\cite{donoho-grimes:2003} showed how traditional dimensionality reduction techniques reveal intrinsic manifold coordinates from complex image sets. Lee \etal~\cite{lee2005acquiring} further extended these ideas by confirming that transitions between nearby images on such manifolds correspond to small, continuous transformations of the underlying objects, thereby highlighting that image collections are not arbitrary point clouds, but structured entities with geometric coherence. Building on these foundational insights, subsequent studies explored the geometric and topological properties of image manifolds more systematically. Specifically, Peyr\'e~\cite{peyre2009manifold} proposed viewing images as functions defined on manifolds, allowing image processing tasks such as denoising, interpolation, and segmentation to leverage manifold geometry for improved interpretability. Around the same time, Turaga~\etal\cite{turaga2008statistical} analyzed image sequences as trajectories on Grassmann manifolds, providing a statistical framework to capture variability in both appearance and shape. In parallel, the notion of an appearance manifold~\cite{shan2005appearance, einecke2007walking, rahimi2005learning} has also gained attention, highlighting how the visual characteristics of objects (for example, facial expressions or texture changes) evolve in a continuous and smooth manner. For instance, Shan~\etal~\cite{shan2005appearance} investigated how facial appearance transforms under different expressions. Einecke~\etal~\cite{einecke2007walking} examined the change in the appearance of the object over time through appearance-based embeddings. These works bridged the gap between early manifold observations and structured geometric modeling, demonstrating how traditional dimensionality reduction methods can guide the construction of meaningful manifold representations for complex visual data. However, these methods typically view the embedding space as unordered point clouds, and thus do not offer shape comparisons.
\\

\noindent {\bf Advanced Learning Techniques and Latent Space Geometry}: 
In recent years, deep neural networks (DNNs) have provided powerful tools to learn the image manifold, albeit implicitly. A significant breakthrough came with Kingma and Welling's~\cite{kingma2013auto} introduction of the variational autoencoder (VAE), which maps input data to a low-dimensional latent space (encoder) and reconstructs them back to the image space (decoder). Building on these advances in generative modeling, Goodfellow~\etal~\cite{goodfellow2020generative} proposed the basic framework and training procedure for generative adversarial networks (GANs). 
Understanding the geometric properties of these learned latent spaces has emerged as a crucial research direction. Bengio~\etal~\cite{bengio-review:2013} emphasized the fundamental importance of understanding the geometric properties of latent space representations. Subsequent studies~\cite{shao-etal-arxivL2017,kuhnel:2021,shukla-etal:2018} investigating this geometry revealed an unexpected characteristic: latent spaces tend to be surprisingly flat. However, these investigations focused mainly on analyzing existing architectures geared towards image synthesis rather than pursuing designs for learning geometries.
More recently, VAEs, GANs, and other DNNs have been extended to learn geometry-preserving representations~\cite{chen2020learning, lee2022regularized, nazari2023geometric, limgraph, liang2024learning, pai2019dimal, duque2022geometry, gong6recovering}. For instance, Lee~\etal~\cite{lee2022regularized} formulated a regularized autoencoder to preserve relative distances and angles.  Singh~\etal~\cite{singh2021structure} and Lim~\etal~\cite{limgraph} focused on preserving local geometry with autoencoders by building graphs from neighbors. Liang~\etal~\cite{liang2024learning} introduced a geometry-preserving StyleGAN by penalizing changes in global pairwise distances and local curvatures (angles between tangent spaces). Gong~\etal~\cite{gong6recovering} proposed manifold representation meta-learning based on autoencoders to recover the underlying manifold structures. However, these methods focused mainly on developing algorithms to preserve geometric properties in DNNs without explicitly comparing the shapes of the underlying image manifolds. This leaves an important gap in understanding how the overall manifold structure changes through different datasets.
\\

\noindent {\bf Shape Analysis of Manifolds}:
Analysis of manifold shapes~\cite{mardia-dryden-book,small-shapes,kendall-barden-carne,FDA} has become a crucial aspect in understanding these data structures. The field has developed sophisticated methods for analyzing manifold geometry, from shape metrics and deformation visualization to statistical modeling and hypothesis testing. Kendall et al.~\cite{kendall1977diffusion} established the mathematical foundations by defining shape as a geometric property invariant to position, orientation, and scale. However, it assumed dense point correspondences across objects. Modern shape analysis has incorporated registration as an integral component through elastic Riemannian metrics. These methods have evolved from analyzing scalar functions and Euclidean curves~\cite{FDA,younes-distance,younes-michor-mumford-shah:08} to more complex geometries such as 2D surfaces~\cite{su2020shape,jermyn2017elastic}, and 2D/3D shape graphs~\cite{guo2022statistical, bal2024statistical,liang2024shape,beaudett2024reducing}. Additionally, Khrulkov~\etal~\cite{khrulkov2018geometry} proposed a geometry score that compares the topological features of real versus generated data in deep generative models, which is conceptually related to shape-based metrics. These advances in shape analysis complement the manifold learning approaches.
While the analysis of shapes for curves, surfaces, and graphs is well-established, the comparison of higher-dimensional manifold shapes remains a significant challenge. The fundamental difficulty lies in both the mathematical representation and computational tractability; as dimensionality increases, traditional shape metrics become increasingly difficult to define and compute, and the visualization and interpretation of deformations between high-dimensional manifolds remains an open problem. 


\section{Proposed Framework: Manifold Learning and Shape Analysis}
\label{sec:framework}


\begin{figure*}[h]
    \centering
  \includegraphics[width=0.35\textwidth, scale=1]{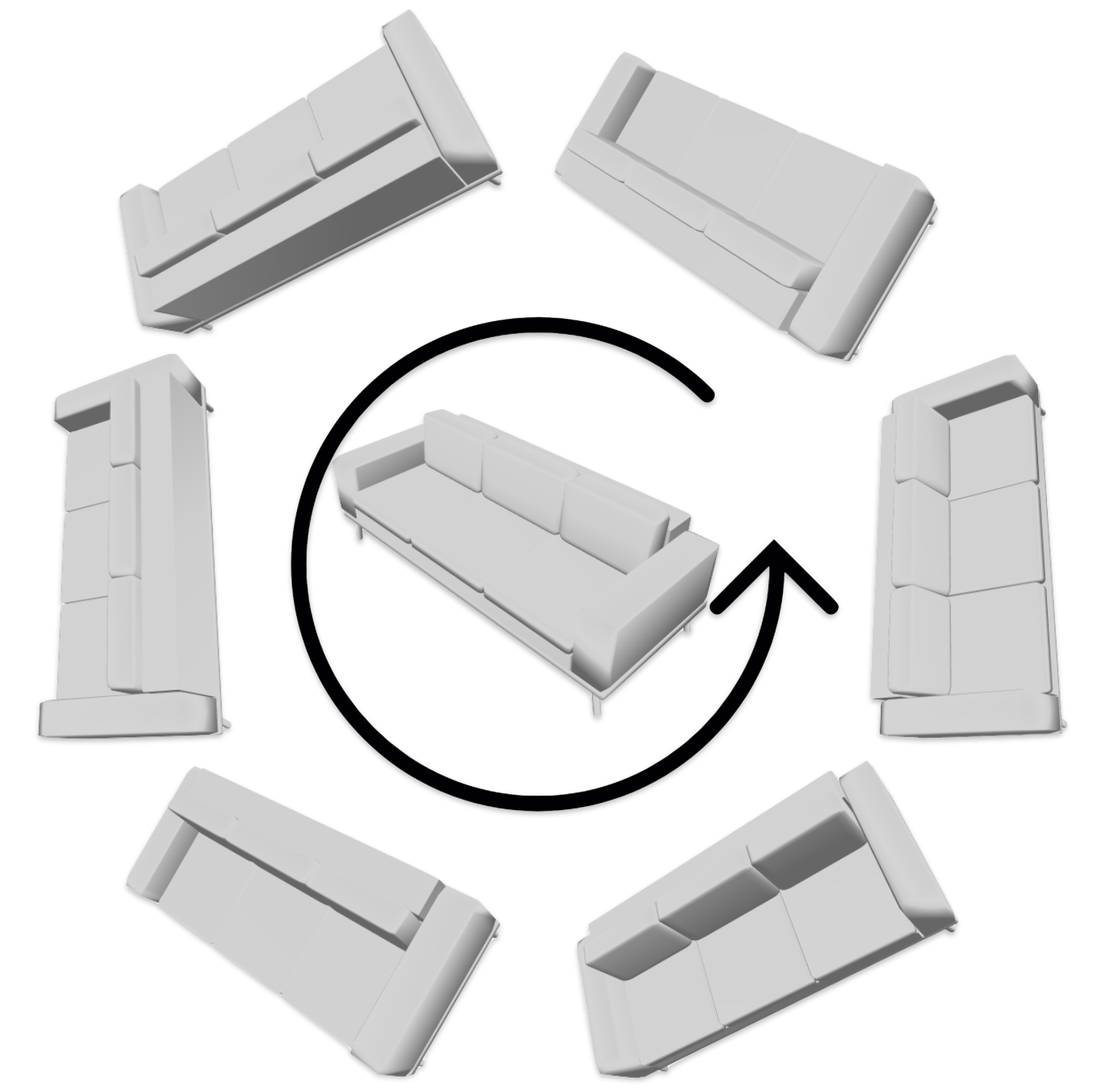}
    ~~~~\includegraphics[draft=false, width=0.5\linewidth]{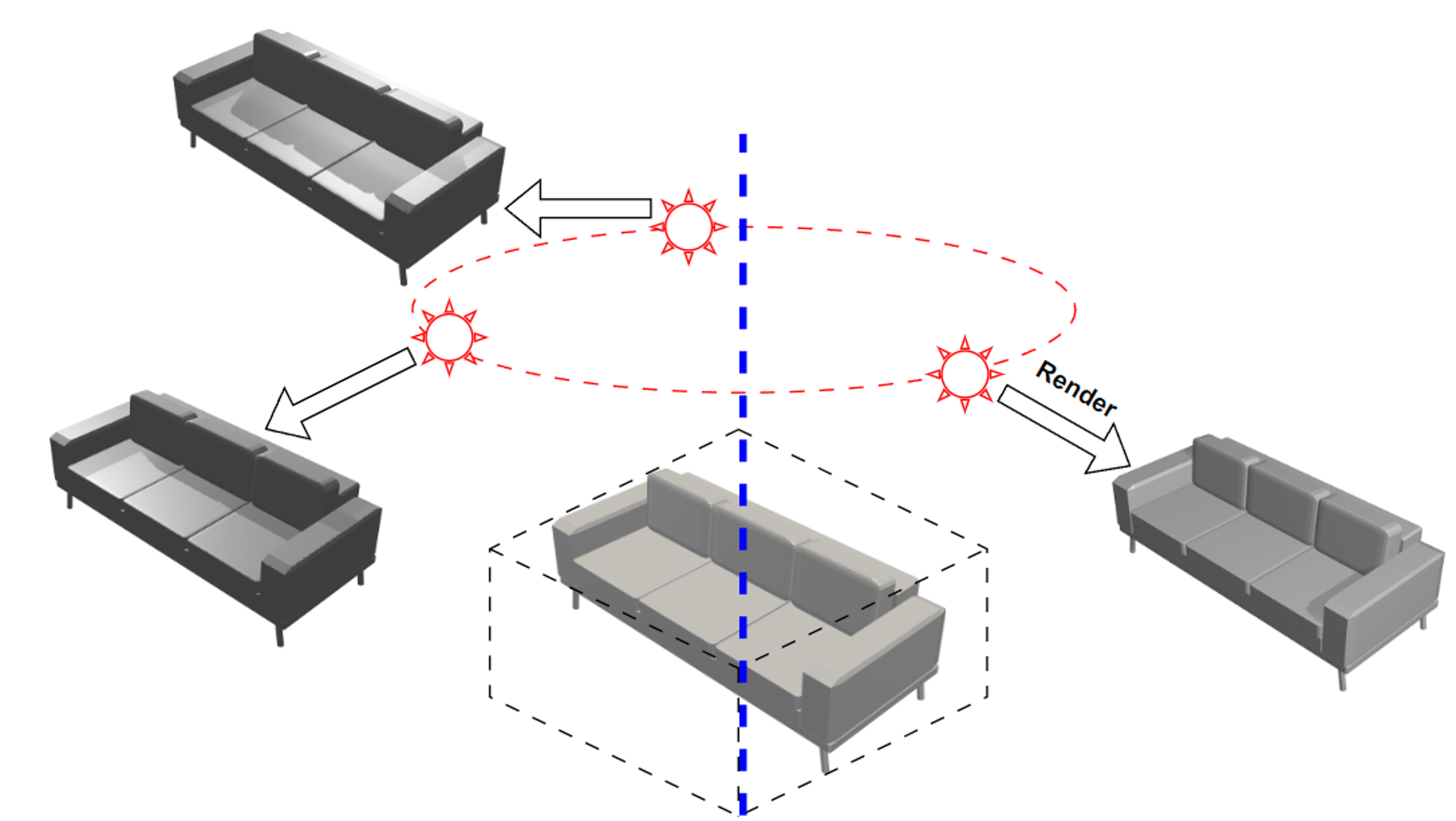}
    \caption{
      \emph{Left:} Pose sampling process. Rotations are applied at regular intervals.
        \emph{Right:} Illumination sampling process. The light source moves along a circular path at constant height.
    }
    \label{fig:samplingProcesses}
\end{figure*}


In this section we establish the notion of an image manifold, and more specifically a pose image manifold. We describe how we generate representations of these manifolds in the image space, how we embed them into lower dimensions to create latent space manifolds, and how we use these latent space representations to study shapes of embedded manifolds.

\subsection{Experimental Setup and Image Data}
\label{ssec:data}

To create an image set for a three-dimensional object $\object$, we begin with its geometry and reflectance model $\geomReflect$ oriented at a default position. 
We use \ac{CAD} objects downloaded from clara.io~\cite{ClaraIO} and processed using meshio~\cite{meshio}.
Fig.~\ref{fig:objectImageExamples} in section~\ref{sec:poseManifolds} shows the CAD objects at their default poses.

Given a finite set of rotations $\soThreeSet\subset SO(3)$, we apply each element of $\soThreeSet$ to an object $\alpha$ and project into the focal plane of the camera to form the rotational pose image set 
$\imageSpace^{\object}_{\soThreeSet} = \{\project(\soThreePoint\geomReflect) \vert \soThreePoint\in\soThreeSet\}$. 
We analyze pose image manifolds built using structured samplings of three subsets of $SO(3)$ which are progressively complex, having intrinsic dimensions of 1, 2, and 3.
The first is a subset with the topology of the circle $SO(2)$, formed by leaning the object forward $\pi/4$ radians then rotating it about its internal $z-$axis.
We analyze this subset in section~\ref{ssec:so2Data}. 
Fig.~\ref{fig:samplingProcesses} illustrates pose image sampling on the $SO(2)$ set. 
Section~\ref{ssec:t2Data} uses a subset with the topology of the torus $\torus^2$, formed by leaning the object by $\pi/4$ radians in all directions over a circle and rotating it about its internal $z-$axis at each of these positions.
The $SO(2)$ set we use is a subset of this torus.
Finally, section \ref{ssec:completeSo3Data} uses a set of points dispersed in a balanced way throughout all of $SO(3)$.

The rotation sets are formed using parameter grids that induce $SO(3)$-neighbor graphs for the data points. Following these graphs, there is a natural structure on the data with a topology that matches the parameter set. For the one-dimensional $SO(2)$ and two-dimensional $\torus^2$ sets, we create 3D images of the latent spaces that emphasize this structure. The $SO(2)$ graphs form looping curves, and the $\torus^2$ graphs form looping surfaces which we slice into bands to improve visibility.

In addition to the pose image manifolds, we analyze image sets produced by varying the lighting conditions while keeping other variables fixed.
Specifically, we form a finite illumination set $\lightSet$ whose elements $\lightPoint$ are defined by point light sources sampled uniformly at angles $\phi_{\lightPoint} \in [0,2\pi)$ along a circle in a horizontal plane with its center directly above the origin. (We abuse notation and use the same symbol $\lightSet$ to denote this finite set.)
Constant ambient light and shadow mapping are also included to ensure base visibility and realistic rendering.
We apply each $\lightPoint$ to an object $\object$ while keeping its rotational pose fixed at the default position to create an illumination image manifold 
$\imageSpace^{\object}_{\lightSet} = \{\project(\geomReflect; \lightPoint) \vert \lightPoint\in\lightSet\}$.
Fig.~\ref{fig:samplingProcesses} illustrates this illumination image sampling. 
We analyze the illumination manifolds in section~\ref{ssec:changingLightSource}.


\begin{figure}
    \centering
    \begin{tabular}{c}
  \includegraphics[width=0.9\linewidth]{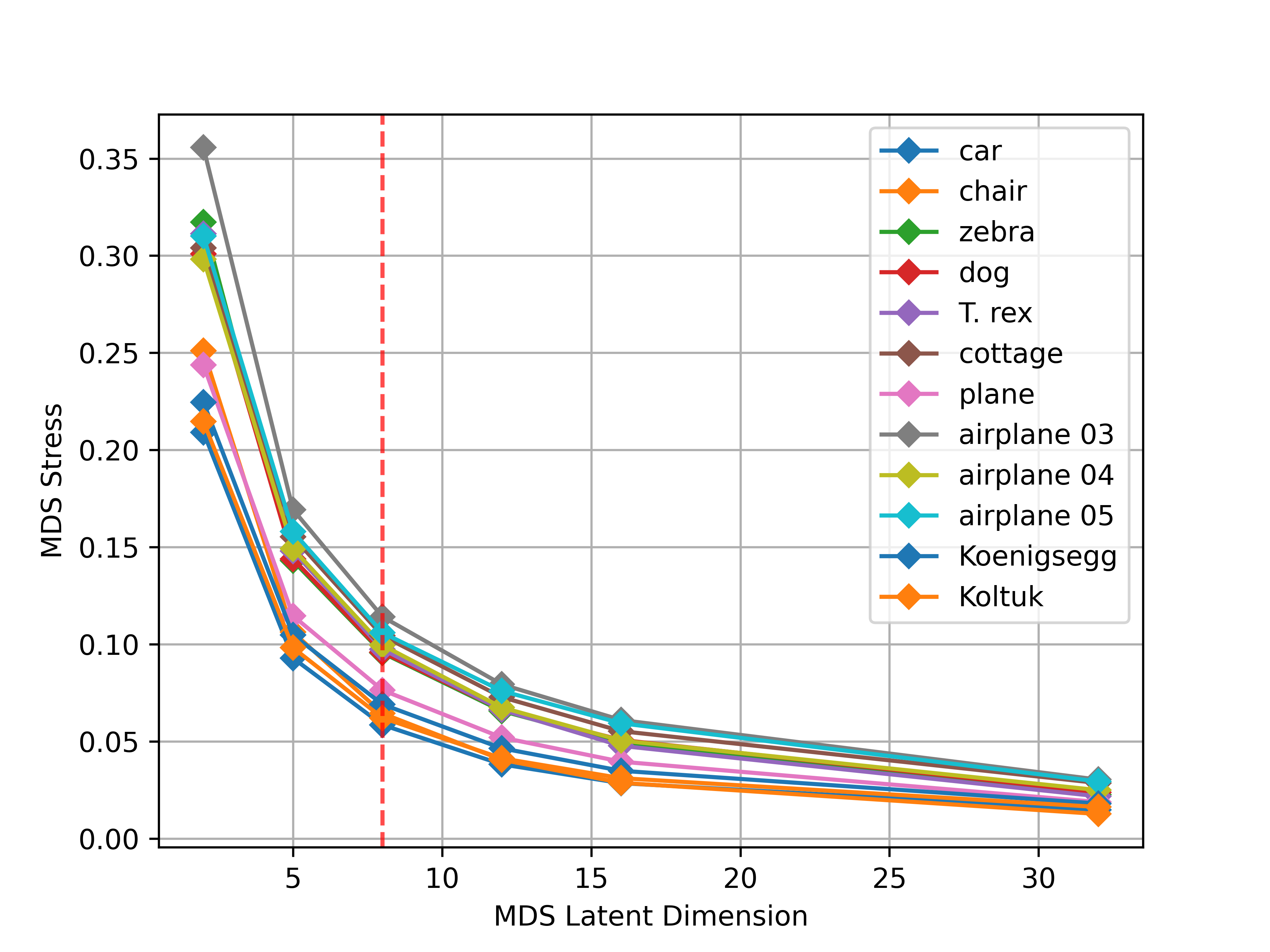}
    \end{tabular}
    \caption{
        \ac{MDS} stress vs. latent dimensions with elbow dimensionality (red dashed line) at $d=8$ on the $SO(2)$ pose set.
        }
    \label{fig:stress-dim}
\end{figure}


\subsection{Geometry-Preserving Map: Multidimensional Scaling (MDS)}
\label{ssec:mdsOverview}

For geometry-preserving image embeddings we use Multidimensional Scaling (MDS), which has the benefits of accuracy in cases with a known ground truth (section \ref{sec:simulationStudies}) and fast computation.
Specifically, we use the form of metric MDS whose objective is to directly preserve global distances from the image space as Euclidean distances in the latent space.
Given a distance matrix $(d_{ij})_{ij=1}^{\sampleSize}$, MDS produces latent space points $\{z_i\}_{i=1}^{\sampleSize}$ with the aim to minimize a loss function called {\it normalized stress} defined as 
$$
  \bigg[\sum_{i<j\le n}(d_{ij}-\Vert\latentPoint_i-\latentPoint_j\Vert)^2 / \sum_{i<j\le n}d_{ij}^2\bigg]^{\frac{1}{2}} \, .
$$
Embedding an image manifold $\imageSpace^{\object}$, we use $d_{ij}=\Vert\imagePoint_i-\imagePoint_j\Vert$ to produce latent points $z_i=\latentMap^{\object}_{MDS}(I_i)$ which form $\latentManifold^\object = \latentMap_{MDS}^\object(\imageSpace^\object)$.
Visually assessing elbow points in dimension-stress curves (Fig.~\ref{fig:stress-dim}), we selected $\real^8$ as the latent space dimension for the $SO(2)$, $\torus^2$, and illumination sets, and selected $\real^{32}$ for the full-$SO(3)$ set.
We use a SMACOF MDS built on the Scikit-learn~\cite{scikitLearn} implementation.

We note in passing that MDS is invariant to rigid transformations in the image space $\real^{D\times D}$. This is useful in creating invariance to certain image transforms that are irrelevant in the current context.

MDS is limited in that $\latentMap_{MDS}^\object$ is not truly a map on $\real^{\imageDimension\times\imageDimension}$, but only creates an embedding of the specific distance matrix it takes as input. More precisely, we denote $\latentMap_{MDS,\soThreeSet}^\object$ to emphasize that the domain is limited to the sample $\imageSpace^\object_\soThreeSet$ produced by a finite set of imaging conditions $\soThreeSet$.
This also means that there is no inverse map for any latent space points outside of $\latentManifold^\object_\soThreeSet$.
For applications where more flexible capabilities are required, a method such as GP-StyleGAN2 \cite{liang2024learning} can be used. GP-StyleGAN2 produces very similar embeddings to MDS but also provides an approximately invertible map that can read out-of-sample points, which allows for more advanced analysis. Everything that follows here can be implemented with GP-StyleGAN2 and the results will be similar.
However, the computational cost is higher than MDS by orders of magnitude.
Access to latent embedding of a determined point set is sufficient for our current study, so we choose MDS for its simplicity and speed.

\subsection{Manifold Comparisons Using Shape Analysis}
\label{ssec:latentSpaceCompare}

Given a 3D model $\geomReflect$ of an object $\object$, a rotational pose $\soThreePoint\in SO(3)$, and illumination conditions $\lightPoint$,
we denote the corresponding image
$\imagePoint = \project(\soThreePoint\geomReflect; \lightPoint)$.
Fixing $\soThreePoint$ and changing $\lightPoint$ results in a change in the illumination pattern on $\geomReflect$.
On the other hand, fixing $\lightPoint$ while varying the poses within some $\soThreeSet\subset SO(3)$ allows us to study the pose image manifold
$\imageSpace^{\object}_{\soThreeSet} = \{ \project(\soThreePoint \geomReflect; \lightPoint) \vert\, \soThreePoint \in \soThreeSet\}$.
Further given a map 
$\latentMap:\real^{\imageDimension\times\imageDimension}\to\real^\latentDimension$, 
we have the latent pose manifold $\latentManifold^\object_\soThreeSet = \latentMap(\imageSpace^\object_\soThreeSet)$, with points $\latentPoint$ parameterized by $\soThreeSet$ such that the neighborhood structure of $\imageSpace^{\object}_{\soThreeSet}$ is retained.
We thus pass our study of the shape of $\object$ to the shape of $\latentManifold^\object_\soThreeSet$, where we ultimately have a representation
$\latentManifold^\object_\soThreeSet : \soThreeSet\to\real^\latentDimension$
with $\soThreeSet$ as the domain and $\real^\latentDimension$ as the range. 
\\

\noindent {\bf Shape Analysis - Preserving Invariances of $\latentMap$}: 
Why are we interested in the {\it shape} of $\latentManifold^{\object}_\soThreeSet$? Shape is a property that is invariant to rigid transformations, global scaling, and re-parameterizations. Rigid transformations -- translations, rotations, and reflections -- form isometries of the range space $\real^\latentDimension$ that are meaningless from our point of view. Scaling and re-parameterizations refer to changes in the imaging process but are not inherent to the 3D object. For all of these, we want to treat them as nuisance variables and remove them from the comparisons.
We will consider them in more detail: 


\begin{figure}
    \centering
    \begin{tabular}{c}
      \includegraphics[width=0.9\linewidth]{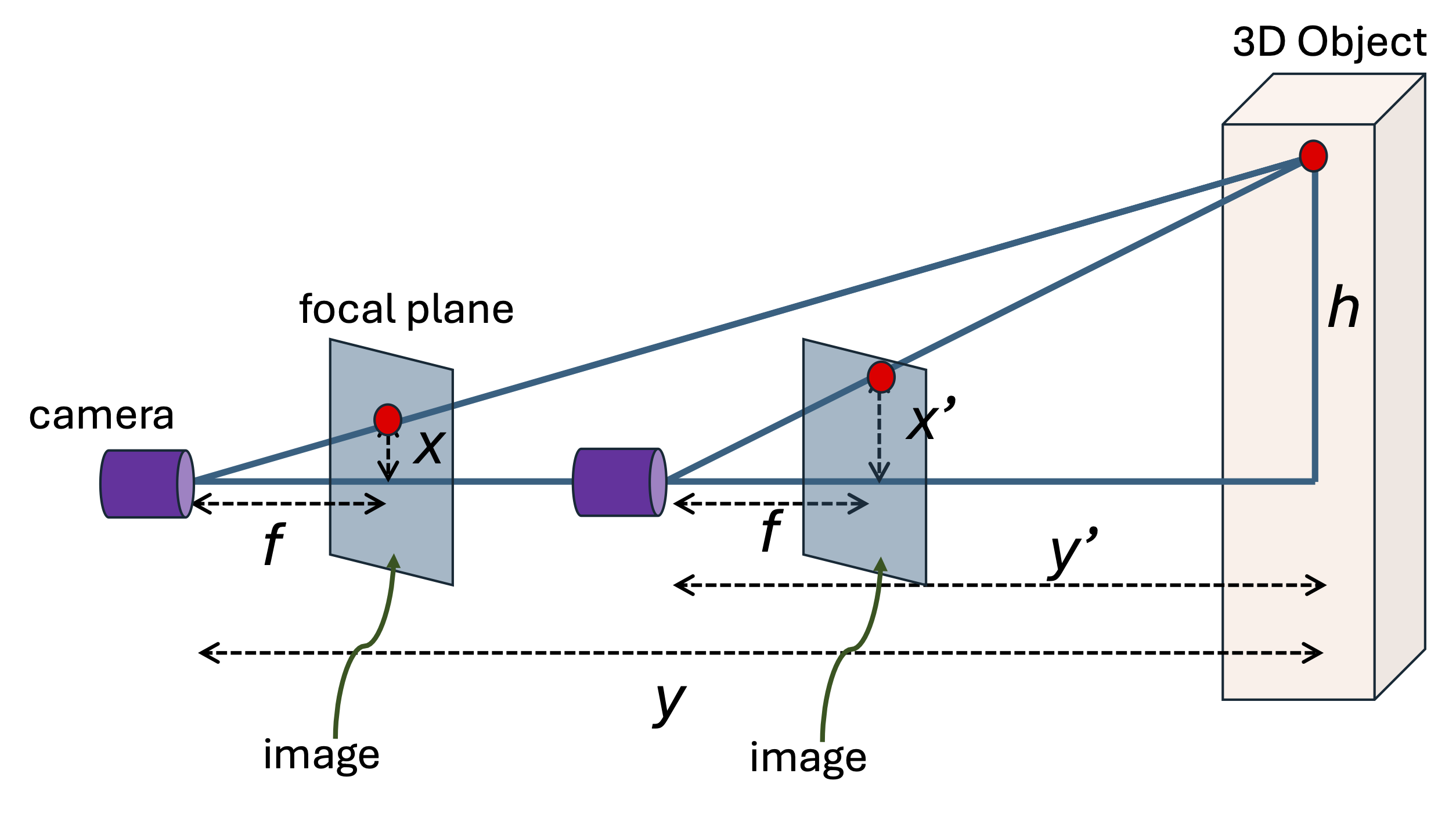}
    \end{tabular}
    \caption{
        Changing the camera-object distance $y$ scales the image pixel height $x$ proportionately.
    }
    \label{fig:camDistProportion}
\end{figure}


\begin{itemize}

  \item {\bf Rigid Transformations of the Range $\real^\latentDimension$}: 
  The manifold $\latentManifold^{\object}_\soThreeSet$ is defined by the mapping $\latentMap: \real^{\imageDimension\times\imageDimension} \to \real^\latentDimension$ that, in turn, is defined by Euclidean distances though the MDS stress terms $(\Vert \imagePoint_i-\imagePoint_j \Vert - \Vert \latentPoint_i-\latentPoint_j \Vert)^2$.
  Since Euclidean distances are preserved by rigid transformations, 
  the composition $\Psi\circ\latentMap$ for any $\Psi:\real^\latentDimension\to\real^\latentDimension$ composed of translations, rotations, and reflections is an equally good solution of the MDS problem as $\latentMap$ itself.
  Thus, they carry no information for analyzing the geometry of $\latentManifold^{\object}_\soThreeSet$. 
  We can remove translation by centering the manifold $\latentManifold^{\object}_\soThreeSet$ so that $\sum_{\latentPoint \in \latentManifold^{\object}_\soThreeSet} \latentPoint = 0$.
  When comparing latent manifolds $\latentManifold^{\object_1}_\soThreeSet$ and $\latentManifold^{\object_2}_\soThreeSet$ from two objects, we can remove rotation and reflection by multiplying with a matrix $\rotationMatrix\in O(\latentDimension)$ determined using the Procrustes method.

  \item {\bf Global Scaling of $\real^\latentDimension$}: The global scale of a centered latent manifold relates to the distance between the camera and $\geomReflect$, and has little relevance to the shape of the object $\object$. 
  We show this by considering images on a continuum rather than on a discrete grid. 

  \begin{prop}
    Using a perspective projection $\project$ and uniform illumination, assume that all objects are far enough from the camera so that all aspects are essentially flat relative to the distance to the camera.
    Then the $\ltwo$ distance between any two (continuous) images $I_1,I_2:\real^2\to\real$, when recomputed for images $I_1',I_2'$ which differ only by altering the camera distance from $y$ to $y'$, scales according to $a = y/y'$.
  \end{prop}
    {\bf Proof}:
    Fig.~\ref{fig:camDistProportion} shows an illustration of this idea. Here, $h$ is the height of an imaged part on the object, $y, y'$ are two camera distances, $f$ is the focal length of the camera, and $x, x'$ are the pixel coordinates (from the image center). From the diagram, we can see that
    $\frac{h}{y} = \frac{x}{f}\ \ \mbox{and}\ \  \frac{h}{y'} = \frac{x'}{f}$.
    Therefore, $x' = \frac{y}{y'} x = a x$ and the pixel coordinate in the image scales inversely to the ratio of camera distances. The new image resulting from a change in camera distance is given by $I'(x') = I(x)$. Comparing the squared $\ltwo$ distances between two images at different distances:
    \begin{multline*}
    \|I_1' - I_2'\|^2 
    = \int_{\real^2} (I_1'(x') - I_2'(x'))^2~dx'
    \\
    = \int_{\real^2} (I_1(x) - I_2(x))^2~a^2~dx 
    = a^2 \|I_1 - I_2\|^2\ .
    ~~~~~~~\ensuremath{\Box}
    \end{multline*}
 
    A similar argument applies to the Euclidean distances between images on a finite, discrete grid, but in an approximate way. 
  Since the camera distance to the object is variable in an arbitrary way, we want the shape analysis of $\latentManifold^{\object}_\soThreeSet$ to be invariant to scaling of the values $\Vert \imagePoint_i-\imagePoint_j \Vert$.
  This global scaling in image distance is accommodated in the MDS problem by simply scaling the latent point norms analogously, so we are free to scale $\latentManifold^{\object}_\soThreeSet$ as needed.


\begin{figure*}
    \centering
    \includegraphics[width=1.0\linewidth]{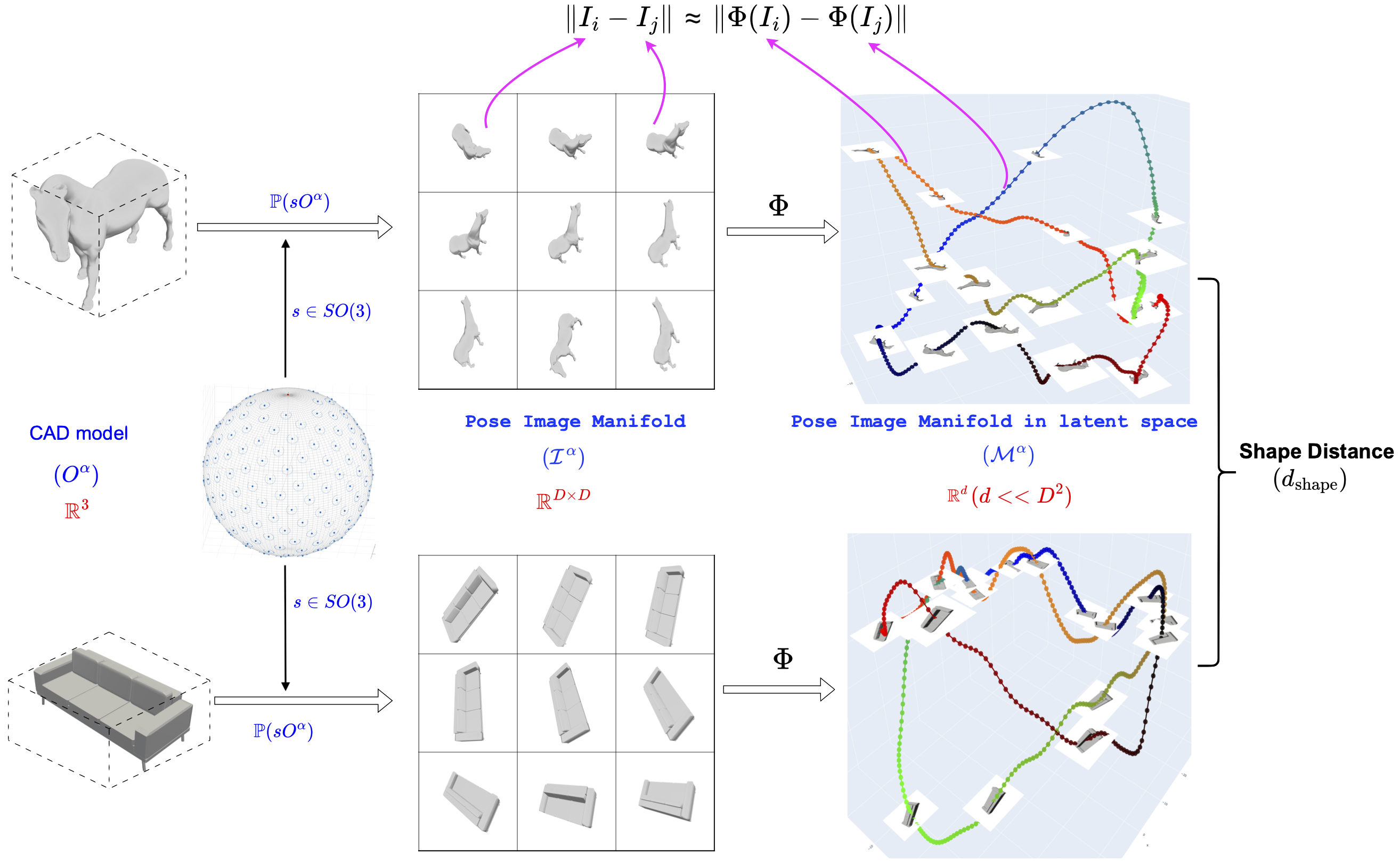} 
    \caption{Pipeline from 3D objects to pose image manifold shape distances.}
    \label{fig:pipeline}
\end{figure*}


  \item {\bf Re-Parameterization of the Domain $S$}:
  If we change the default position of an object by applying a rotation $\soThreePoint_0\in SO(3)$, this results in a `rigid' alteration of the image set: 
  $\widehat{\imageSpace}^\object_\soThreeSet = \{ \project(\soThreePoint\cdot\soThreePoint_0 \geomReflect; \lightPoint) \vert\, \soThreePoint \in \soThreeSet\}$.
  In the case of the entire space $\soThreeSet=SO(3)$, then considered as point sets we have the same pose manifolds $\widehat{\imageSpace}^\object_{SO(3)}=\imageSpace^\object_{SO(3)}$ and $\widehat{\latentManifold}^\object_{SO(3)}=\latentManifold^\object_{SO(3)}$. The difference is simply a re-parameterization 
  $\widehat{\latentManifold}^\object_{SO(3)}(\soThreePoint)=\latentManifold^\object_{SO(3)}(\soThreePoint\cdot\soThreePoint_0)$,
  so that comparisons of latent pose manifolds $\latentManifold^{\object_1}_{SO(3)}$ and $\latentManifold^{\object_2}_{SO(3)}$ of two objects should be invariant to global action of a single $\soThreePoint_0\in SO(3)$.
  We account for this in practice by allowing registration using a restricted set of `rigid' permutations $\permutationSet_\soThreeSet$ determined by the finite set $\soThreeSet$ so that the adjacency graph structure is preserved in the point order.

  Besides global rotation, an equivalent pose image manifold can be produced by changing the order of points in $\soThreeSet$ in a way that preserves neighborhood structure but reflects directions (for example, traversing a circle clockwise versus counterclockwise). The permutation sets $\permutationSet_\soThreeSet$ account for this kind of change as well.
  However, we note that the efficacy of this approach to registration is limited when working with finite subsets of $SO(3)$. Our manually selected default poses for the objects constitute a pre-registration which addresses this. 

\end{itemize}
In summary, given any manifold $\latentManifold^{\object}$, we are interested in studying its shape in a manner that is invariant to rigid transformation, global scaling, and rigid re-parameterization.
For any two objects $\object_1$ and $\object_2$, and images generated from a set of imaging conditions $\soThreeSet$, 
we want to define a latent space shape metric so we can compare the objects using 
$$
d_{\soThreeSet}(\object_1, \object_2) = 
d_{shape}(\latentManifold^{\object_1}_\soThreeSet, \latentManifold^{\object_2}_\soThreeSet).
$$


\noindent {\bf Kendall's Shape Analysis of Pose Manifolds}:
This framework provides metrics that are invariant to location, scale, and rotational orientation in $\real^{\latentDimension}$ of pose manifolds. With a finite $\vert \soThreeSet \vert=\sampleSize$, this shape metric is computed using the following steps: 
(1) First we remove translations in 
$\{\latentPoint_i\}_{i=1}^{\sampleSize}=\latentManifold^\object_\soThreeSet$
by centering them to have mean 
$\frac{1}{\sampleSize}\sum_{i=1}^{\sampleSize}\latentPoint^c_i=0$.
(2) We remove global scales by scaling the points so that they satisfy
$\sum_{i=1}^{\sampleSize}\Vert\tilde{\latentPoint}_i\Vert^2=\sampleSize$.
(3) Rotation, reflection, and re-parameterization are removed by pairwise registration and alignment between points in two manifolds $\latentManifold^{\object_1}_\soThreeSet$ and $\latentManifold^{\object_2}_\soThreeSet$.
Given two standardized latent point sets $\{\tilde{\latentPoint}^{(j)}_i\}_{i=1}^{\sampleSize},~j=1,2$, the registration defines an index permutation function $\permutation(\cdot)$. 
The registered point sets are aligned with an orthogonal linear transformation $\rotationMatrix \in O(\latentDimension)$ determined using the Procrustes method.
The allowed set of permutations $\permutationSet_\soThreeSet$ is confined to rigid index shifts that preserve adjacencies.
Given a primary matching between $\tilde{\latentPoint}^{(1)}_1$ and any $\tilde{\latentPoint}^{(2)}_i$, the rest of the matchings are determined in tandem by choosing one of $k$ directionalities appropriate to $\soThreeSet$.
The number $k$ is related to the intrinsic dimension of $\soThreeSet$, and in our applications it is always small.
Thus there is no computational difficulty in checking all 
$\vert\permutationSet_{\soThreeSet}\vert=k\sampleSize$ permissible registrations along with their optimal $\rotationMatrix$ and selecting the one with the lowest \ac{RMSE}.
This defines the shape distance
\begin{multline}
  d_{\soThreeSet}(\object_1,\object_2) = 
  d_{shape}(\latentManifold^{\object_1}_\soThreeSet, \latentManifold^{\object_2}_\soThreeSet) :=
  \\
  \operatorname*{min}_{\permutation\in\permutationSet_{\soThreeSet},\rotationMatrix\in O(\latentDimension)}
  \bigg[ \frac{1}{\sampleSize}\sum_{i=1}^{\sampleSize}\big\Vert \rotationMatrix\tilde{\latentPoint}^{(1)}_{\permutation(i)}-\tilde{\latentPoint}^{(2)}_i \big\Vert^2 \bigg]^{\frac{1}{2}}.
  \label{eq:shapeDistance}
\end{multline}
The complete process of obtaining shape distances between a pair of 3D objects is illustrated in Fig.~\ref{fig:pipeline}. 

\section{Manifold Learning: Simulation Studies}
\label{sec:simulationStudies}


\begin{figure*}[h]
  \begin{center}
  \begin{tabular}{|c|c|c|c|c|c|}
    \hline
     Ground Truth & MDS & Isomap & Lapl. Eig. & LLE & t-SNE \\
    \hline
    {\adjustbox{margin=1pt, trim={0} {0} {0} {0}, clip}{%
    \includegraphics[width=0.135\textwidth]{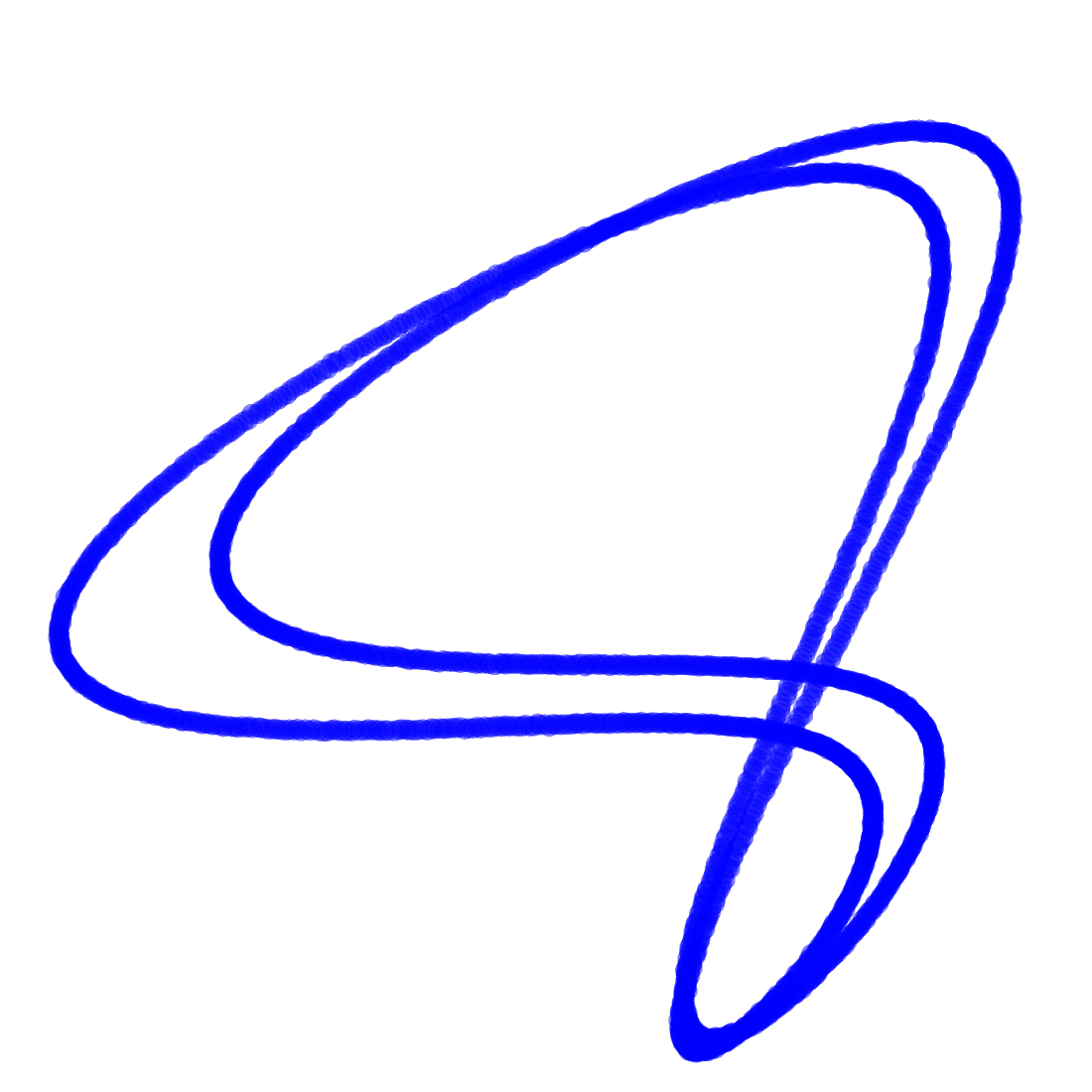}}} &
    {\adjustbox{margin=1pt, trim={0} {0} {0} {0}, clip}{%
    \includegraphics[width=0.135\textwidth]{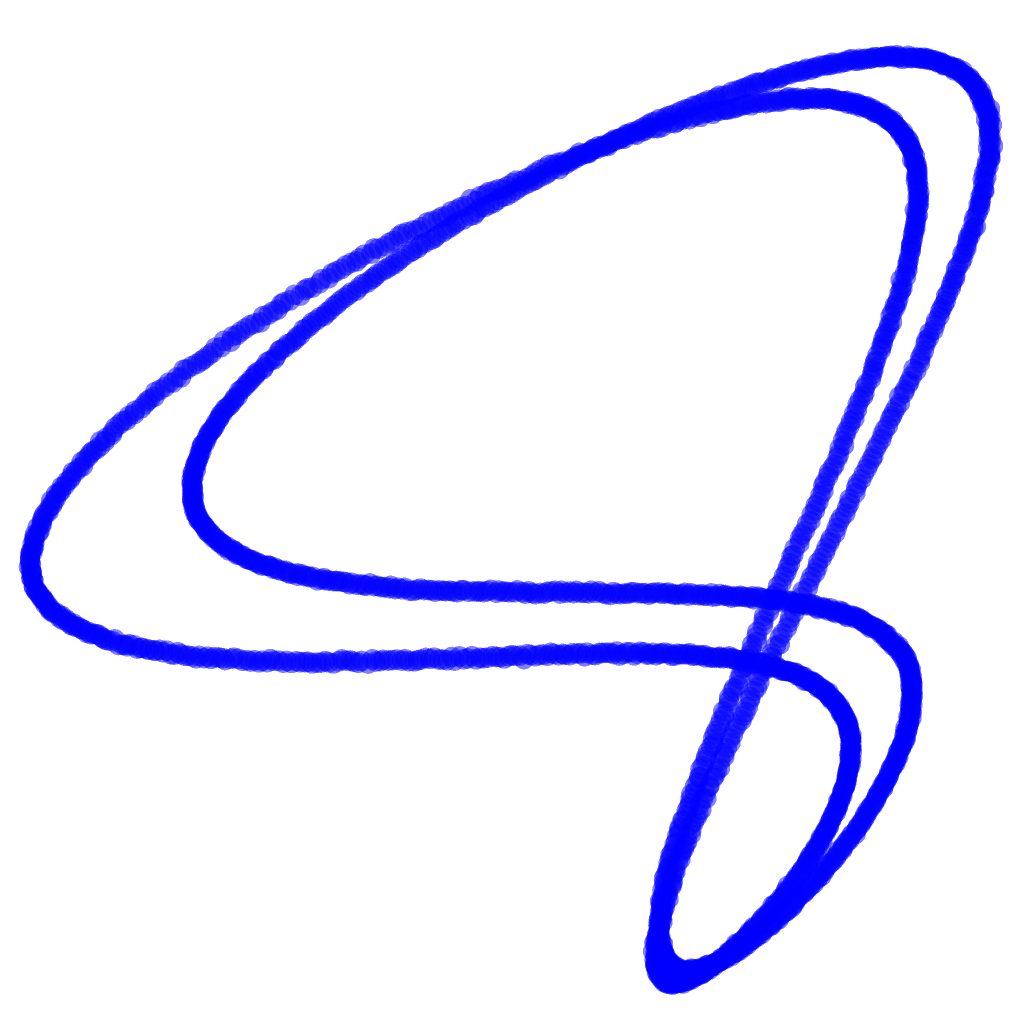}}} &
    {\adjustbox{margin=1pt, trim={0} {0} {0} {0}, clip}{%
    \includegraphics[width=0.135\textwidth]{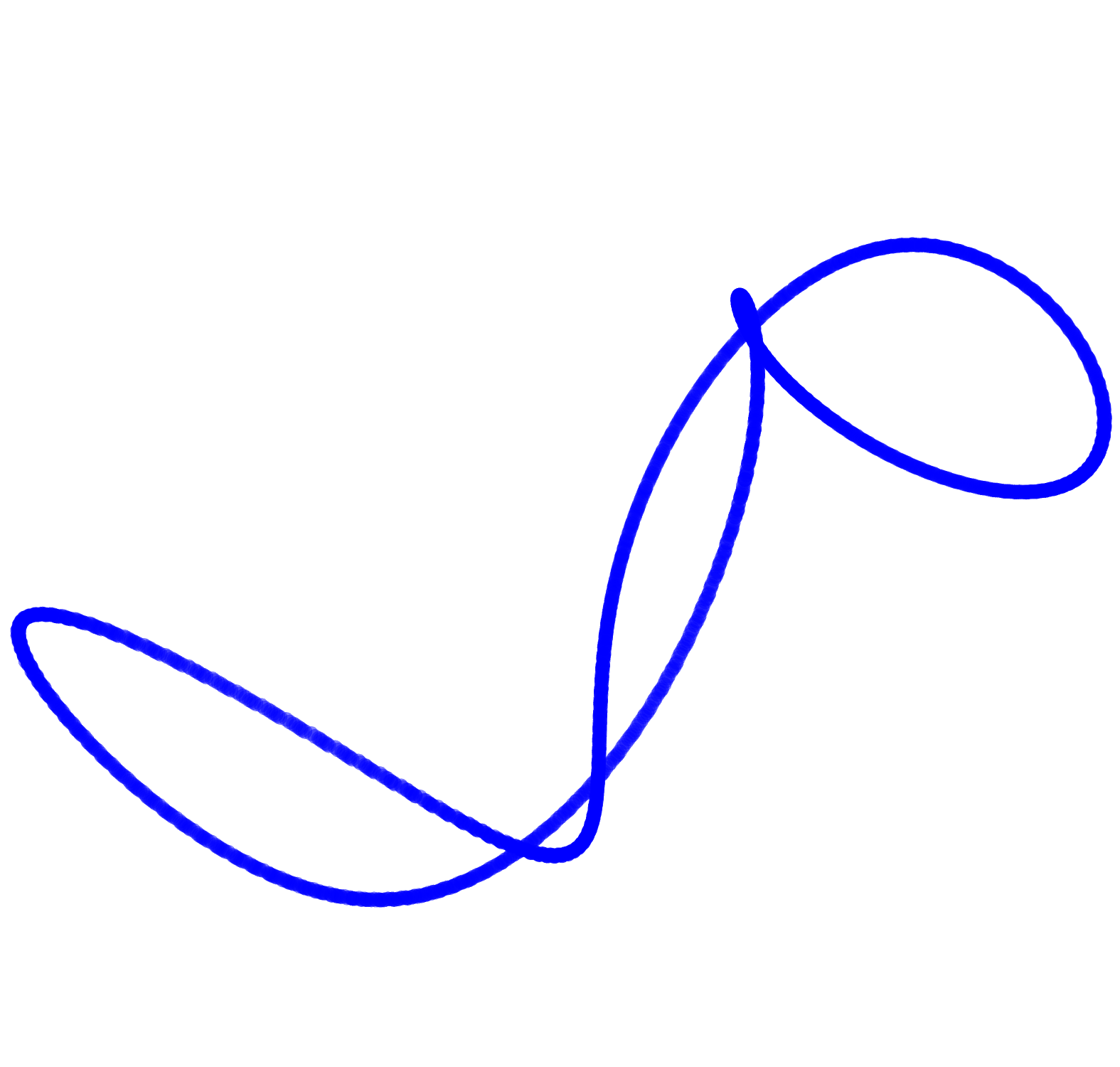}}} &
    {\adjustbox{margin=1pt, trim={0} {0} {0} {0}, clip}{%
    \includegraphics[width=0.135\textwidth]{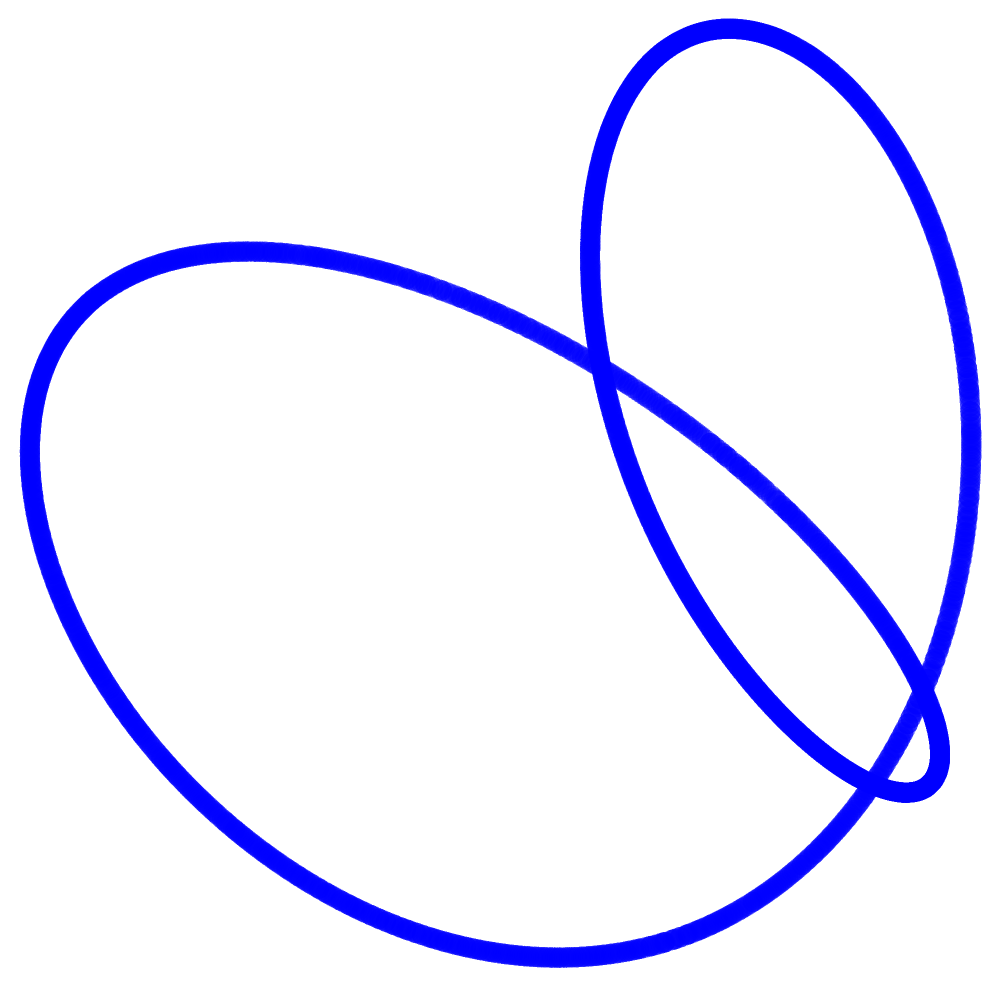}}} &
    {\adjustbox{margin=1pt, trim={0} {0} {0} {0}, clip}{%
    \includegraphics[width=0.135\textwidth]{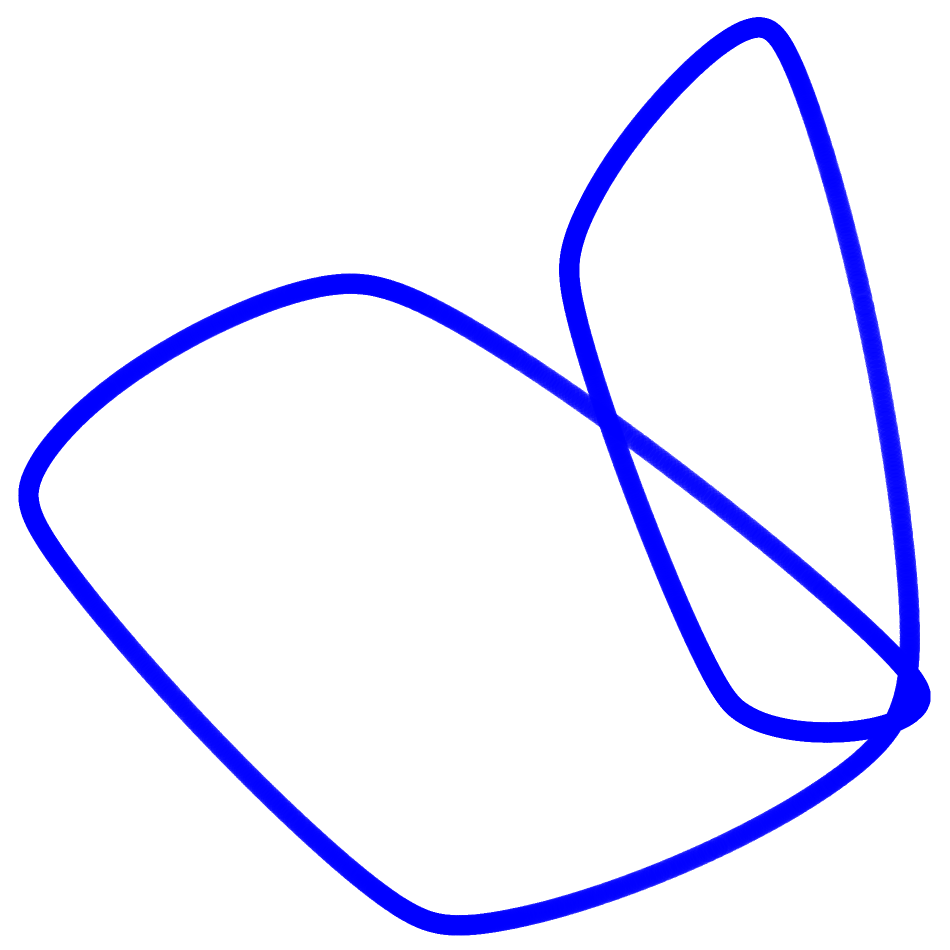}}} &
    {\adjustbox{margin=1pt, trim={0} {0} {0} {0}, clip}{%
    \includegraphics[width=0.135\textwidth]{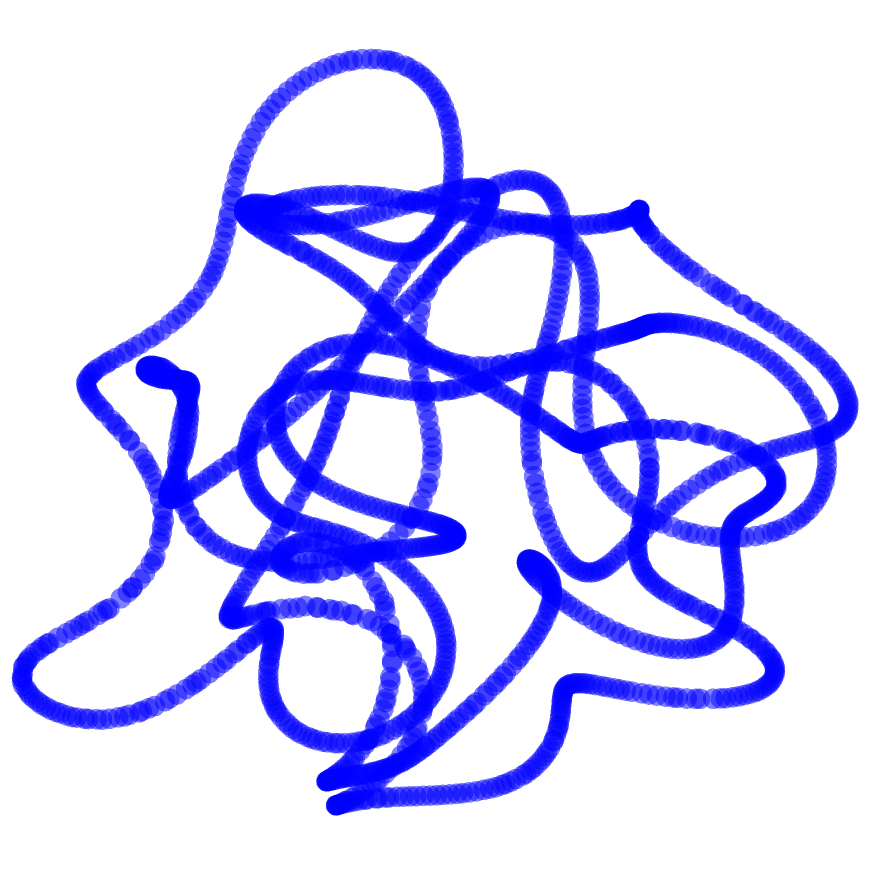}}} \\
    \hline
    {\adjustbox{margin=1pt, trim={0} {0} {0} {0}, clip}{%
    \includegraphics[width=0.135\textwidth]{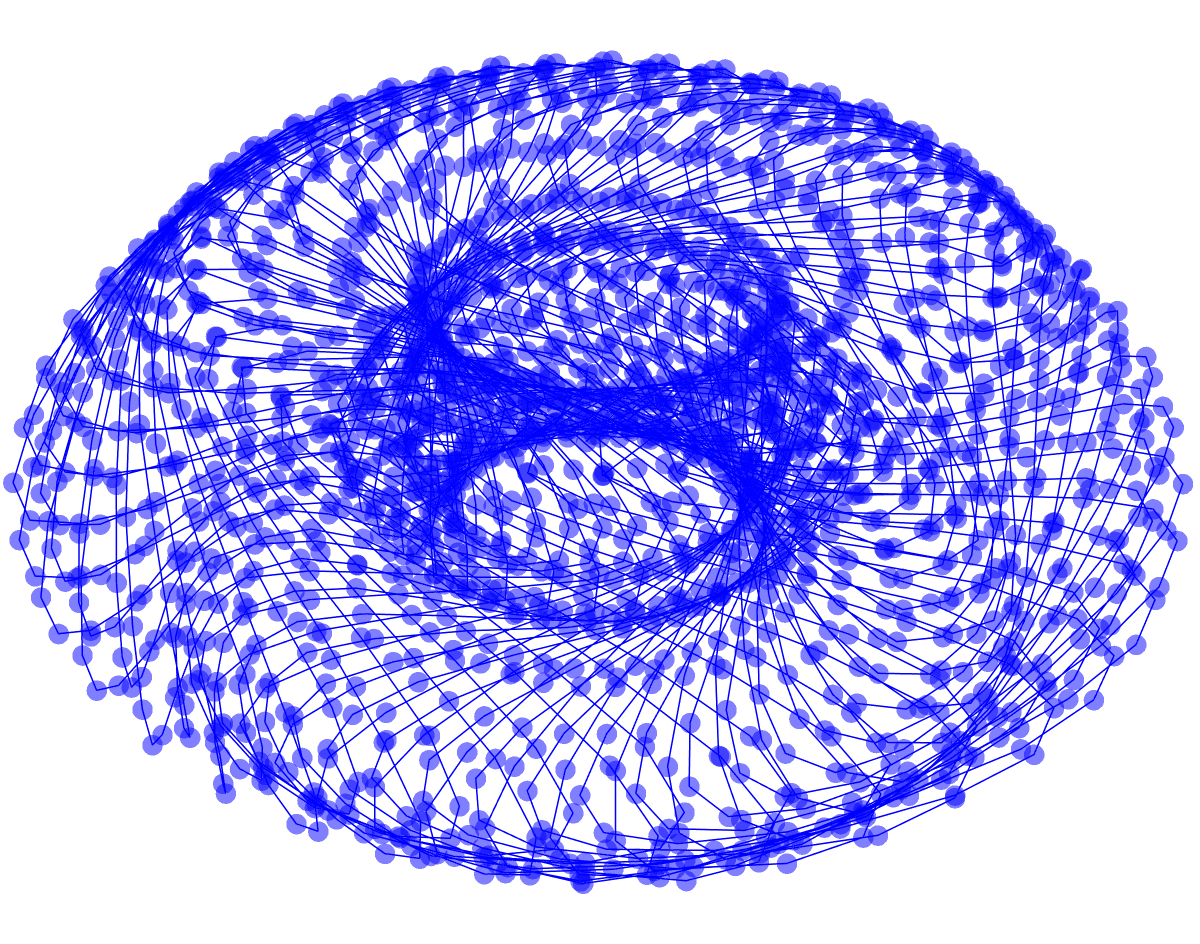}}} &
    {\adjustbox{margin=1pt, trim={0} {0} {0} {0}, clip}{%
    \includegraphics[width=0.135\textwidth]{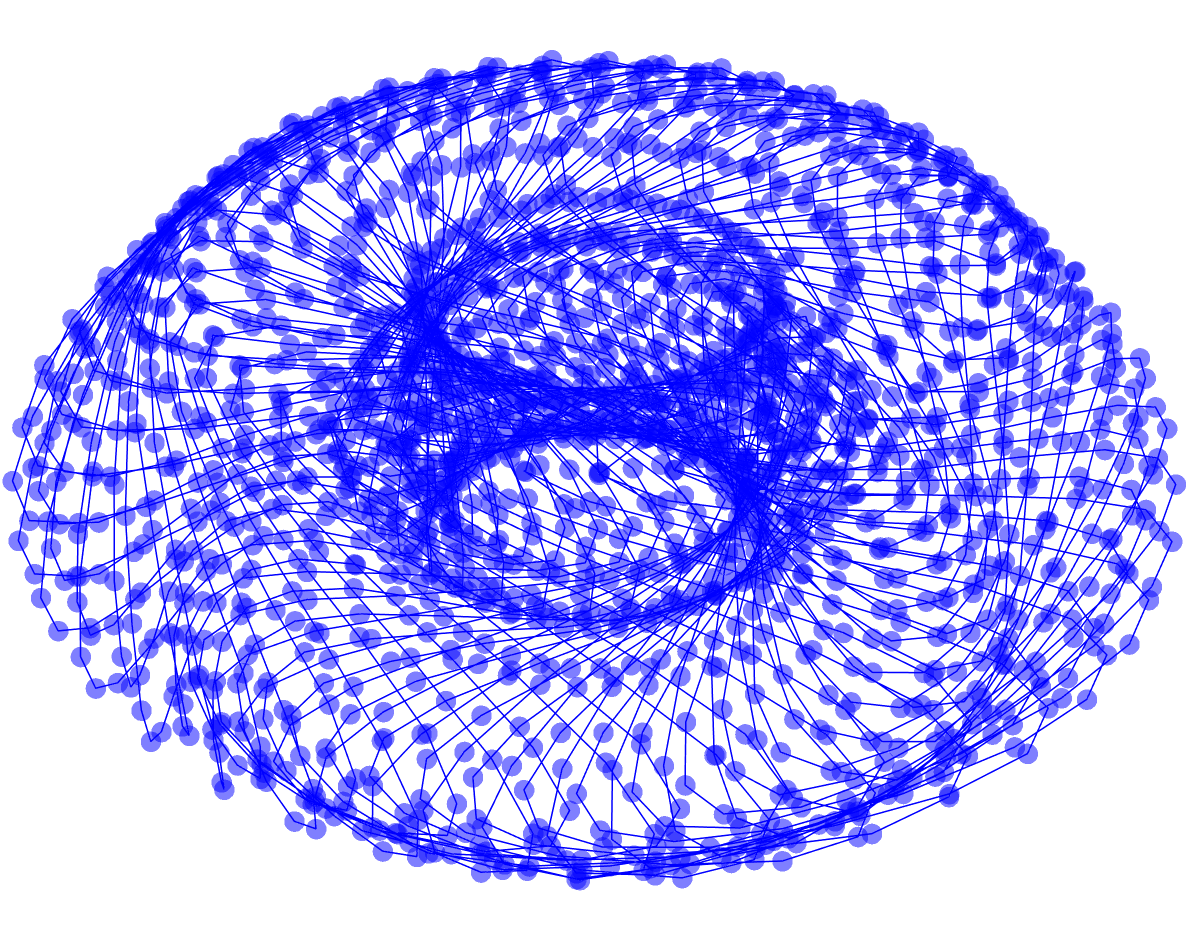}}} &
    {\adjustbox{margin=1pt, trim={0} {0} {0} {0}, clip}{%
    \includegraphics[width=0.135\textwidth]{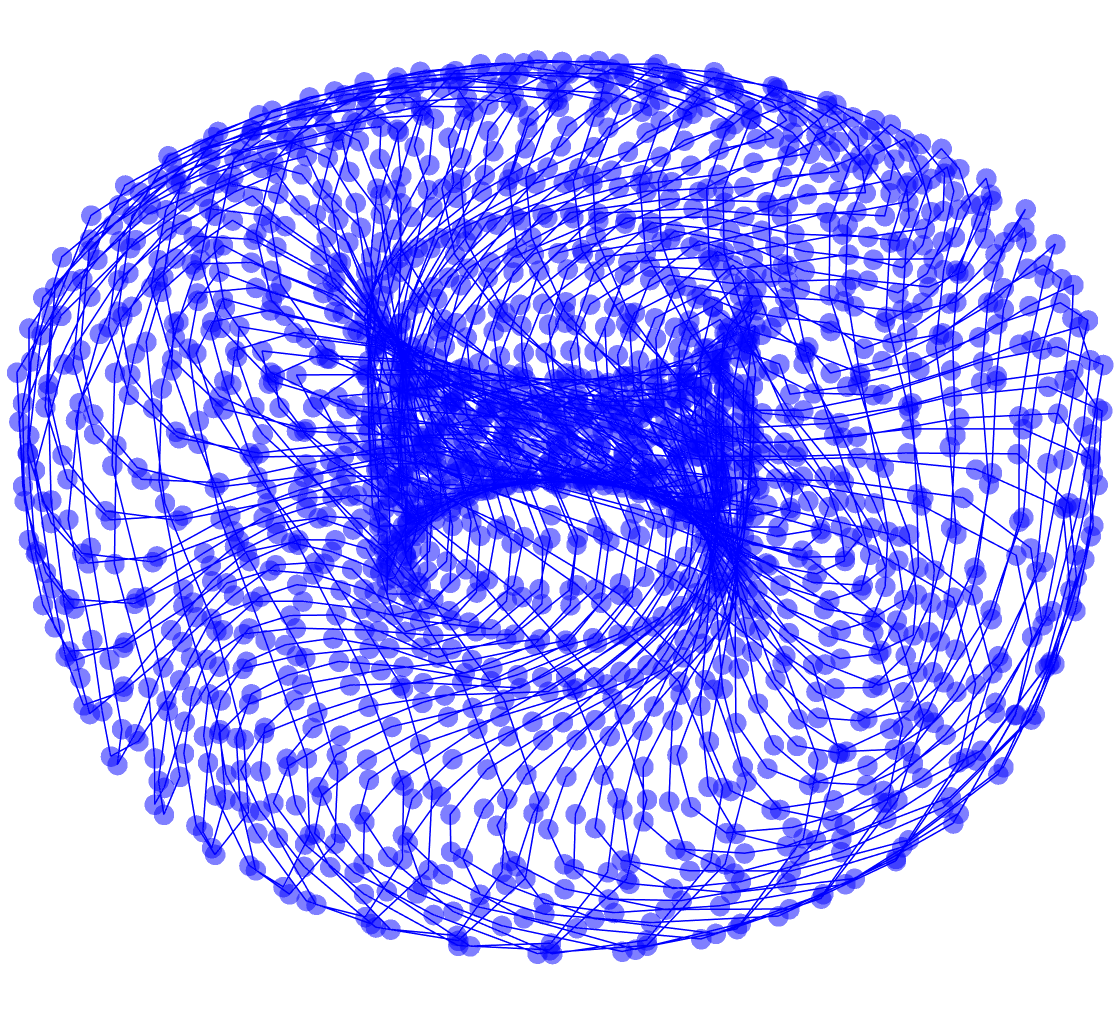}}} &
    {\adjustbox{margin=1pt, trim={0} {0} {0} {0}, clip}{%
    \includegraphics[width=0.135\textwidth]{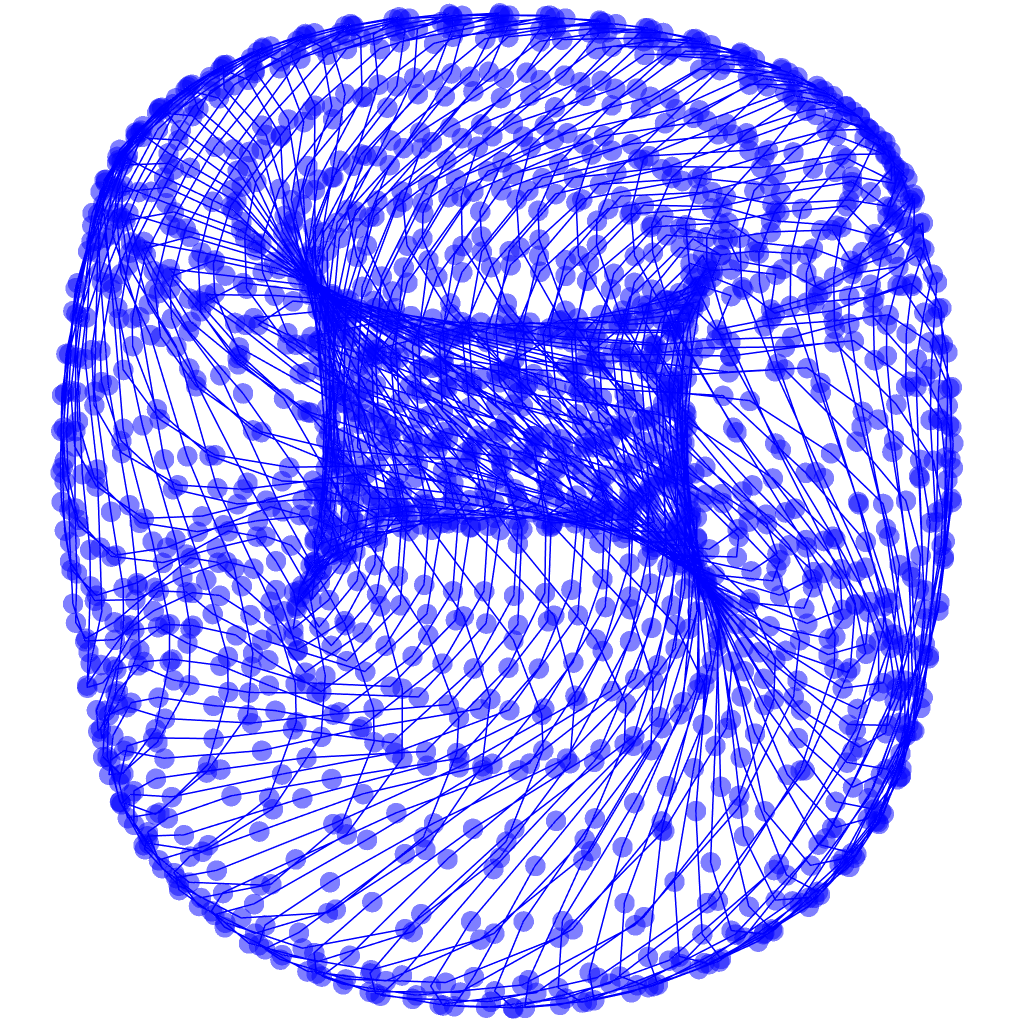}}} &
    {\adjustbox{margin=1pt, trim={0} {0} {0} {0}, clip}{%
    \includegraphics[width=0.135\textwidth]{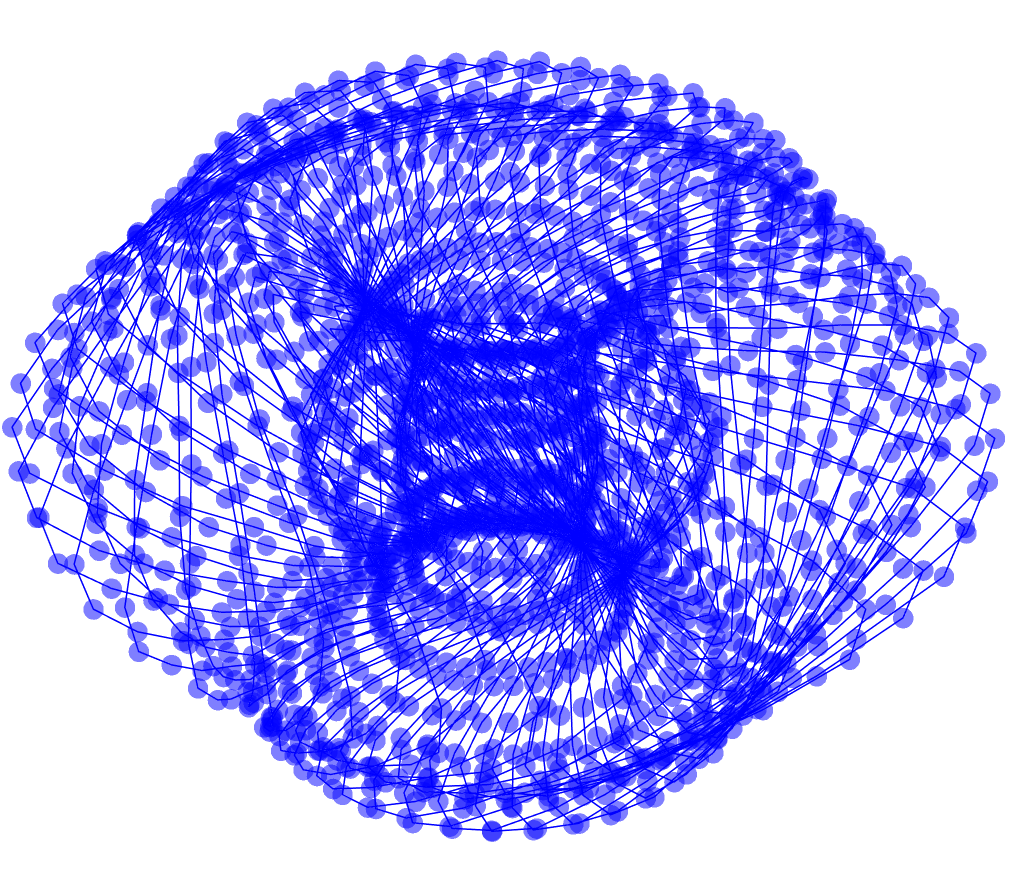}}} &
    {\adjustbox{margin=1pt, trim={0} {0} {0} {0}, clip}{%
    \includegraphics[width=0.135\textwidth]{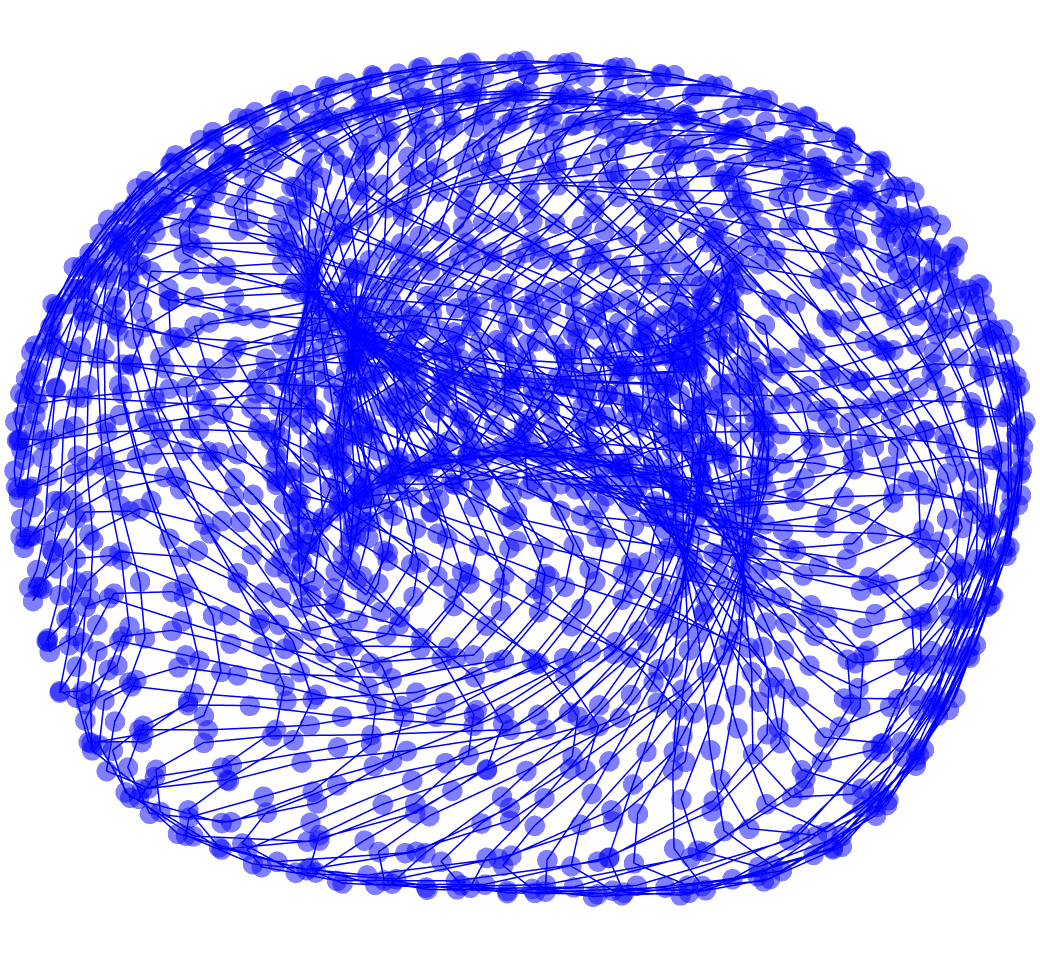}}} \\
    \hline
  \end{tabular}
  \caption{
    First three PCA axes of the latent space learned by MDS and other non-linear dimension reduction methods for the noisy double saddle (top row) and noisy looped double saddles (bottom row) datasets.
    }
  \label{fig:comparison_dsaddle_noisy}
  \end{center}
\end{figure*}



\begin{table*}[h]
  \centering
    \caption{Shape distances between learned latent spaces under different methods and the ground truth manifold. Lower values indicate closer alignment to the ground truth.}
    \label{tab:sim_quant_compare}
  \begin{adjustbox}{max width=.9\textwidth}
    \begin{tabular}{@{}l *{7}{>{\centering\arraybackslash}p{1.7cm}}@{}}
        \toprule
         & 
        \makecell{Isomap} & 
        \makecell{Lapl. Eig.} & 
        \makecell{LLE} & 
        \makecell{t-SNE} & 
        \makecell{MDS}   \\
        \midrule
        Double saddle        & 1.26 & 0.66 & 0.76 & 0.76 & \textbf{0.01} \\
        Noisy double saddle & 1.27 & 0.68          & 0.76 & 0.76 & \textbf{0.01} \\
        Loop of double saddles & 0.31 & 0.32          & 0.74 & 0.81 & \textbf{0.11} \\
        Noisy loop of double saddles & 0.32 & 0.32         & 0.77 & 0.81 &  \textbf{0.12} \\
        \bottomrule
    \end{tabular}
  \end{adjustbox}
\end{table*}


In this section we investigate the effectiveness of MDS for manifold learning in a controlled setting. In this simulated setting, we know the ground truth and can compare learning results from MDS and other non-linear embedding techniques that rely on different objective functions.
The simulation takes some smooth manifolds in a low-dimensional space, embeds them in a higher-dimensional Euclidean space, and uses several learning techniques to estimate the original manifolds. We use two manifolds in this study, each inspired by shapes observed in later real-data experiments. 

We embed points on these manifolds into a large space ($\real^{768}$) by appending zeros and applying a random rotation from the orthogonal group $O(768)$.
These high-dimensional points are then mapped back into the original lower dimensions using a variety of manifold learning methods: 
MDS, Isomap, Laplacian Eigenmaps, \ac{LLE}, and t-SNE. This is performed both with and without adding random Gaussian noise to the points in $\real^{768}$ before embedding.

In this section and throughout the paper, we include three-dimensional plots representing manifolds whose embedding dimensions are low but higher than three.
In creating these images, we express the embedded points in a \ac{PCA} basis such that the variance along each successive axis is nonincreasing. In cases where multiple sets of latent space points are being compared, the PCA is defined using the points from all sets under consideration.
\\

\noindent {\bf Case 1: Curve with Double-Saddle Shape}:
The first manifold is a double loop where each loop has a saddle shape, as shown in the top-left panel of Fig.~\ref{fig:comparison_dsaddle_noisy}. 
It has the one-dimensional topology of a circle and is embedded in $\real^4$ with positions specified by a single parameter, say $\psi$:
$x_1(\psi)=\cos(2\psi)$,
$x_2(\psi)=\sin(2\psi)$,
$x_3=\cos(4\psi)$ with distortion added to separate the two loops, and
$x_4(\psi)=\frac{1}{4}\cos(\psi)$.
The higher amplitude in the first three axes makes it possible to select them correctly using PCA for 3D visual comparisons.


\setlength{\tabcolsep}{1pt} 
\renewcommand{\arraystretch}{0.1} 
\begin{figure*}[h]
  \centering
    \begin{tabular}{ccccccc}
      \hspace{-4mm}
      airplane 03 &
      airplane 04 &
      airplane 05 &
      plane &
      car &
      Koenigsegg &
      Porsche
      \\
      \hspace{-4mm}
      \includegraphics[width=0.134\textwidth, scale=1]{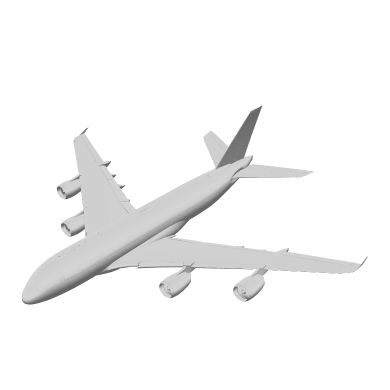} &
      \includegraphics[width=0.134\textwidth, scale=1]{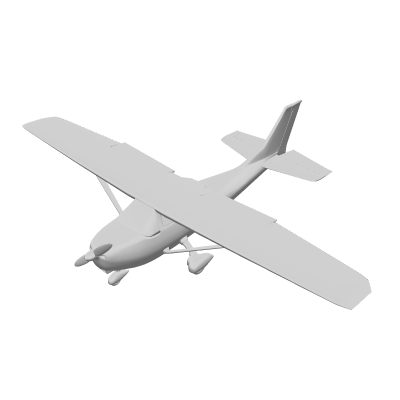} &
      \includegraphics[width=0.134\textwidth, scale=1]{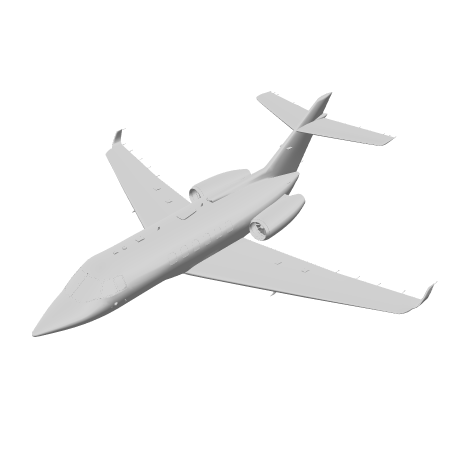} &
      \includegraphics[width=0.134\textwidth, scale=1]{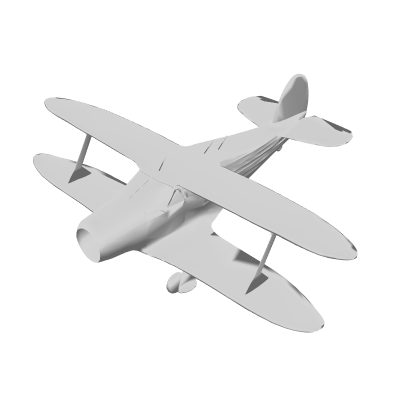} &
      \includegraphics[width=0.134\textwidth, scale=1]{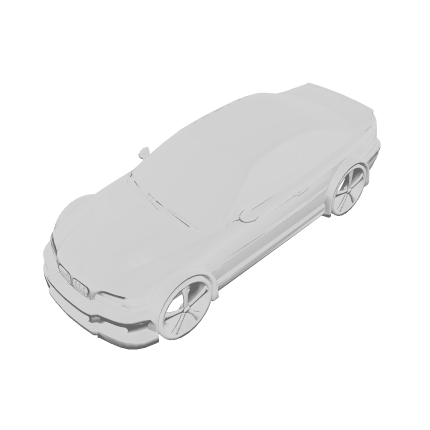} &
      \includegraphics[width=0.134\textwidth, scale=1]{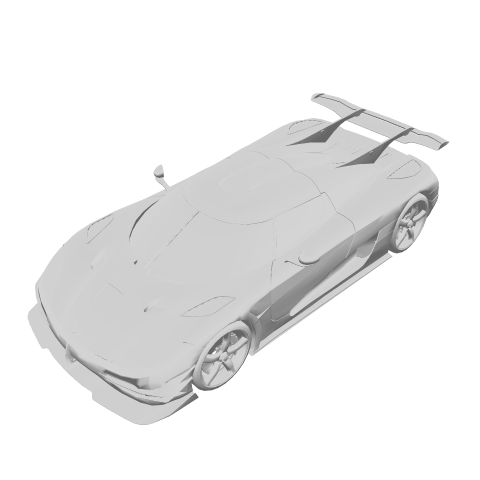} &
      \includegraphics[width=0.134\textwidth, scale=1]{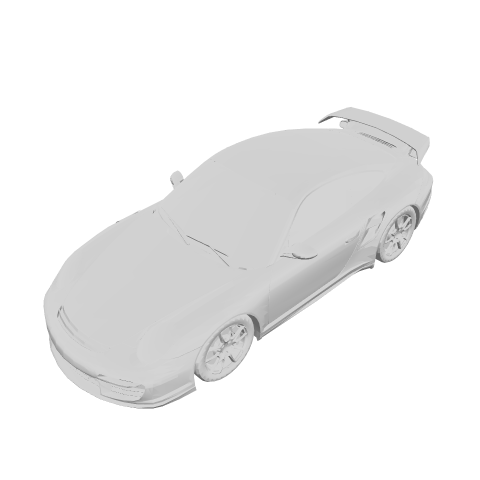} 
      \\
      \hspace{-4mm}
      \includegraphics[width=0.134\textwidth, scale=1]{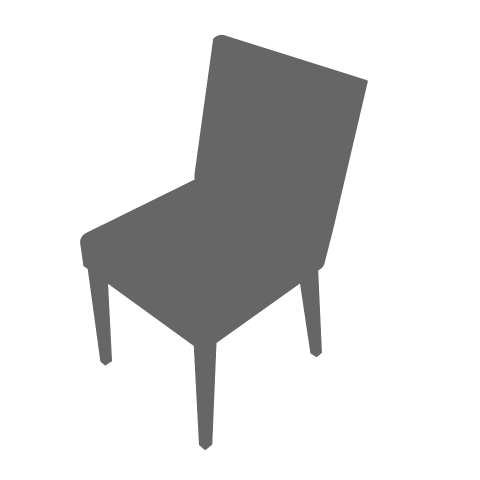} &
      \includegraphics[width=0.134\textwidth, scale=1]{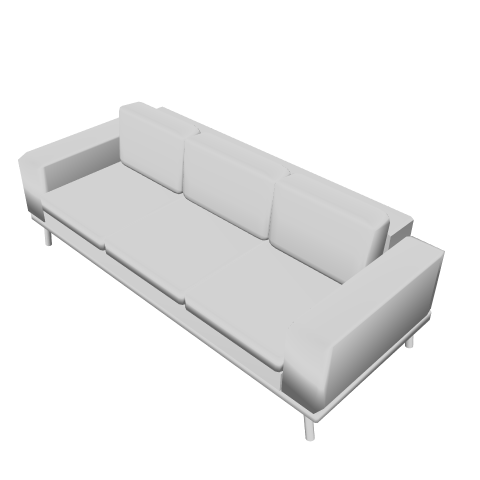} &
      \includegraphics[width=0.134\textwidth, scale=1]{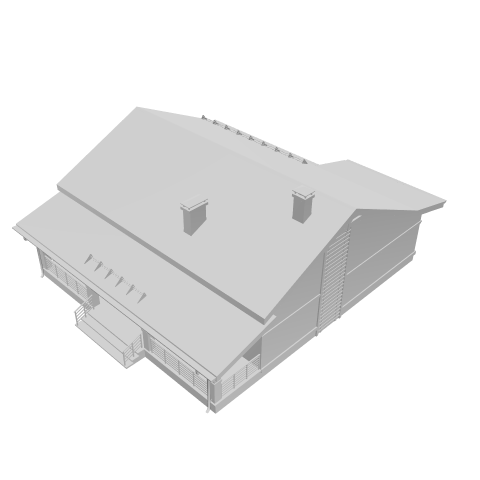} &
      \includegraphics[width=0.134\textwidth, scale=1]{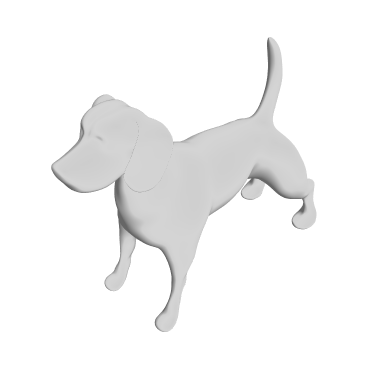} &
      \includegraphics[width=0.134\textwidth, scale=1]{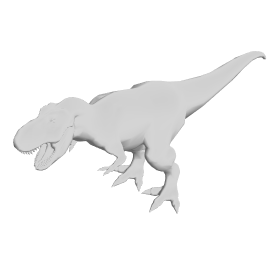} &
      \includegraphics[width=0.134\textwidth, scale=1]{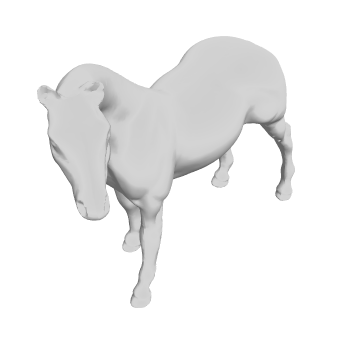} &
      \includegraphics[width=0.134\textwidth, scale=1]{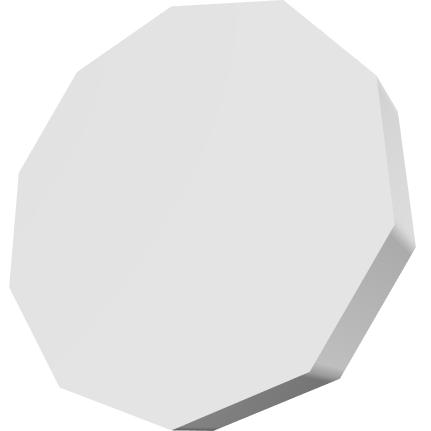}
      \\
      \hspace{-4mm}
      chair &
      Koltuk &
      cottage &
      dog &
      T. rex &
      zebra &
      prism 10
      \\
    \end{tabular}
    \caption{
      Images of the \ac{CAD} objects at their default positions and a prism angled to display its face.
      }
    \label{fig:objectImageExamples}
\end{figure*}
\setlength{\tabcolsep}{6pt} 
\renewcommand{\arraystretch}{1.0} 


The top row of Fig.~\ref{fig:comparison_dsaddle_noisy} plots manifolds estimated using MDS, Isomap, Laplacian Eigenmaps, LLE, and t-SNE.  
The MDS embedding is visually much more true to the originals than the others. 
This is simply explained by considering the nature of each method.
MDS is expected to have good performance on this task, since it is known by construction that the global Euclidean distance matrix from $\real^{768}$ can be perfectly embedded in $\real^4$ with the desired form.
Isomap attempts to embed a matrix of global graph geodesic distances, and any intrinsically isometric deformation of the original curve would create an identical objective. The manifold has many spatially nearby points that are far away intrinsically, so Isomap is not expected to faithfully recover the original curve.
The other methods are designed to preserve local distances while paying less attention to larger ones. For a one-dimensional manifold with consistently proximate neighboring points, there is little enforcing adherence to the global structure.
In particular, the t-SNE similarity function $\exp(-\Vert x_i-x_j \Vert^2 / 2\sigma^2)$ makes almost no distinctions between any but the smallest distances, and its stochastic nature seeds a chaotic global appearance with nothing to rein it in. Furthermore, the authors of \cite{van2008visualizing} note that the basic application of t-SNE is not well-suited to embedding in dimensions higher than three.

We also evaluate the results quantitatively by computing the shape distances (eq.~\ref{eq:shapeDistance}) between each embedding and the ground truth.
These values are shown in Table~\ref{tab:sim_quant_compare}. MDS substantially outperforms all of the other methods by this metric.
\\

\noindent {\bf Case 2: Surface of Loop of Double Saddles}: 
The second simulated object is formed using a loop of double saddles. It has the two-dimensional topology of the $\torus^2$ torus and is embedded in $\real^6$ with positions specified by the two parameters, say $(\phi,\psi)$:
axes $x_1,x_2,x_3$ are formed by rotating first three axes of the one-dimensional double saddle around the $x_3$-axis in a circle parameterized by $\phi$, then 
$x_4(\phi,\psi)=\frac{1}{4}\cos(\psi)$,
$x_5(\phi,\psi)=\frac{1}{4}\cos(\phi)$, and
$x_6(\phi,\psi)=\frac{1}{4}\sin(\phi)$.

Again, we plot visual comparisons of the manifolds produced by each embedding method, and compute the shape distances between them and the ground truth.
The bottom row of Fig.~\ref{fig:comparison_dsaddle_noisy} shows MDS to give the best results, although less dramatically so than before.
The higher intrinsic dimensionality of the manifold and its radial symmetry make it more difficult to warp the curved surface without affecting local distances. We see visually and quantitatively that the local methods produce reconstructions that are globally more similar to the original than in the 1D case.
Table~\ref{tab:sim_quant_compare} shows that MDS again substantially outperforms the other methods by the shape distance metric.

\section{Experimental Results: Pose Image Manifolds}
\label{sec:poseManifolds}

\noindent
In this section, we present results on learning of pose image manifolds and submanifolds, and comparisons of their shapes. 

~~\\
\noindent {\bf Experimental Setup}: 
We form pose image manifolds $\imageSpace^{\object}_{\soThreeSet}$ by applying structured sets of rotations $\soThreeSet$ to 3D objects $\object$ and collecting the 2D images at each pose.
These are mapped with MDS to latent image manifolds $\latentManifold_{\soThreeSet}^{\object} = \latentMap_{MDS,\soThreeSet}^{\object}(\imageSpace^{\object}_{\soThreeSet})$. 
We compute the pairwise shape distances between these manifolds using $d_{\soThreeSet}$ from Eqn.~\ref{eq:shapeDistance}, display them as heatmaps, and view the proximities in two dimensions by performing a further MDS on the shape distance matrix.
When the intrinsic dimension of $\soThreeSet$ is 1 or 2, we also visualize examples of the first three PCA axes of $\latentManifold_{\soThreeSet}^{\object}$. It is more difficult to visualize the intrinsically 3-dimensional full-$SO(3)$ set, and we do not do so.

We primarily focus on 13 \ac{CAD} objects, shown in Fig~\ref{fig:objectImageExamples} at their default poses corresponding to the identity rotation. These are standardized so that the top and front of each object point approximately along the positive $z$ and $x$ axes respectively.
The names we use are simplified versions of those in the source data files. We note that {\it Koenigsegg} and {\it Porsche} are sports cars, and that {\it Koltuk} is a sofa.
To analyze symmetry in a controlled setting, sections~\ref{ssec:so2Data} and~\ref{ssec:t2Data} also use prism surfaces constructed for the purpose using the Plotly package in Python \cite{plotly,pythonRef09}.
Each prism is built by connecting two parallel copies of a regular polygon. We use 11 prisms corresponding to polygons with side numbers 
$\{3,4,...,10,50,100,1000\}$, 
and name them according to the number of sides. The default poses of the prisms have the faces in horizontal planes. An image of the prism with 10 sides is shown in Fig.~\ref{fig:objectImageExamples}.

Sections~\ref{ssec:so2Data}, \ref{ssec:t2Data}, and~\ref{ssec:completeSo3Data} respectively sample subsets of $SO(3)$ which are intrinsically 1-, 2-, and 3-dimensional. The sampling structures are detailed therein.
When indexing $SO(3)$ points, we use a Hopf coordinate system similar to Yershova~et~al.~\cite{yershova2010generating}, with points expressed by triples $(\theta,\phi,\psi) \in [0,\pi]\times[0,2\pi)\times[0,2\pi)$. 
The coordinates can be understood in terms of the resulting orientation of a three-dimensional object after the corresponding $SO(3)$ action has been applied to it when starting from a specified default position. The default identity position implies an internal north pole of the object and an internal longitude-zero prime meridian. The $(\theta,\phi)$ pair are equivalent to the standard $\sphere^2$ spherical (zenith,azimuth) coordinates, and indicate the resulting direction of the internal north pole after the action is applied. The $\psi$ coordinate indicates the heading of the internal prime meridian. The orientation $(\theta,\phi,\psi)$ can be reached from the identity by composing a rotation of $\psi$ radians about the extrinsic $z$-axis followed by the appropriate rotation about an $xy$-plane axis that brings the internal north to point in the $(\theta,\phi)$ direction.
As is the case with spherical coordinates, the Hopf coordinates have degeneracies at $\theta\in\{0,\pi\}$.
Diagrams giving abstract representations of the $SO(3)$ samples used are available in the supplement.

Section \ref{ssec:changingLightSource} extends the analysis to illumination image manifolds, performing analogous experiments using image sets generated by moving a single light source along a circular path while the object remains stationary at its default position.

The latent manifold visualizations in this section and the rest of the paper use a simple smoothing technique which iteratively replaces each point with a weighted mean of itself and its parameter space neighbors.
The number of smoothing steps used is dependent on the parameter density. Manifolds which are intrinsically one-dimensional use ten smoothing steps, and the sparser two-dimensional manifolds use five smoothing steps. 

\subsection{Pose Submanifolds: Single-Angle or $SO(2)$  Parameterizations}
\label{ssec:so2Data}


\begin{figure*}[h]
  \begin{center}
    \includegraphics[width=0.244\textwidth, scale=1]{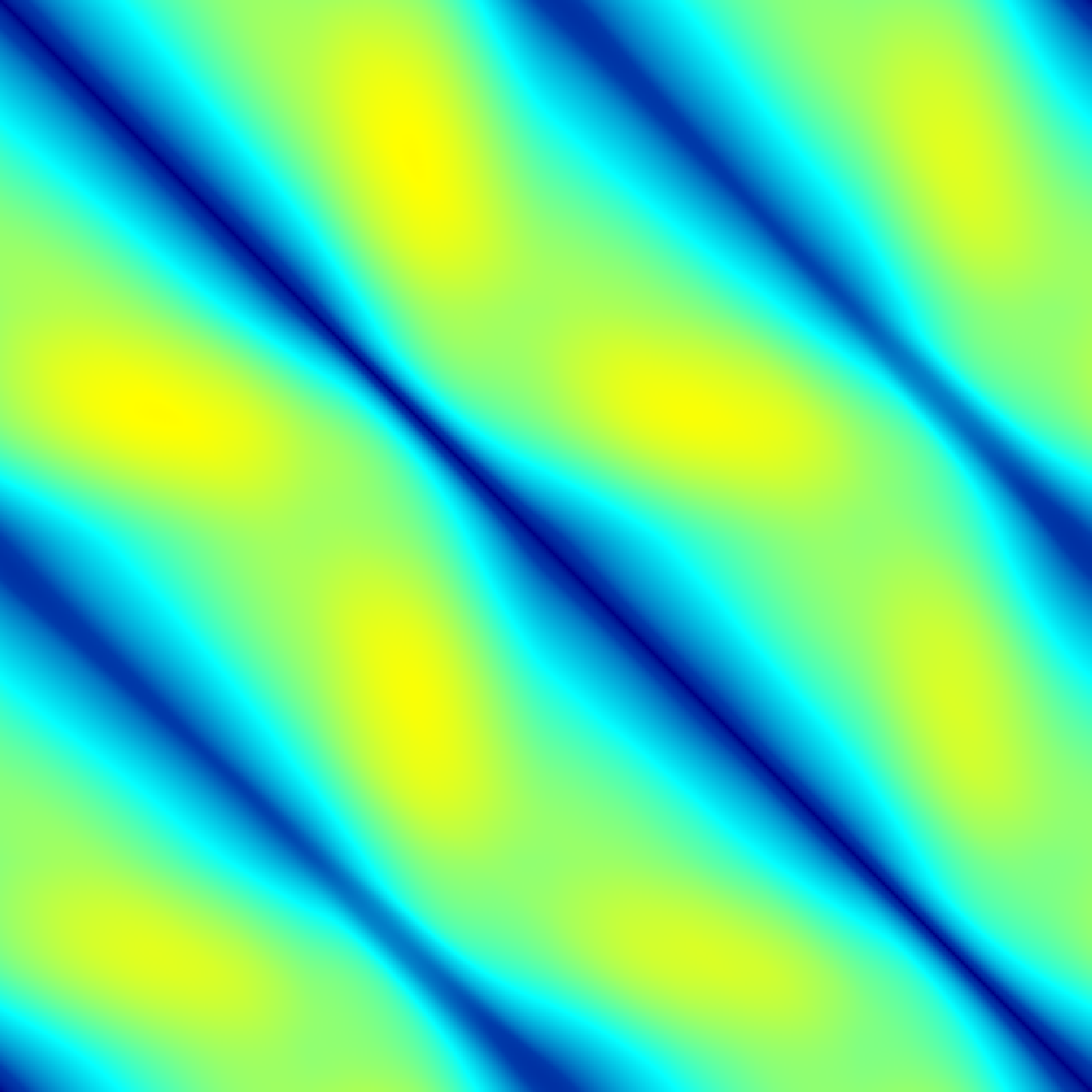}
    \includegraphics[width=0.244\textwidth, scale=1]{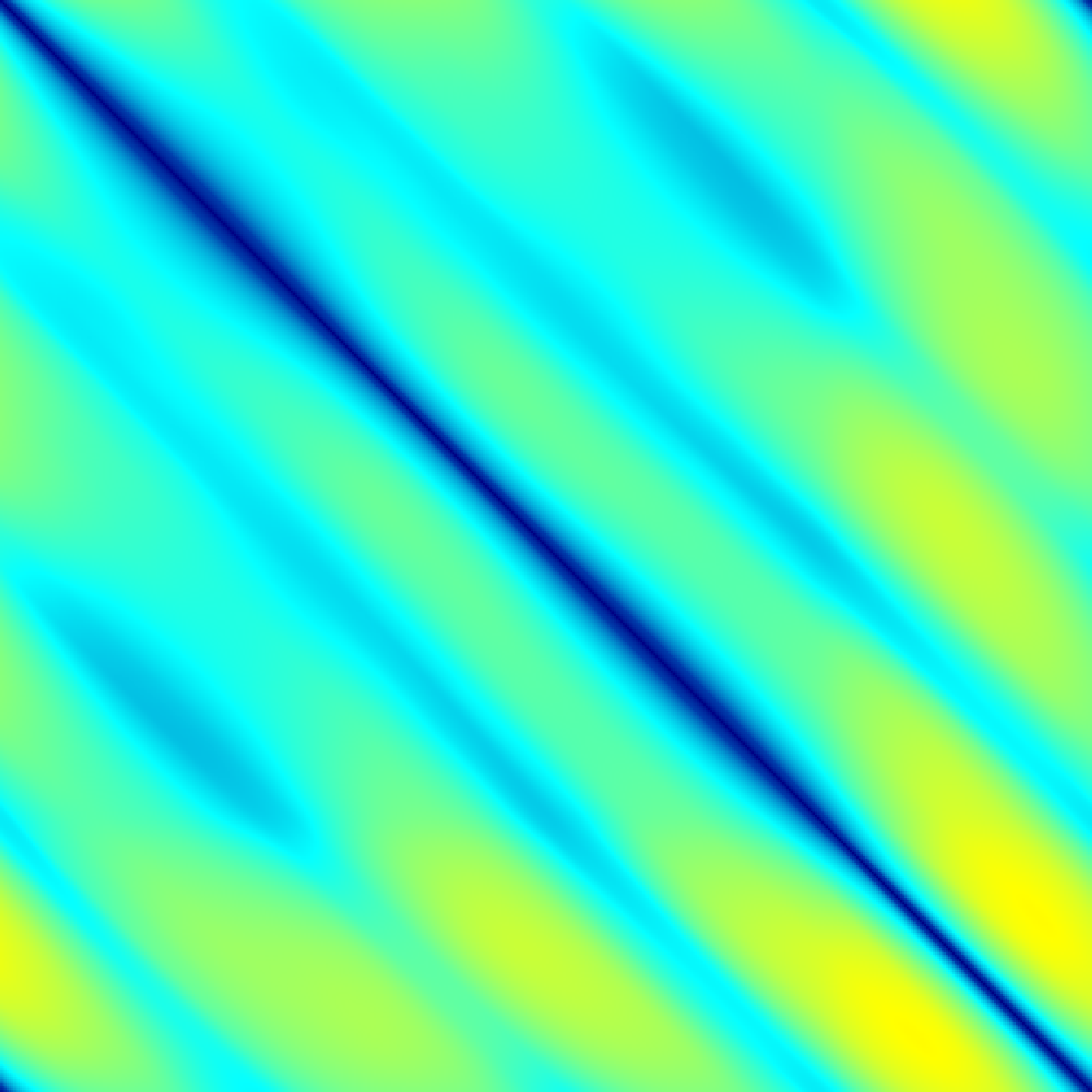}
    \includegraphics[width=0.244\textwidth, scale=1]{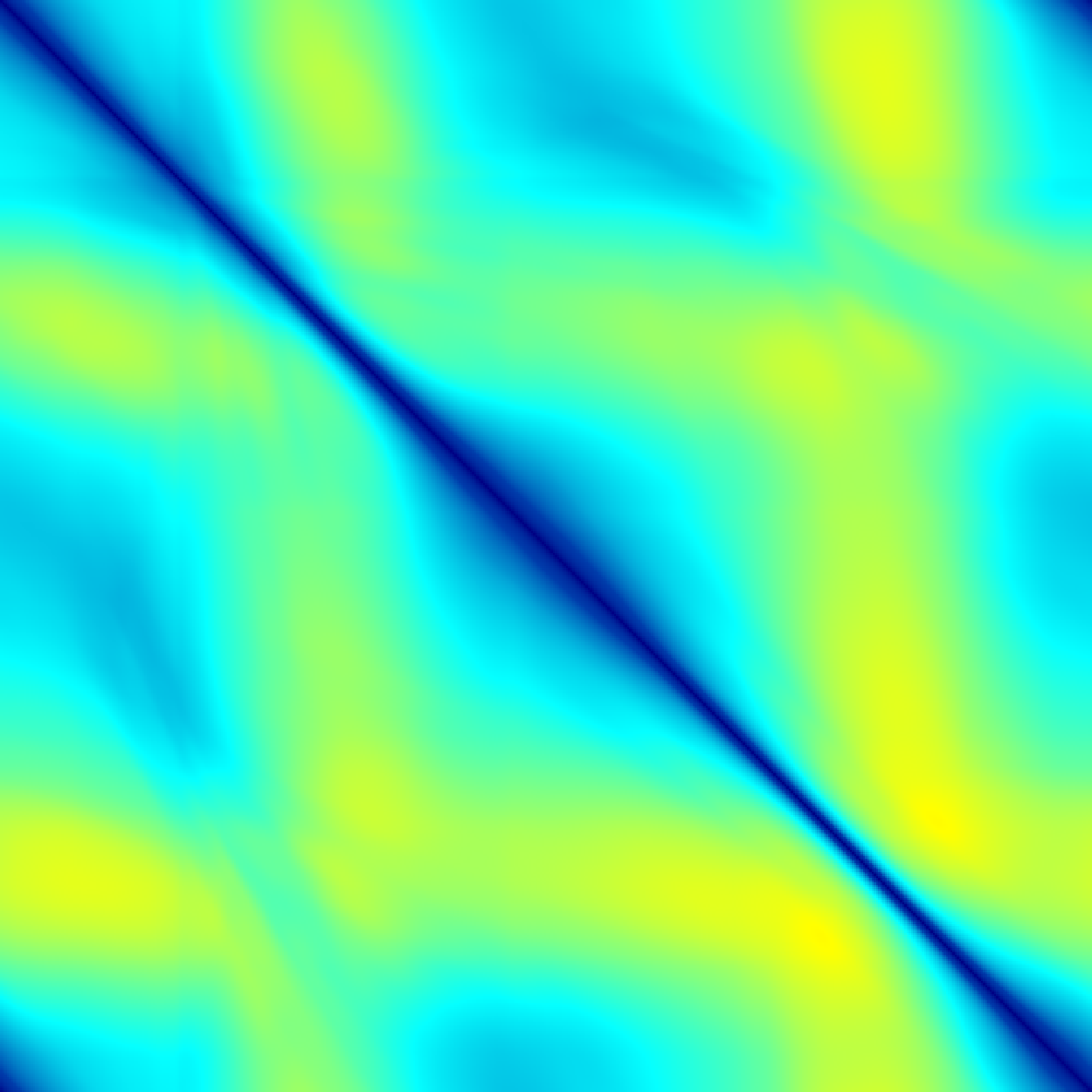}
    \includegraphics[width=0.244\textwidth, scale=1]{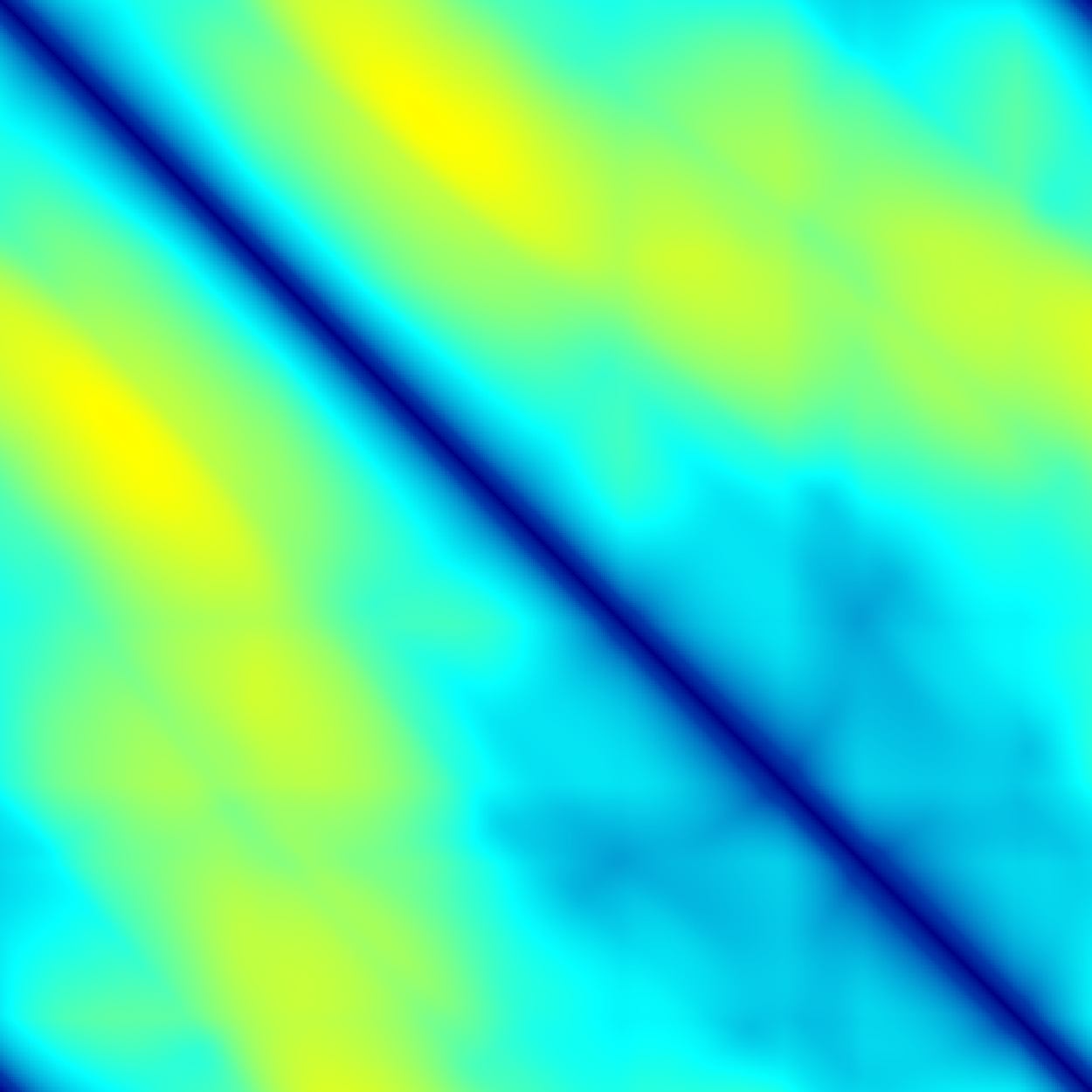}
    \includegraphics[width=0.244\textwidth, scale=1]{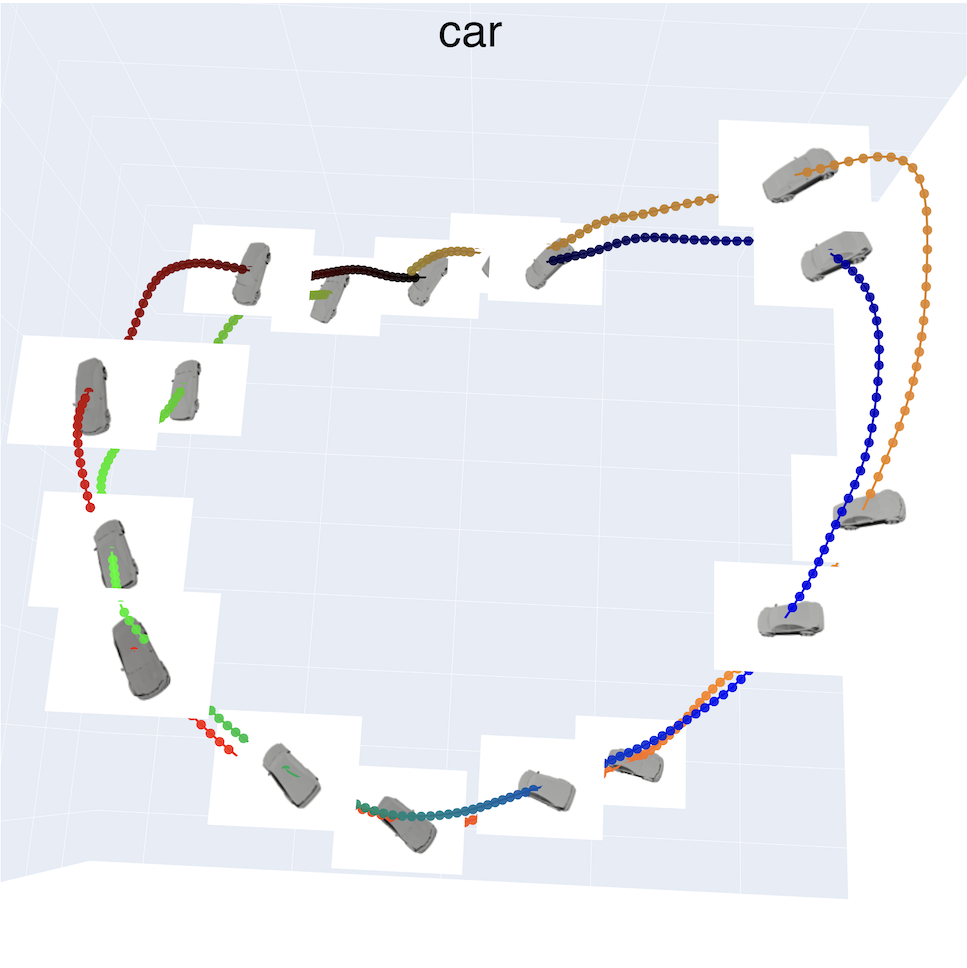}
    \includegraphics[width=0.244\textwidth, scale=1]{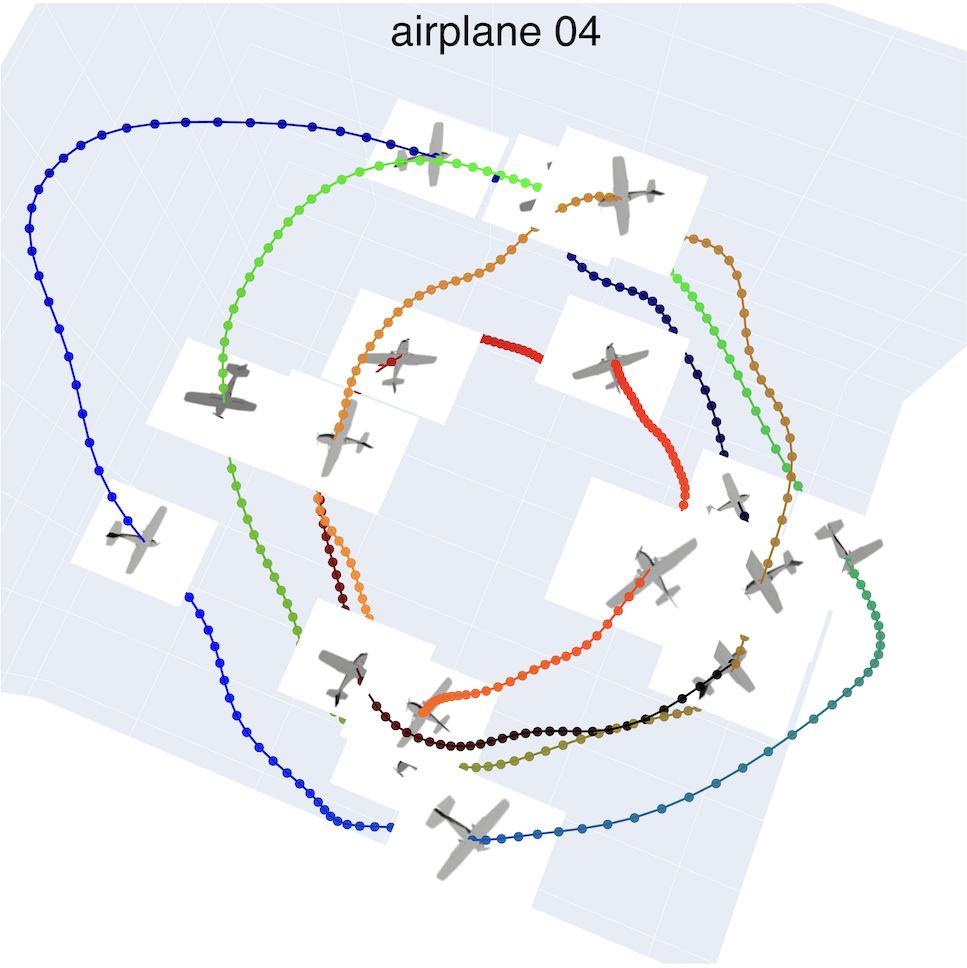}
    \includegraphics[width=0.244\textwidth, scale=1]{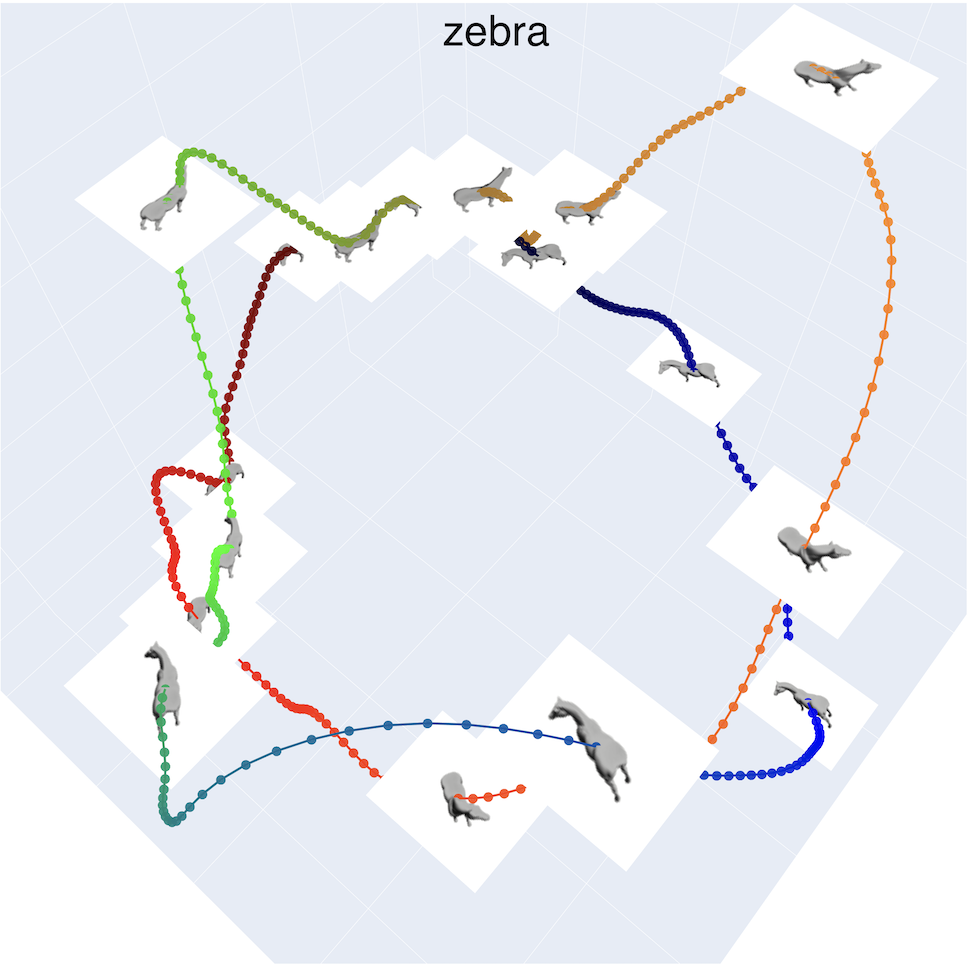}    
    \includegraphics[width=0.244\textwidth, scale=1]{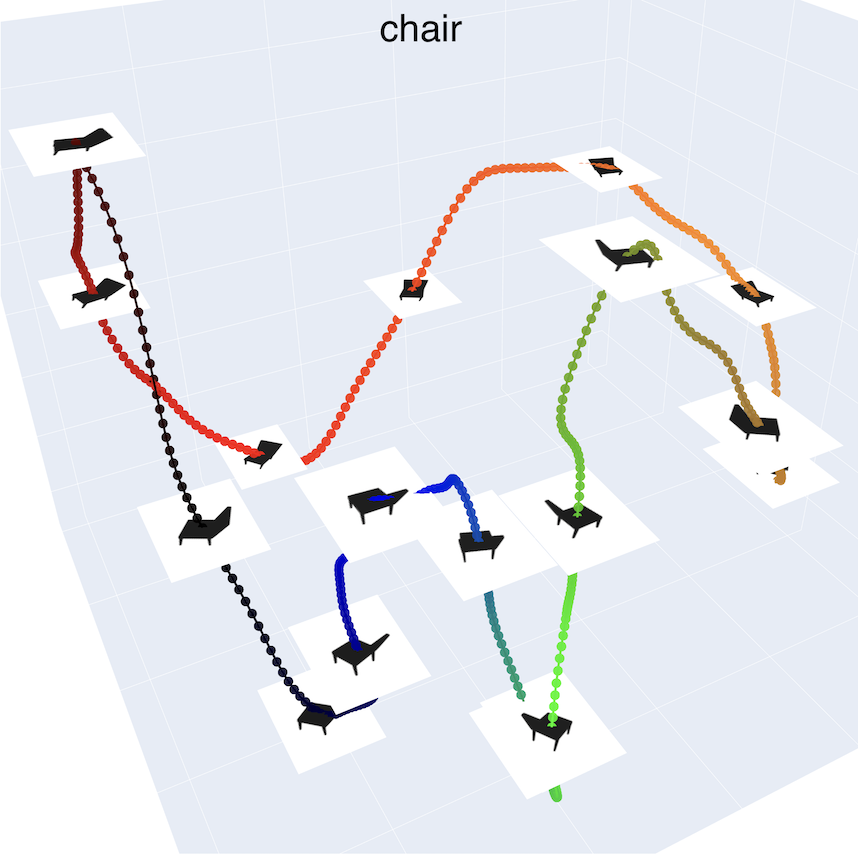}
    \includegraphics[width=0.244\textwidth, scale=1]{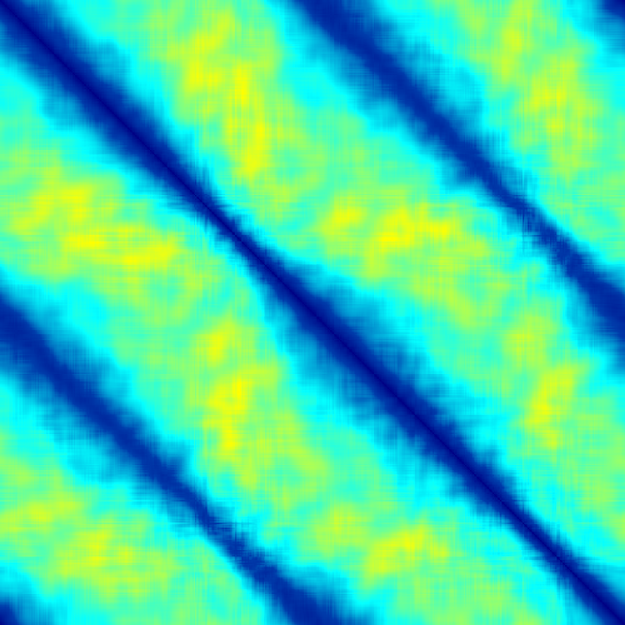}
    \includegraphics[width=0.244\textwidth, scale=1]{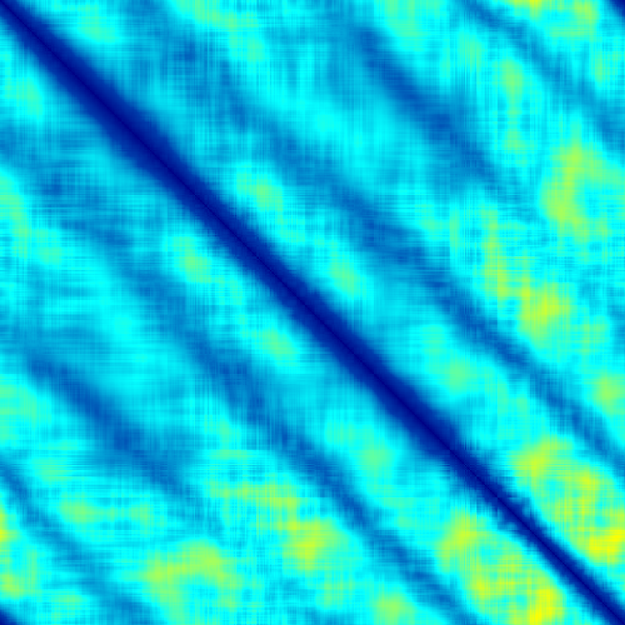}
    \includegraphics[width=0.244\textwidth, scale=1]{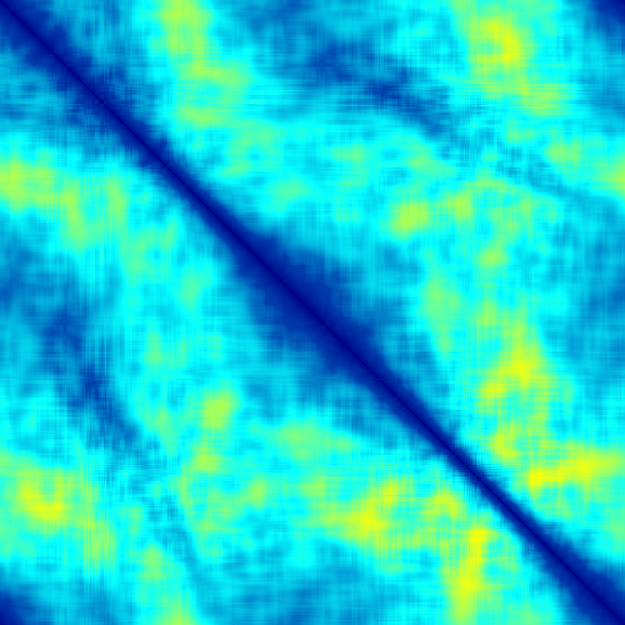}
    \includegraphics[width=0.244\textwidth, scale=1]{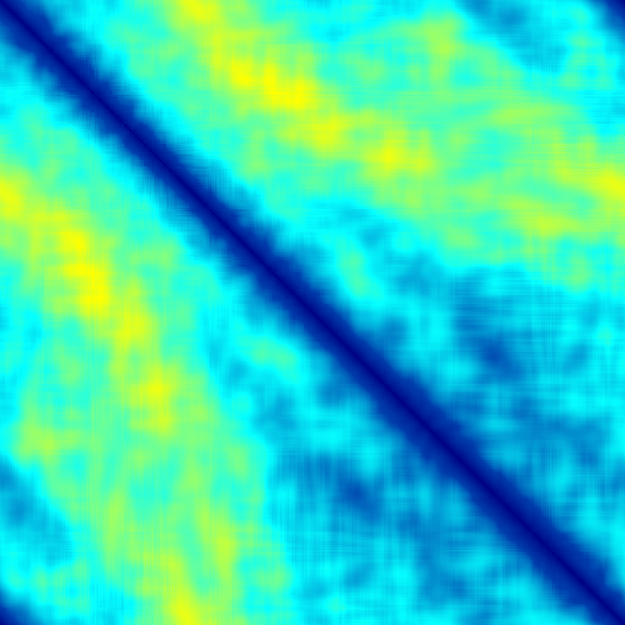}
  \caption{
    \emph{Middle row}:
      $SO(2)$ MDS latent space axes 1-3 for the {\it car}, {\it airplane 04}, {\it zebra}, and {\it chair}.
    \emph{Top row}:
      Image space distance matrices for corresponding objects in the middle row.
      Scales range from dark blue (low) to yellow (high).
    \emph{Bottom row}:
      Latent space distance matrices. 
    }
  \label{fig:so2LatentWithImages_1}
  \end{center}
\end{figure*}


\begin{figure*}[h]
  \begin{center}
    \includegraphics[width=0.59\textwidth, scale=1]{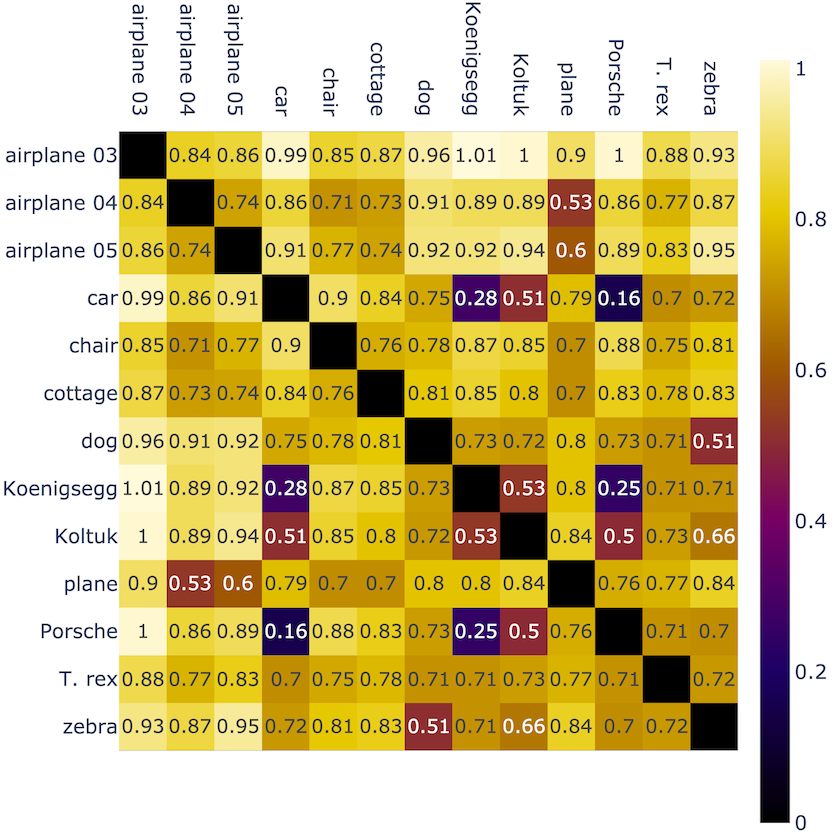}
    \includegraphics[width=0.39\textwidth, scale=1]{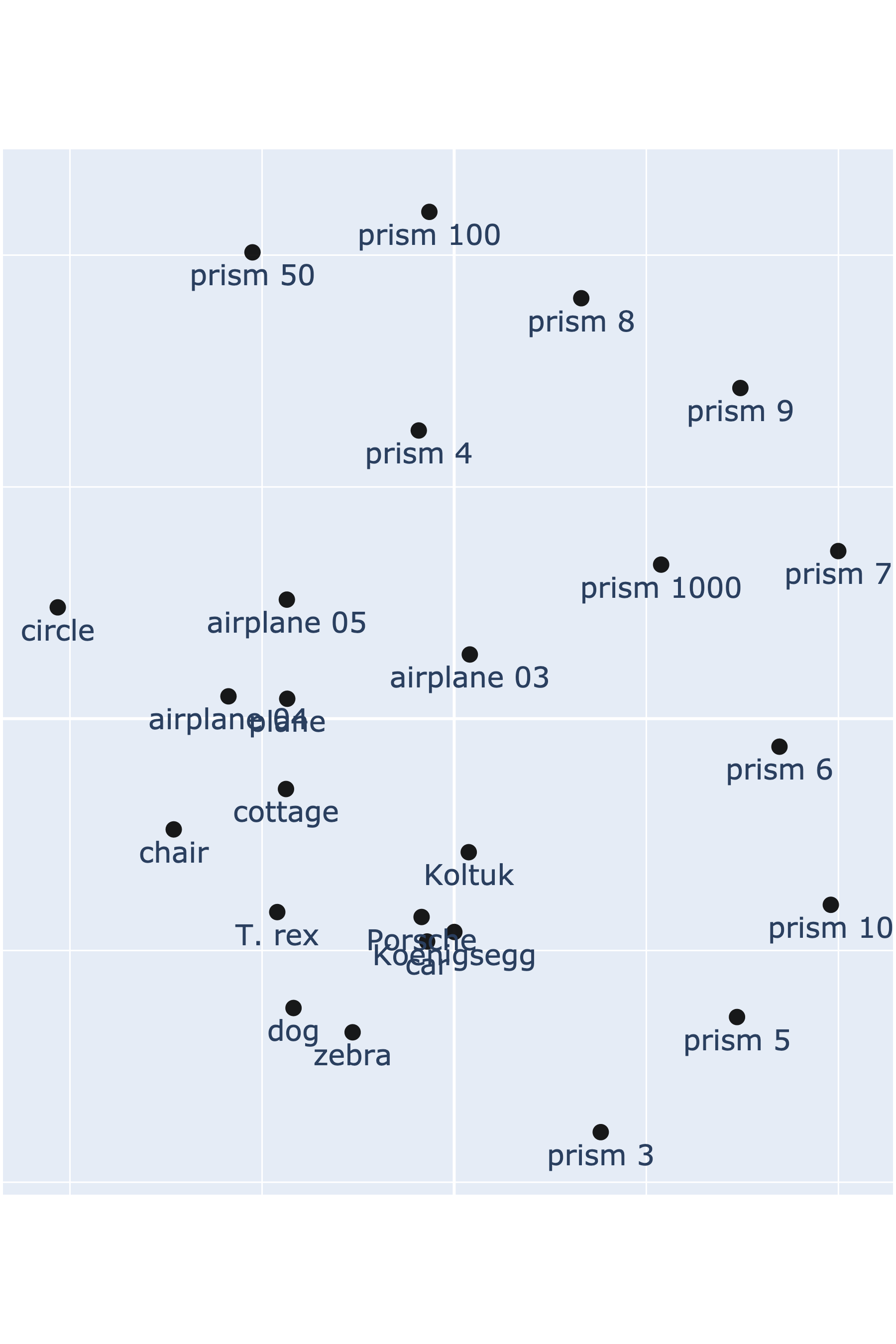} 
    \caption{
      \emph{Left}: Latent space shape distances on the $SO(2)$ subset.
      \emph{Right}: Two-dimensional MDS positions produced by the distances.
      }
  \label{fig:so2ShapeDistances}
  \end{center}
\end{figure*}


\begin{figure*}
  \begin{center}
    \includegraphics[width=0.23\textwidth, scale=1]{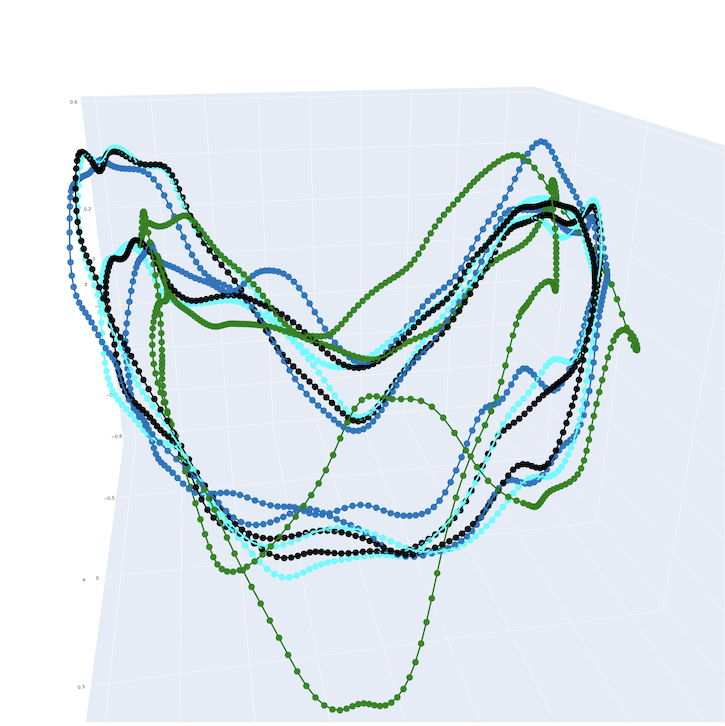}
    \includegraphics[width=0.10\textwidth, scale=1]{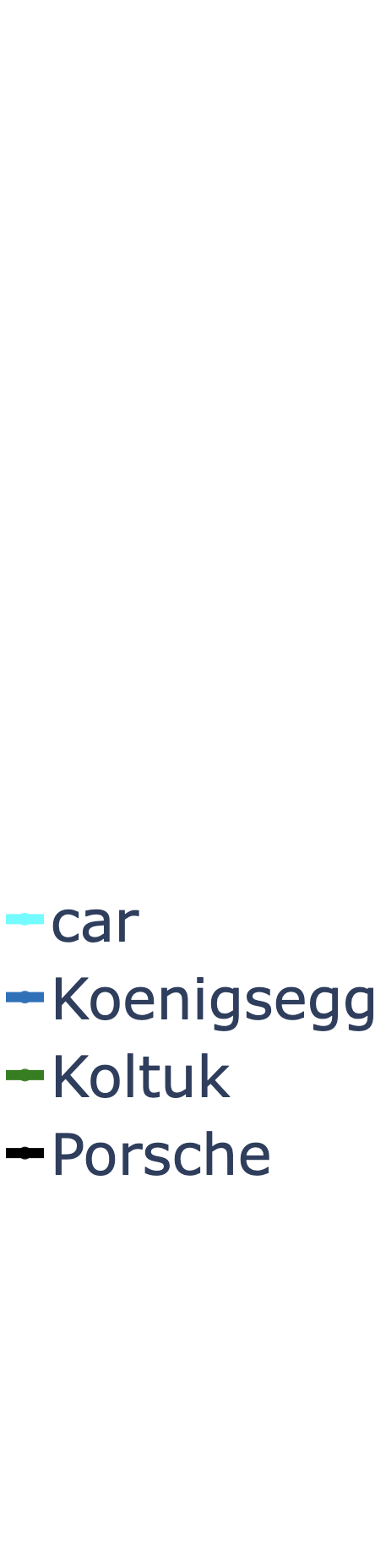}
    \includegraphics[width=0.23\textwidth, scale=1]{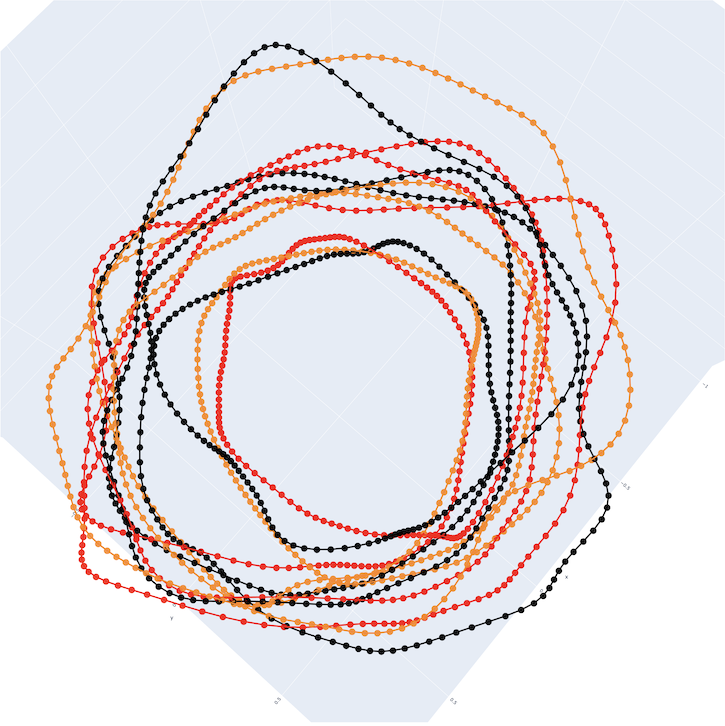}
    \includegraphics[width=0.10\textwidth, scale=1]{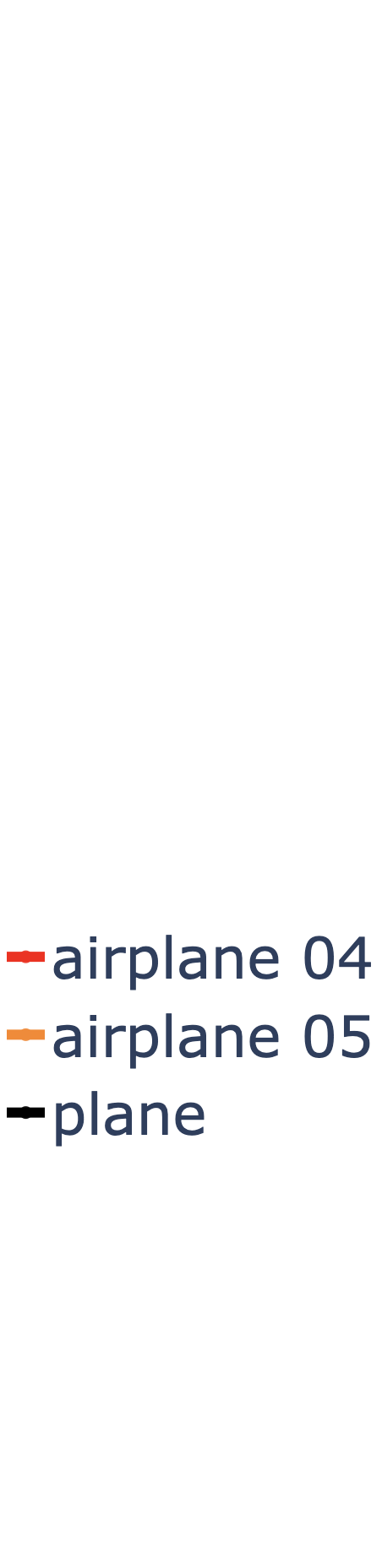}
    \includegraphics[width=0.23\textwidth, scale=1]{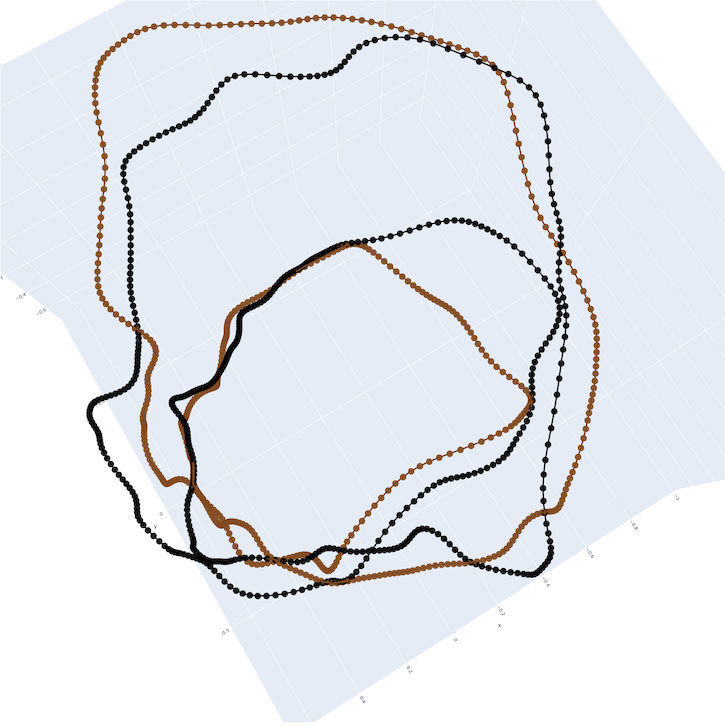}
    \includegraphics[width=0.06\textwidth, scale=1]{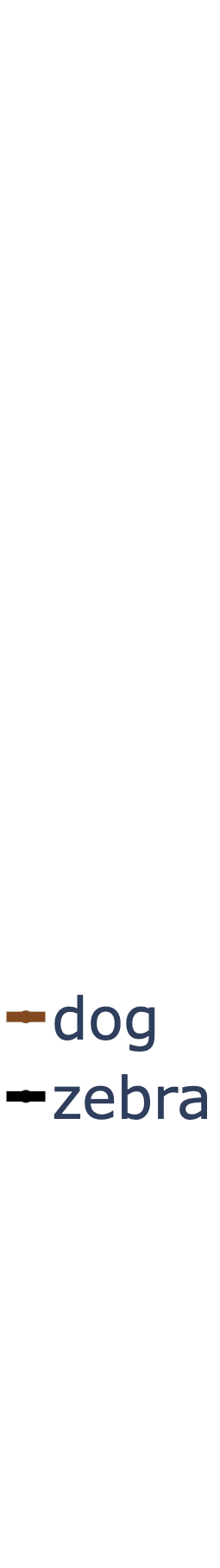}
  \caption{
    $SO(2)$ MDS latent spaces for groups of similar objects. In each panel, colored curves are aligned to the black curve.
  }
  \label{fig:so2LatentSpaceClusters}
  \end{center}
\end{figure*}


The data in this section uses the one-dimensional subset $\soThreeSet_1$ of $SO(3)$ sampled from a copy of $SO(2)$ by keeping the coordinates $(\theta,\phi)=(\pi/4,0)$ fixed and creating a circle of 500 evenly spaced points in the $\psi$ coordinate.
This can be thought of as the poses formed by leaning the object forward until it is halfway towards reaching the vertical position, then rotating it one full turn about its internal $z$-axis as in Fig.~\ref{fig:samplingProcesses}.
The objects have their top aspect prominently visible at $\soThreeSet_1$ poses.
We map the generated image sets $\imageSpace^{\object}_{\soThreeSet_1}$ for the 13 \ac{CAD} objects and 11 prisms to an $8$-dimensional latent space using $\latentMap_{MDS}$.

~\\
\noindent {\bf Manifold Visualizations}: 
Each column of Fig.~\ref{fig:so2LatentWithImages_1} shows a visualization of one of these latent spaces, along with its pointwise Euclidean distance matrices in the image space and latent space.

The distance matrix for the {\it car} displays a strong Toeplitz-like structure, showing that the distance beetween any two points is mostly determined by the angular difference between the poses. This can be explained by the plane of rotation's orientation being nearly orthogonal to the camera position, and to the flat top aspect of the {\it car}. The dark diagonals at angular difference intervals of $\pi$ illustrate the {\it car}'s two-fold rotational symmetry.
A similar but weaker structure is seen for the {airplane 04}, with darker diagonals at intervals of $\frac{\pi}{2}$ showing its four-fold symmetry.
The further departure from the Toeplitz structure for the {\it zebra} and {\it chair} can be attributed to the more pronounced variations in the objects' heights.

The latent space visualizations show the $500$ ordered points with corresponding images included at regular intervals.
The first three PCA axes are plotted. These generally capture about 50-75\% of the total variation for the $SO(2)$ latent spaces, and the further spectral values are nearly equal. While there is still substantial variation beyond the third axis, we found values on the higher axes to be more irregular and lacking interpretive value.


\begin{figure*}[h]
  \begin{center}
  \includegraphics[width=0.215\textwidth, scale=1]{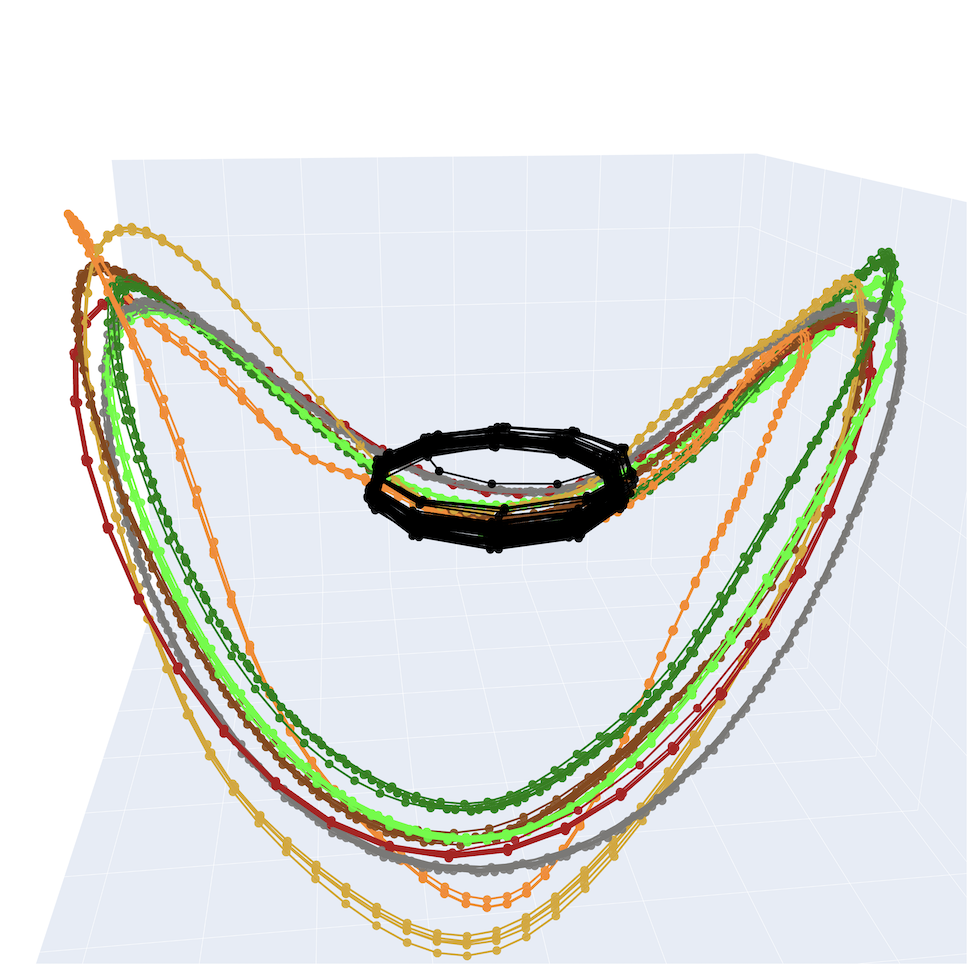}
  \includegraphics[width=0.215\textwidth, scale=1]{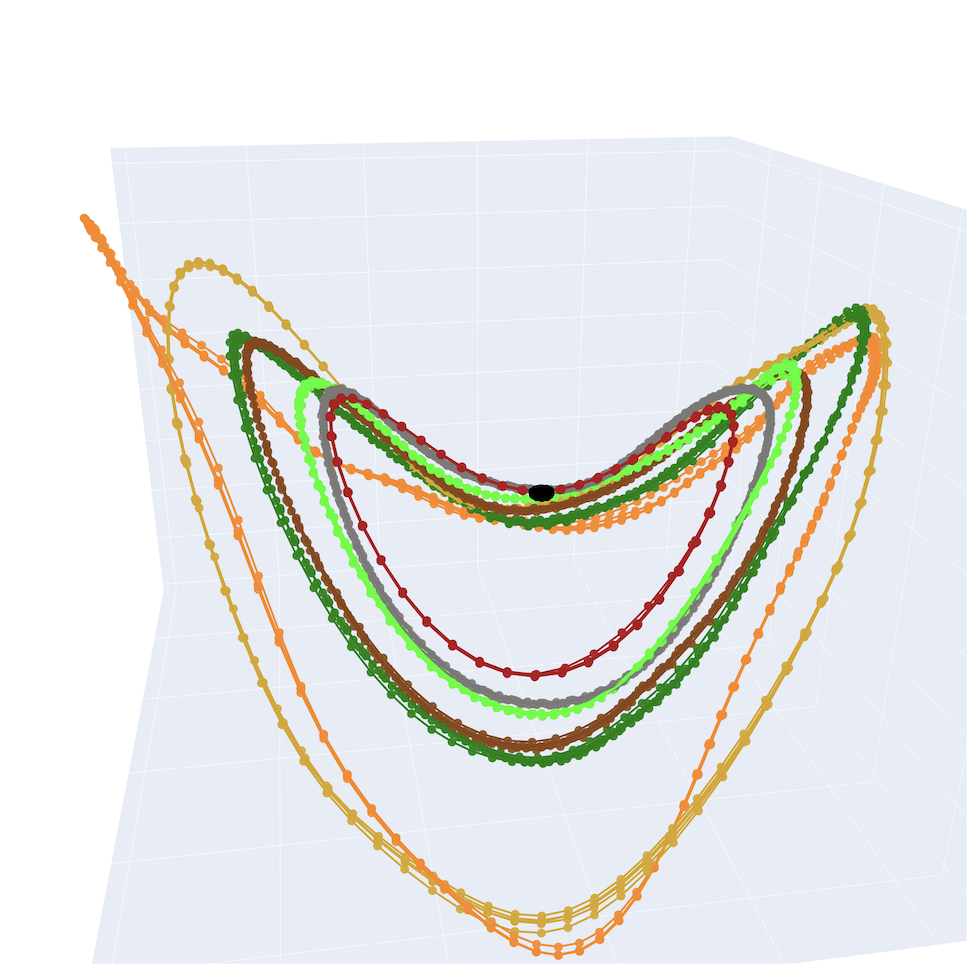}
  \includegraphics[width=0.215\textwidth, scale=1]{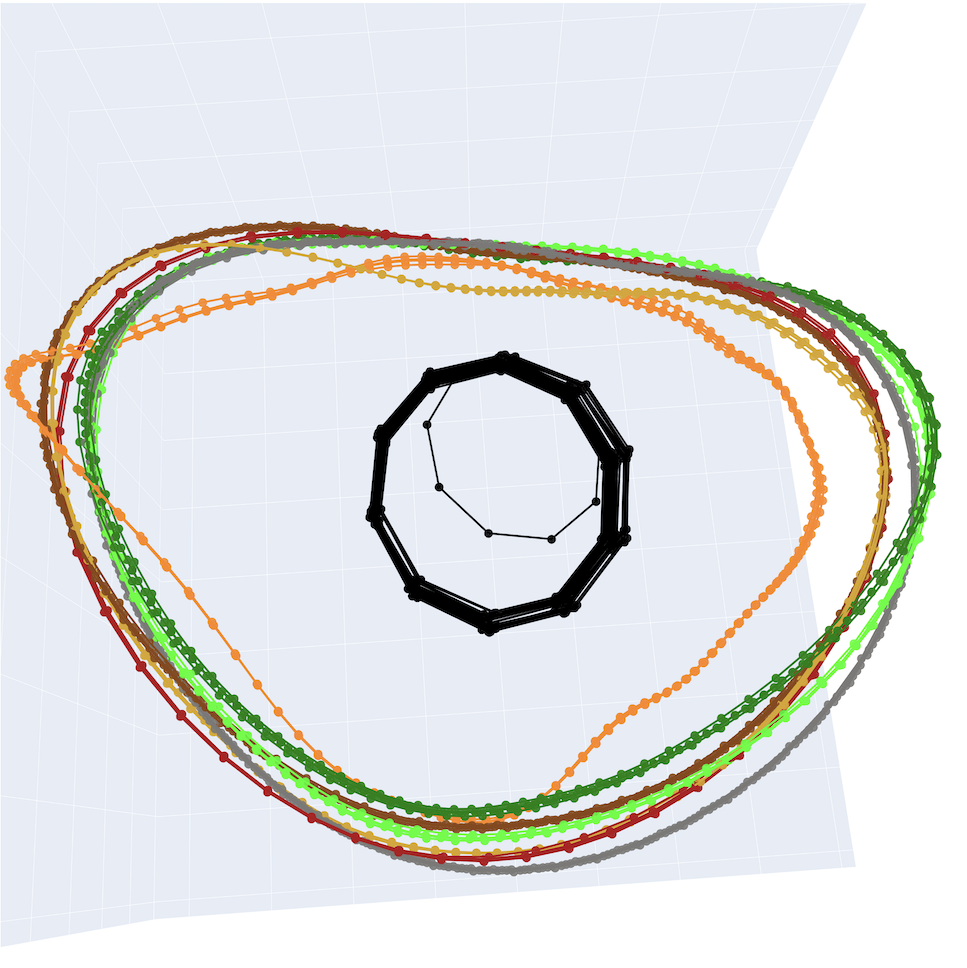}
  \includegraphics[width=0.215\textwidth, scale=1]{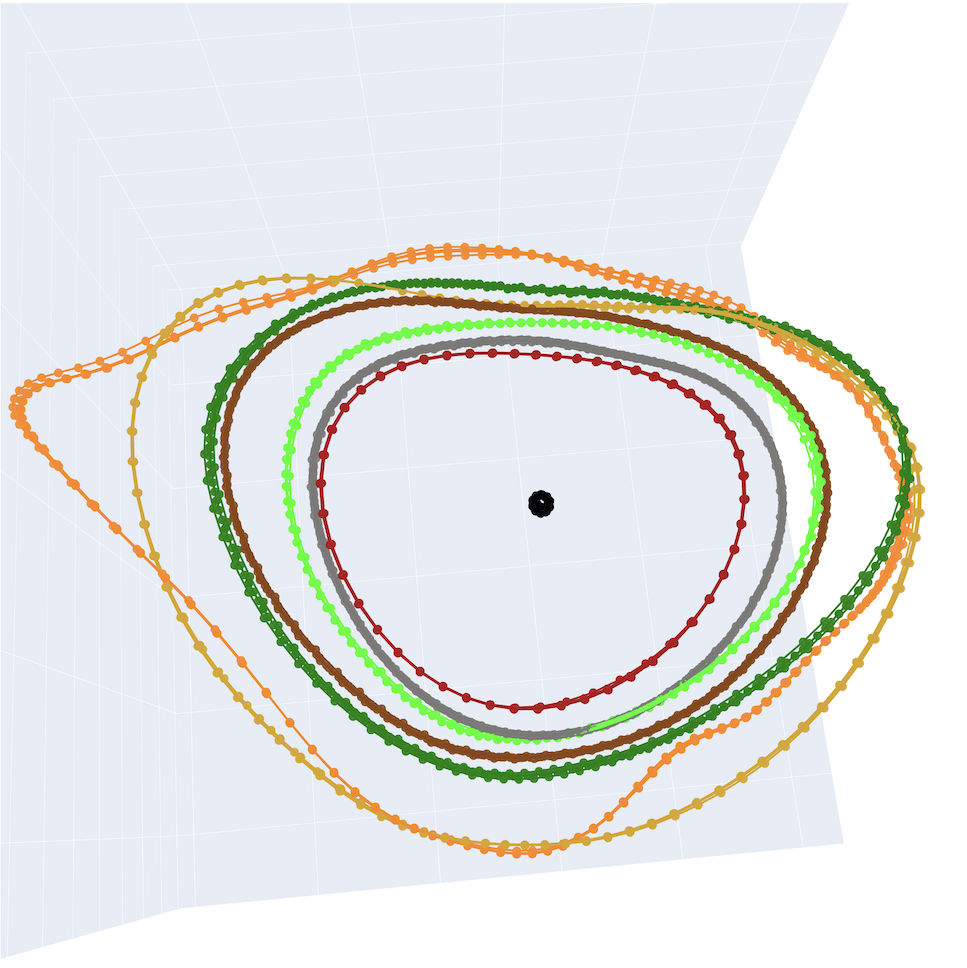}
  \includegraphics[width=0.11\textwidth, scale=1]{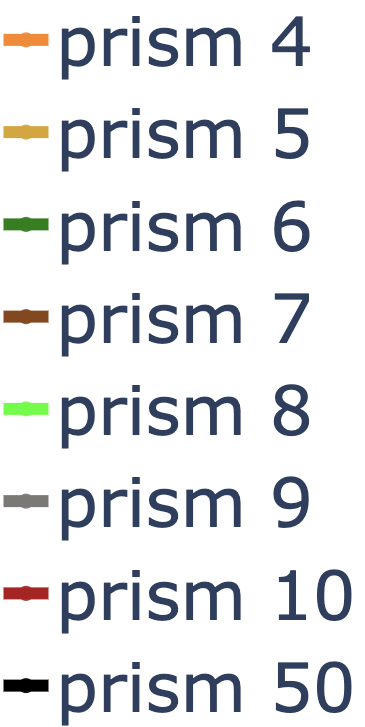}
  \caption{
    $SO(2)$ \ac{MDS} latent spaces for the prisms. All panels show the same data.
    Panels 1,3 from the left show sets standardized to a common scale. Panels 2,4 repeat panels 1,3 returned to original scales.
    }
  \label{fig:so2PrismAlignments}
  \end{center}
\end{figure*}


The latent space for the {\it car} clearly reflects its $\pi$-radian rotational symmetry, closely tracing the same simple path two times as the object completes a single rotation. Looking at nearby points on the separate loops, the images clearly show the {\it car} aligned along similar spatial axes but facing in opposite directions.
The {\it airplane 04} has weak rotational symmetry at $\pi/2$ radian intervals. The path for this object roughly traces four loops, and the images on corresponding regions of different loops show the object at rotational poses separated by intervals of $\pi/2$.
The {\it zebra} has a weak rotational symmetry of $\pi$ radians. Its latent path forms two loops which are loosely similar to each other, and at nearby points the images show the {\it zebra} aligned along the same spatial axis but facing in opposite directions.
The {\it chair} does not have any rotational symmetry for this pose set, and its latent path forms only a single loop.

~~\\
\noindent {\bf Shape Comparisons and Clustering}: 
The shape distances $d_{\soThreeSet_1}$ from Eqn.~\ref{eq:shapeDistance} and their two-dimensional proximity plot are shown in Fig.~\ref{fig:so2ShapeDistances}.
For visual clarity, we only include the subset of the distance matrix pertaining to the \ac{CAD} objects. A complete distance matrix is included in the supplement.
The distances between the prisms and the other objects, as well as amongst themselves, are generally much higher, mostly in the range $[1,1.4]$. 
We also include a simple geometric circle generated in $\real^2$ and embedded in $\real^8$, which has distances from most objects also approximately in $[1,1.4]$.

A few obvious clusters can be formed, and they correspond to natural categories of the objects:
  a car-type cluster \{{\it car}, {\it Koenigsegg}, {\it Porsche}\},
  an animal-type cluster \{{\it dog}, {\it zebra}\},
  and an airplane-type cluster \{{\it airplane 04}, {\it airplane 05}, {\it plane}\}.
The top of the {\it Koltuk} has a rectangular appearance similar to the top of the cars, and it is close to them in shape space for this set of poses.
The {\it airplane 03} has distances that set it apart from the airplane-type cluster. Its wings form a sharper angle with the fuselage than the others, bringing it farther away from the four-part symmetry seen for the {\it airplane 04} in Fig.~\ref{fig:so2LatentWithImages_1}.
The latent spaces for these clusters are visualized in Fig.~\ref{fig:so2LatentSpaceClusters}.

~~\\
\noindent {\bf Prisms and Symmetry}: 
The highly symmetric prism objects provide a clear demonstration of how object symmetries correspond to shape in the latent space manifold.
Fig.~\ref{fig:so2PrismAlignments} shows the first three latent space axes of the prisms with $4-10$ sides aligned to the prism with 50 sides.
The latent manifolds for prisms $4-10$ trace a similar saddle shape as the rectangular car-type objects, repeatedly looping through the path for a number of times equal to their number of sides (e.g. the square prism completes four loops).
The 50-sided prism, with a side count becoming significant compared to the 500-point sampling density of $\soThreeSet_1$, does not show the same saddle shape in this computational setting but does pass repeatedly through similar loops corresponding to its 50 sides.
Fig.~\ref{fig:so2PrismAlignments} shows the latent manifolds both at a standardized common scale and at the original scales determined by the image space distances. As the number of sides increase, the polygonal faces approach a circle.
The single rotational axis here is orthogonal to these faces, so as the prisms become rounder, the images at different poses become progressively more alike and their distances decrease. 
The figure shows that the scales of the latent manifolds become correspondingly smaller as the number of sides increases. 
The prisms with 100 and 1000 sides (see the supplement) are nearly circular and produce $\soThreeSet_1$ latent manifolds that are essentially noise, while their scales continue to shrink.

Fig.~\ref{fig:so2PrismAlignments} also highlights the importance of the choice of registration method. Our shape distance minimizes RMSE limited to registrations that preserve parameter neighbors.
The standardized latent space manifolds for the prisms, modulo rotation, occupy similar regions in $\real^8$. They loop through these regions different numbers of times and at different rates though, and are in fact far away from each other in this shape space, with most distances around $[1.2,1.4]$.
To ease visual comparison, the figure shows alignments defined using a free registration that optimizes the linear sum assignment problem
using the SciPy package \cite{2020SciPy-NMeth},
similar to the Hungarian algorithm.
While this free registration is useful for visualization, it can distort the meaningful correspondence between point order and rotational pose, and produce low RMSE values for objects that actually have very different forms. As case in point, computing the RMSE using this alternative registration, the distances between the prisms drop to around $[0.3,0.4]$, while most of the values amongst the CAD objects remain similar to the shape distances in Fig.~\ref{fig:so2ShapeDistances}.

\subsection{Pose Submanifolds: Two Angles or $\torus^2$ Parameterizations}
\label{ssec:t2Data}

\noindent
The data in this section uses the two-dimensional subset $\soThreeSet_2$ of $SO(3)$ sampled from a copy of $\torus^2$ by taking the set of $(\theta,\phi)$ pairs with fixed $\theta=\pi/4$ and a circle of 40 evenly spaced $\phi$ values, and attaching to each of these a circle of 50 evenly spaced points in the $\psi$ coordinate, for a total of 2000 points in the set.
The 50-point circles at each $(\theta,\phi)$ pair can be thought of as leaning the object in the $\phi$ direction until it is halfway towards reaching the vertical position, then rotating it one full turn about its internal $z$-axis.
We map the generated image sets $\imageSpace^{\object}_{\soThreeSet_2}$ for the 13 \ac{CAD} objects and 11 prisms to an $8$-dimensional latent space using $\latentMap_{MDS}$.


\begin{figure*}[h]
  \begin{center}
    \includegraphics[width=0.59\textwidth, scale=1]{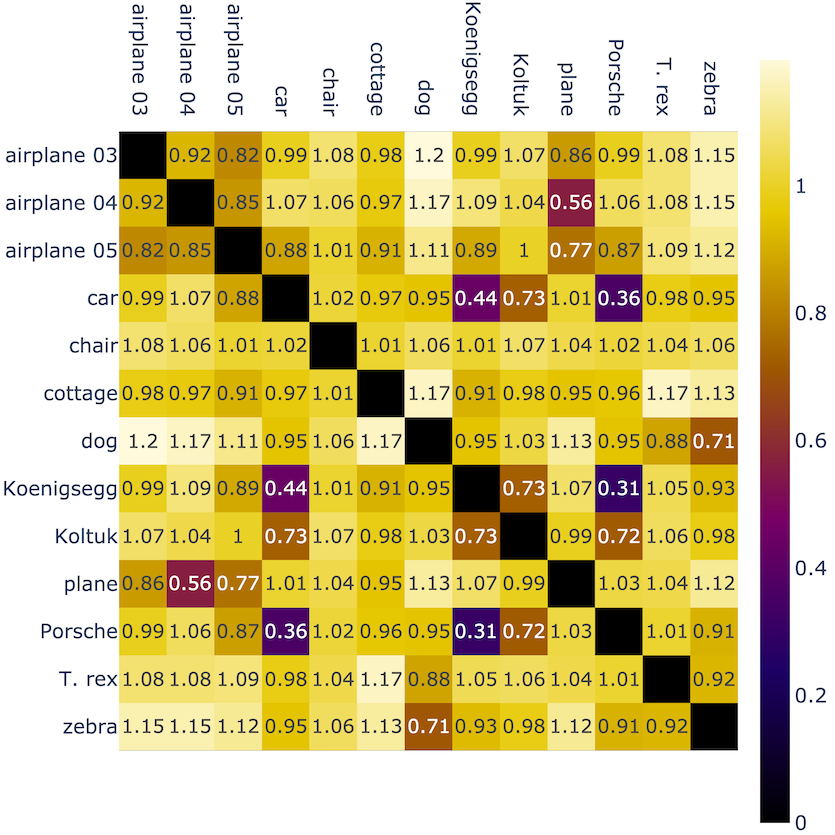}
    \includegraphics[width=0.39\textwidth, scale=1]{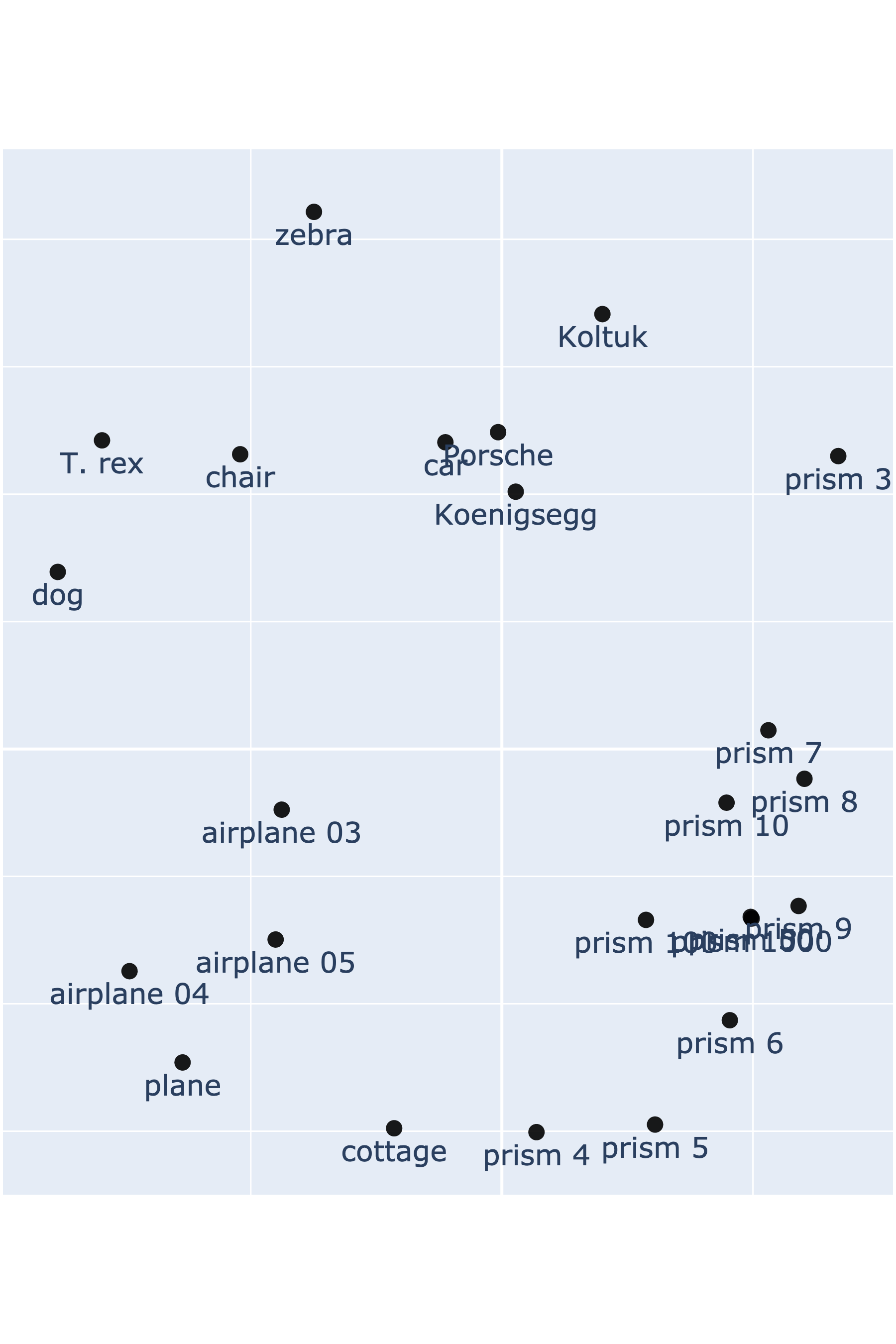}
    \caption{
      \emph{Left}: Latent space shape distances on the $\torus^2$ subset.
      \emph{Right}: Two-dimensional MDS positions produced by the distances.
      }
  \label{fig:t2ShapeDistances}
  \end{center}
\end{figure*}


The shape distances $d_{\soThreeSet_2}$ from Eqn.~\ref{eq:shapeDistance} and their two-dimensional proximity plot are shown in Fig.~\ref{fig:t2ShapeDistances}.
For visual clarity, we only include the subset of the distance matrix pertaining to the \ac{CAD} objects. A complete distance matrix is included in the supplement.
The distances between the prisms and the other objects are high, mostly in the range $[0.9,1.3]$. 
Their distances amongst themselves are smaller, with prisms with more sides being more closely grouped and having distances around $0.5$ or lower. 

The $\soThreeSet_2$ set forms the same car-type, animal-type, and airplane-type clusters as seen in section \ref{ssec:so2Data}.
The {\it Koltuk} displays poses in this set where its border is less similar to the cars, and it is farther away from their cluster here.
The {\it airplane 03} is somewhat more similar to the other airplanes than before.

\begin{figure*}
  \begin{center}
    \includegraphics[width=0.325\textwidth, scale=1]{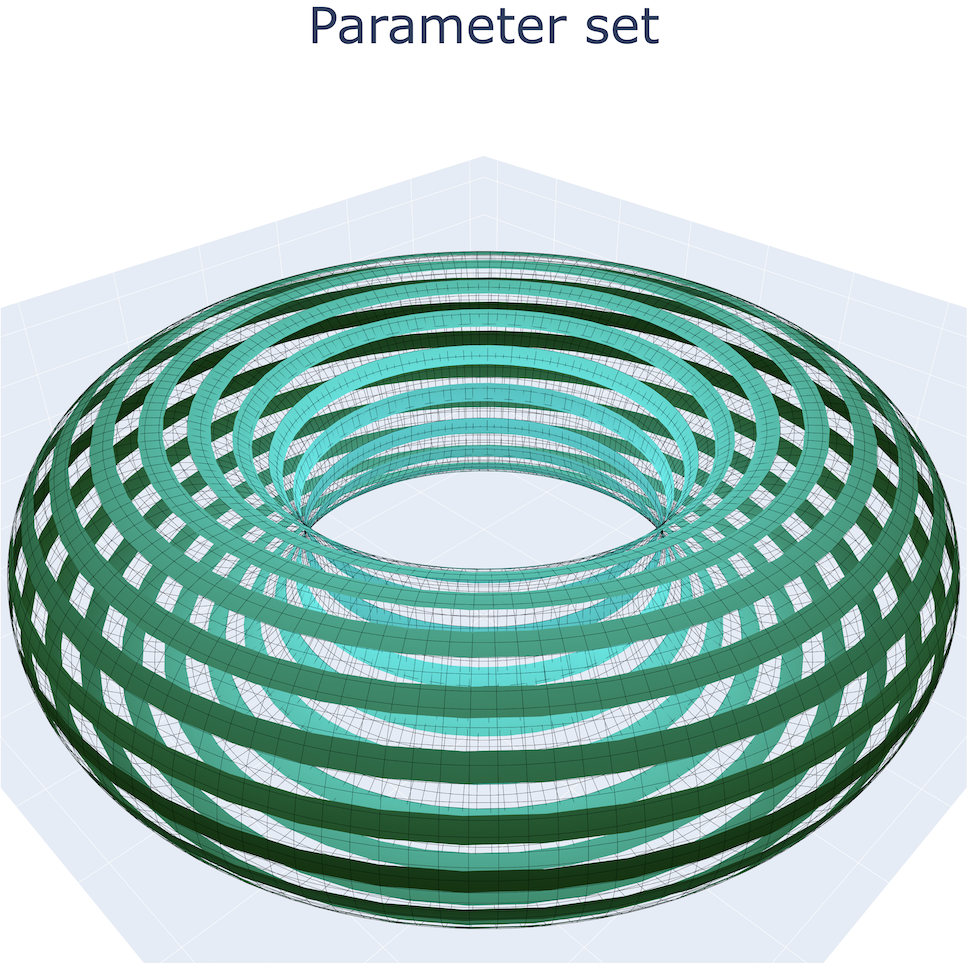}
    \includegraphics[width=0.325\textwidth, scale=1]{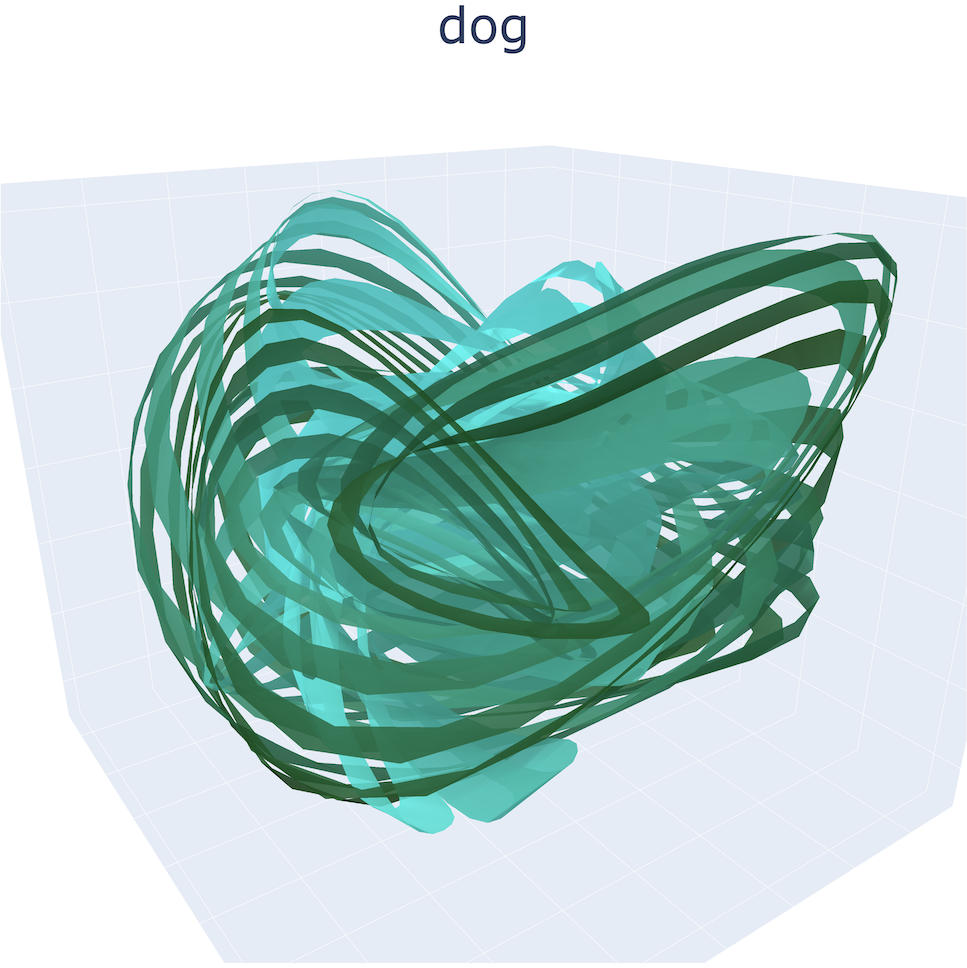}
    \includegraphics[width=0.325\textwidth, scale=1]{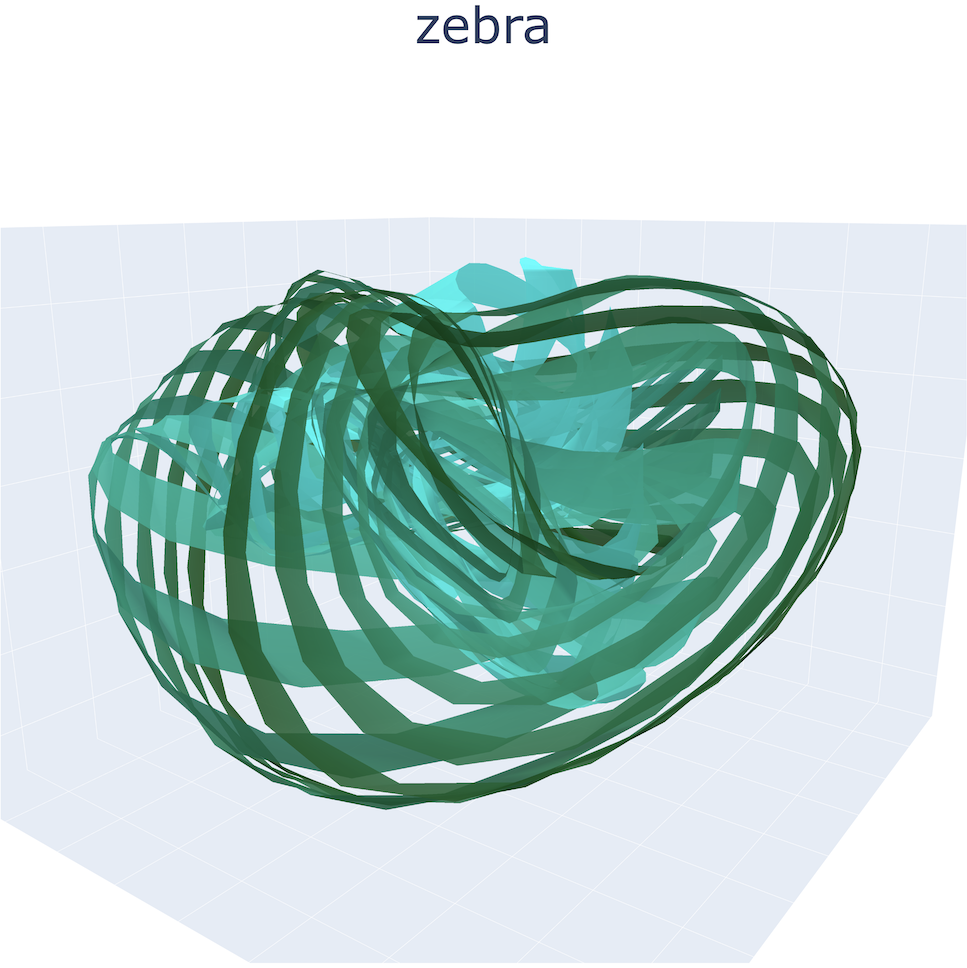}
    \includegraphics[width=0.325\textwidth, scale=1]{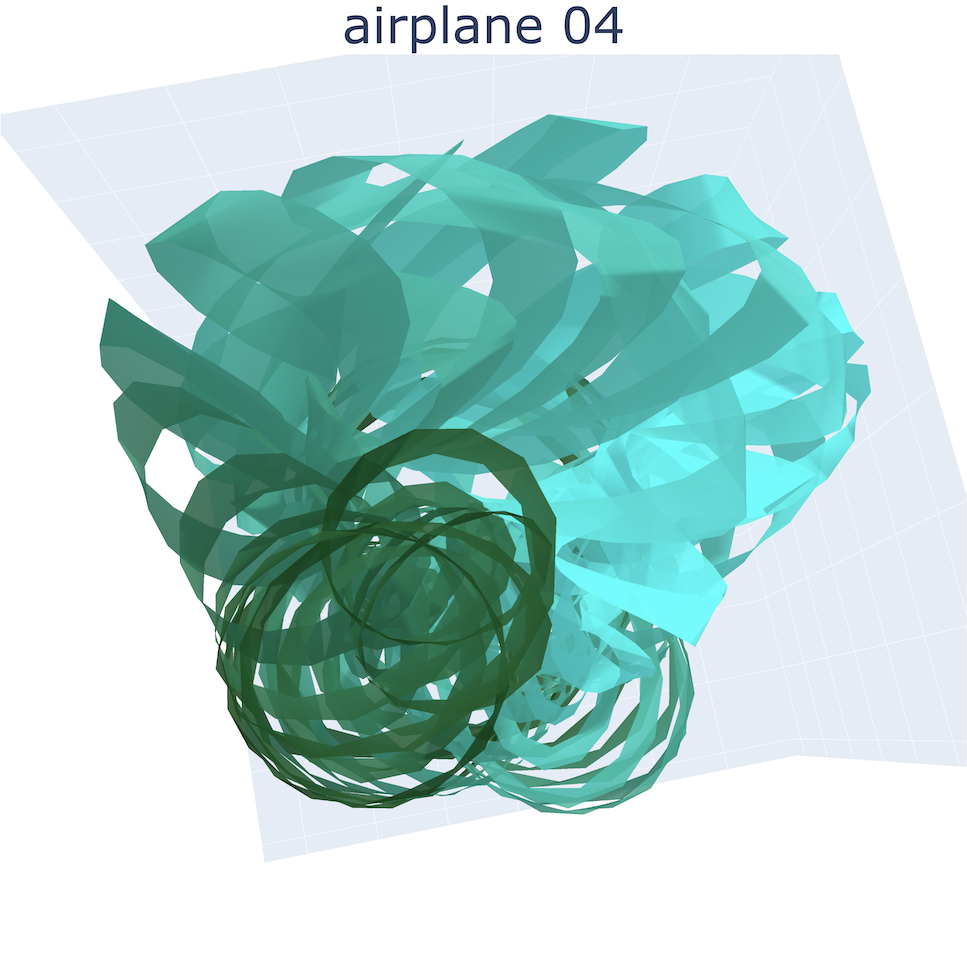}
    \includegraphics[width=0.325\textwidth, scale=1]{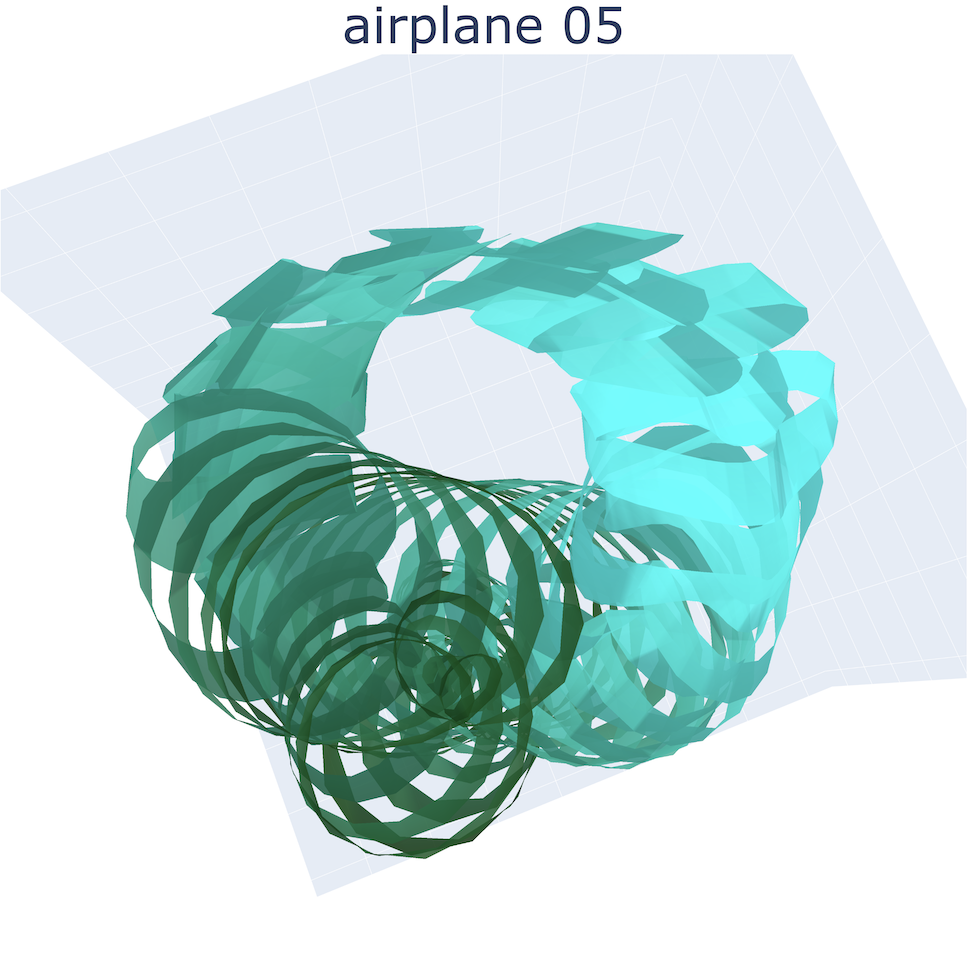}
    \includegraphics[width=0.325\textwidth, scale=1]{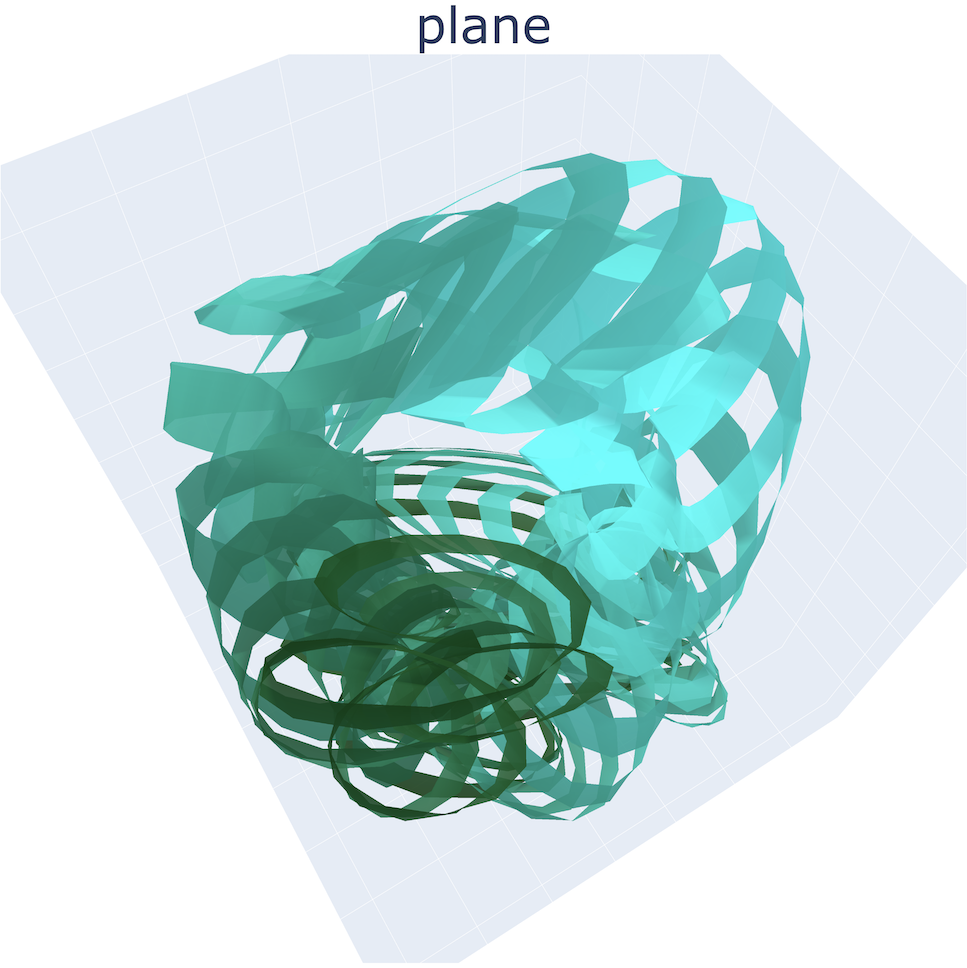}
    \includegraphics[width=0.325\textwidth, scale=1]{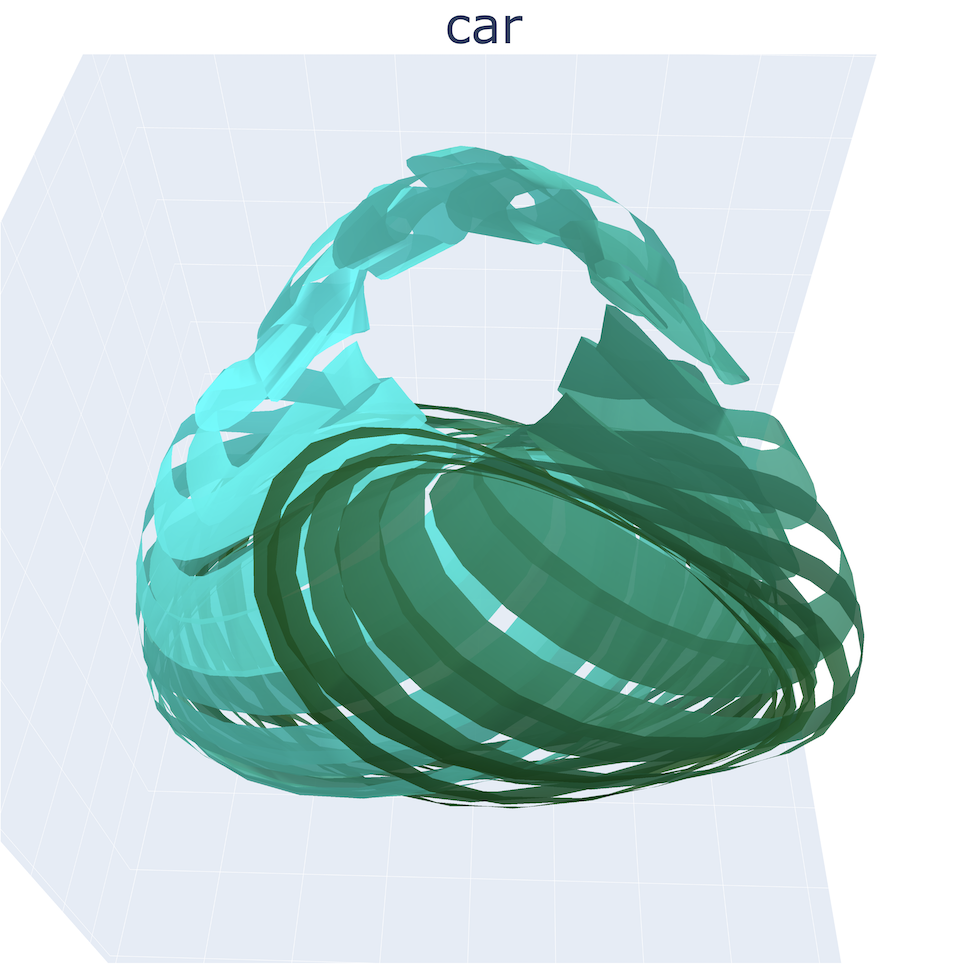}
    \includegraphics[width=0.325\textwidth, scale=1]{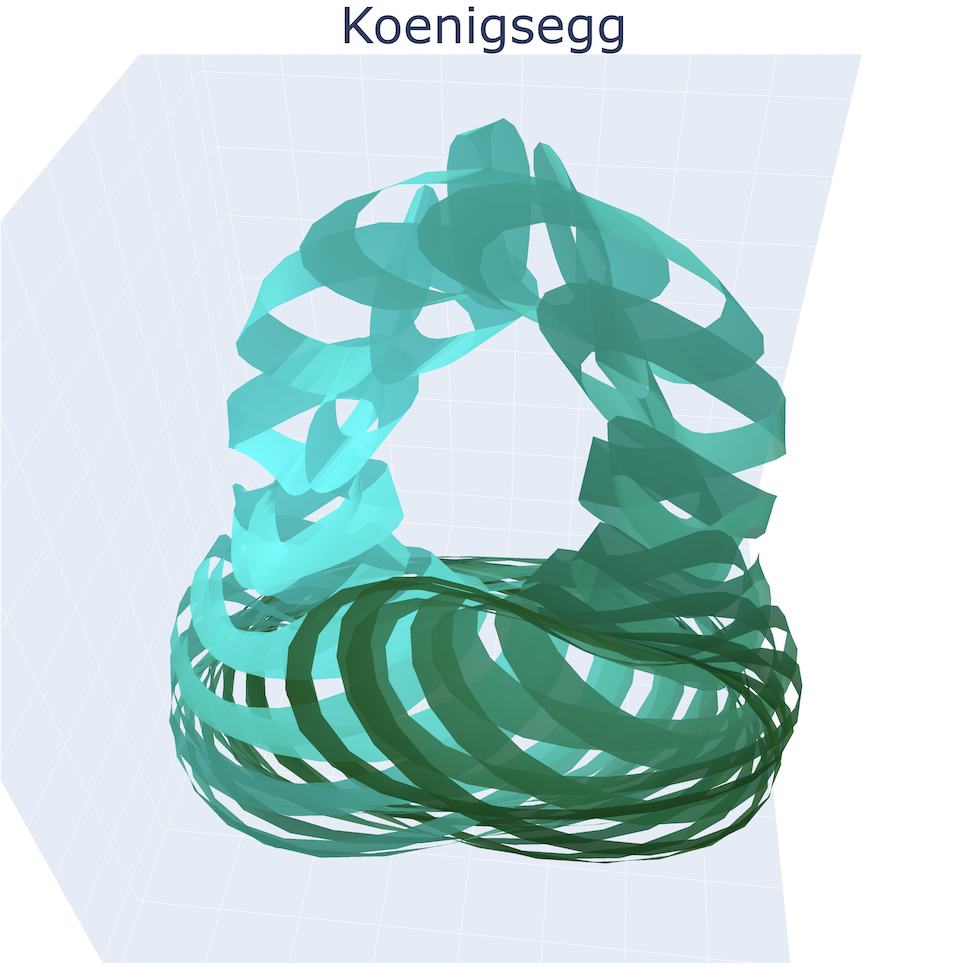}
    \includegraphics[width=0.325\textwidth, scale=1]{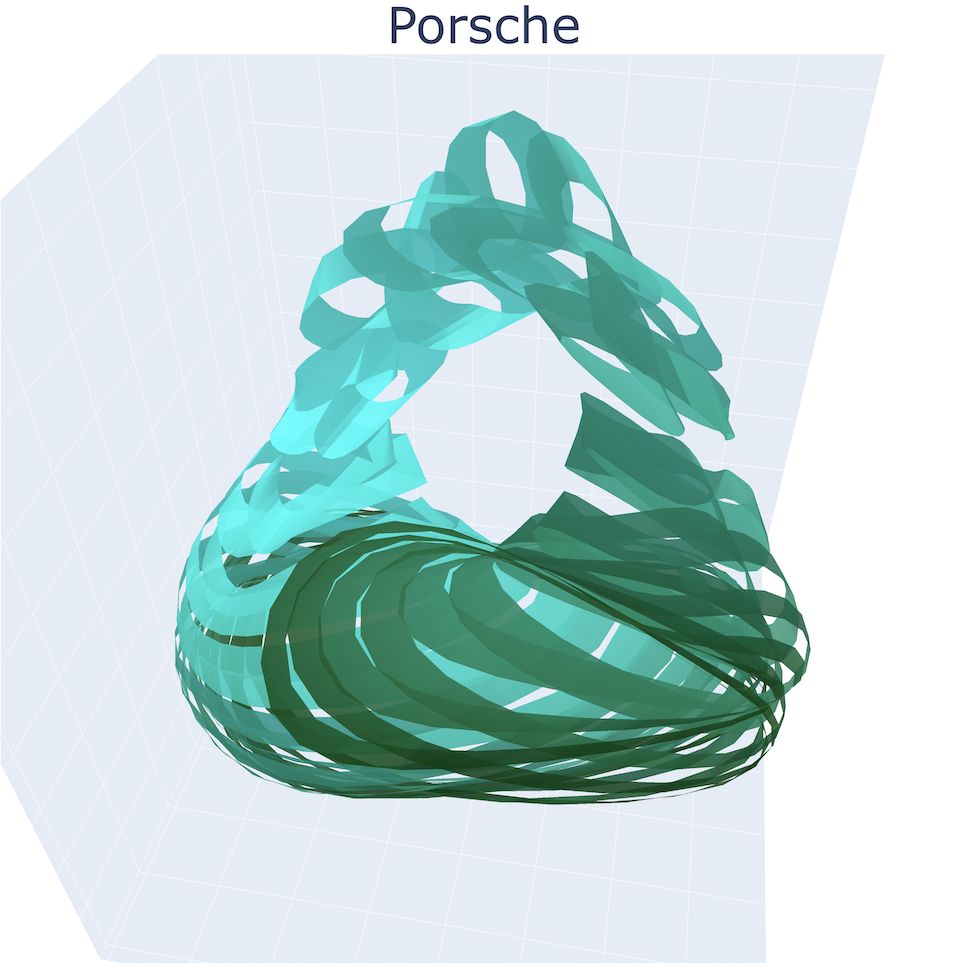}
    \caption{
      $\torus^2$ \ac{MDS} latent spaces for groups of similar objects.
      Each row displays a different group.
      Objects are aligned to the rightmost object in their row.
      Top left panel shows the parameter set used for all images in the figure.
      }
    \label{fig:t2MultiObjectCompare_1}
  \end{center}
\end{figure*}

Fig.~\ref{fig:t2MultiObjectCompare_1} visualizes the latent space manifolds for objects in these clusters.
The top-left panel shows how points correspond to the toroidal parameterization: position on the vertical circles of the torus indicates the $\phi$ coordinate, and position on the horizontal circles indicates the $\psi$ coordinate. The visualization decomposes the manifolds into bands forming loops in the $\psi$ coordinate, with the dark green band showing the same copy of $SO(2)$ used in section \ref{ssec:so2Data}.
The first three PCA axes are plotted. These generally capture about 50\% of the total variation for the $\torus^2$ latent spaces for each object, and the further spectral values are nearly equal.
We note that while there are self-intersections in the first three axes of the latent space, variation in the higher axes allow these to be embeddings in $\real^8$ rather than just immersions. 
The manifold shapes are visually distinct between clusters and similar within them.
It is visible that the airplane-type and car-type manifolds can be formed by looping structures in the $\psi$ coordinate, which together form another loop in the $\phi$ coordinate.
Closer inspection (see the supplement) shows that individual bands in the car-type manifolds form close double loops as in section \ref{ssec:so2Data}.

\begin{figure*}
  \begin{center}
  \includegraphics[width=0.244\textwidth, scale=1]{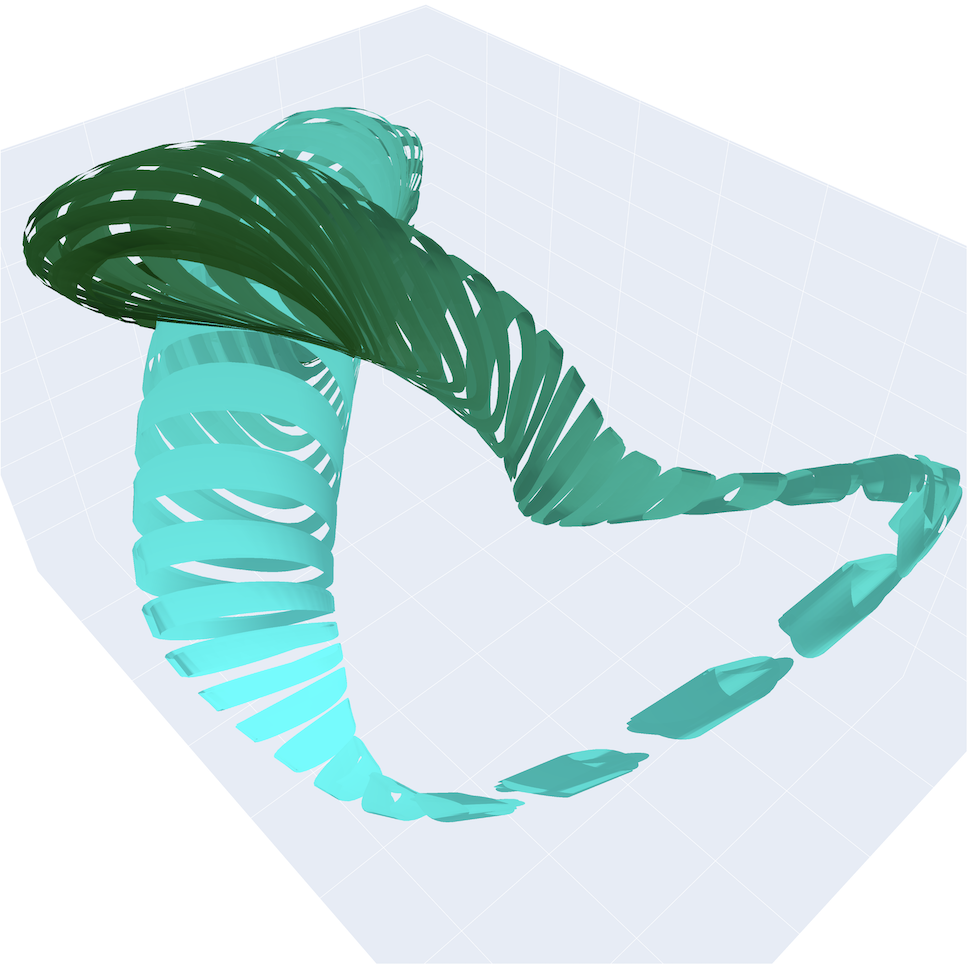}
  \includegraphics[width=0.244\textwidth, scale=1]{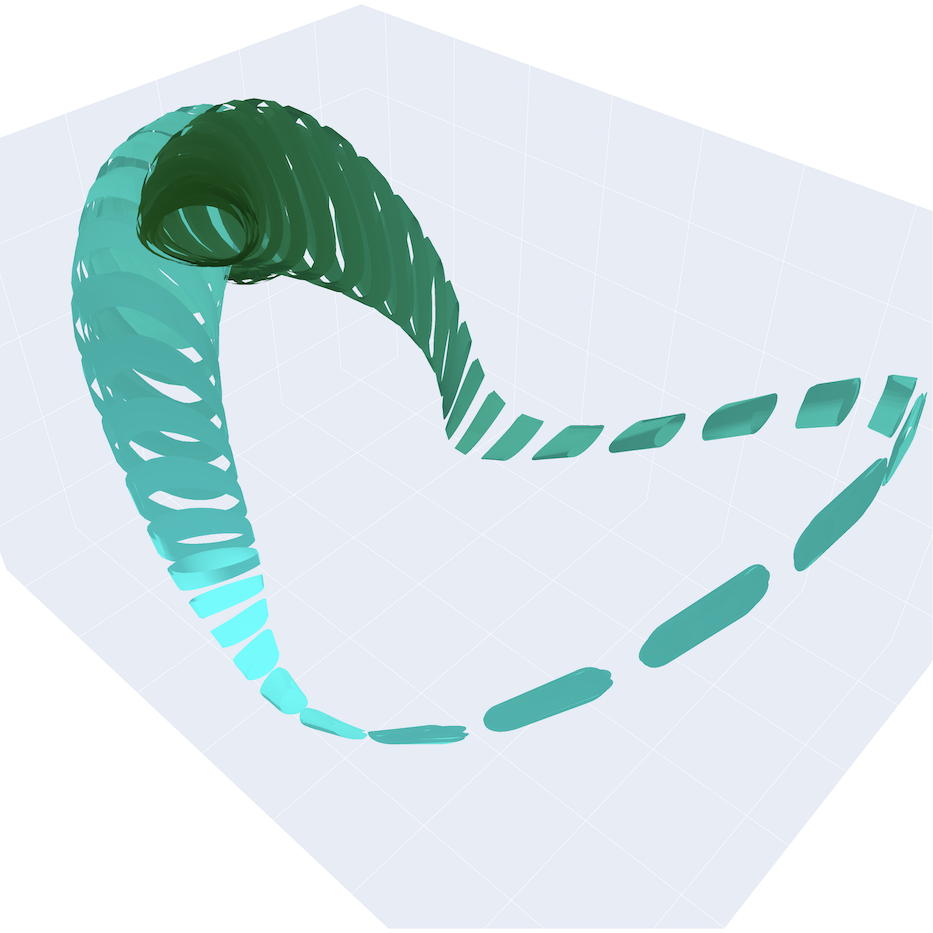}
  \includegraphics[width=0.244\textwidth, scale=1]{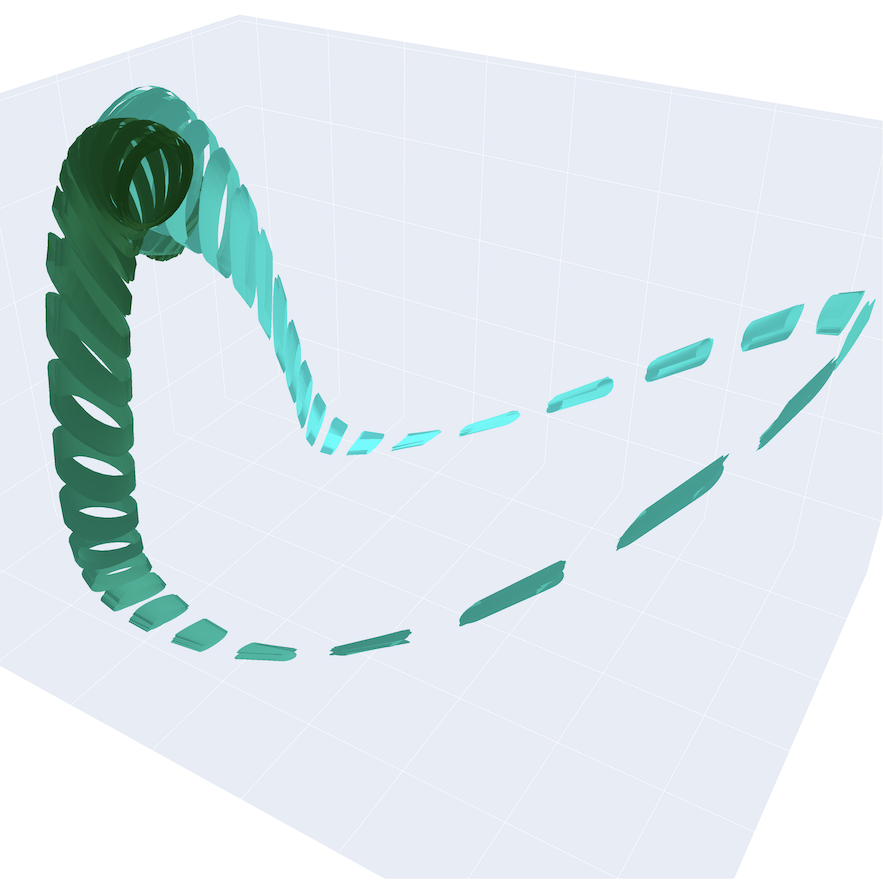}
  \includegraphics[width=0.244\textwidth, scale=1]{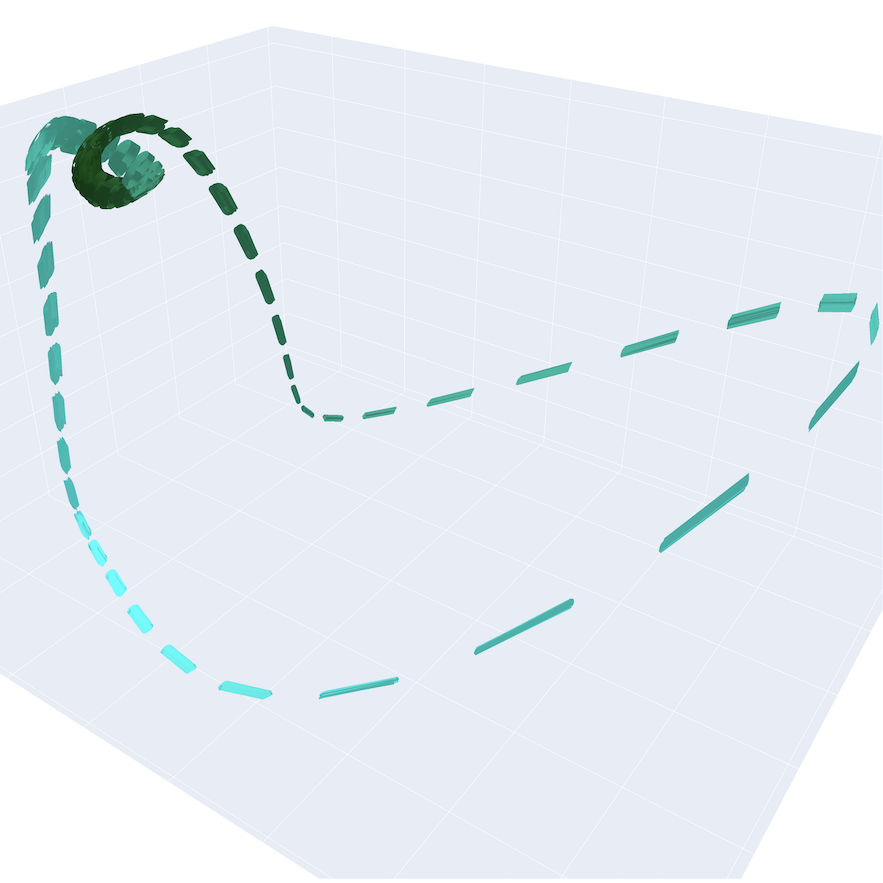}
  \caption{
    $\torus^2$ MDS latent spaces for the prisms. Prisms with 4, 5, 6, and 9 sides from left to right.
    }
  \label{fig:t2Prisms_1}
  \end{center}
\end{figure*}

Fig.~\ref{fig:t2Prisms_1} shows $\soThreeSet_2$ latent manifolds for a few of the prisms.
The manifolds are decomposed into loops in $\psi$ which together form a loop in $\phi$. As in section \ref{ssec:so2Data}, the $\psi$ loops are traced repeatedly according to the number of sides, and they become narrower as the number of sides increases. The number of sides is less relevant for changes in $\phi$, and the scale of the $\phi$ loop remains stable as the $\psi$ loops shrink. 
The $\phi$ scale becomes dominant, resulting in smaller shape distances and clustering of the higher-sided prisms.


{}


\subsection{Full Pose Manifold: $SO(3)$ Parameterizations} 
\label{ssec:completeSo3Data}

The data in this section uses a three-dimensional sampling $\soThreeSet_3$ of points throughout $SO(3)$ designed to give a balanced representation of the entire space. It uses 267 $(\theta,\phi)$ pairs generated over $\sphere^2$ using the Fibonacci spirals of Swinbank and Purser~\cite{swinbank2006fibonacci} (also studied in Hardin~et~al.~\cite{hardin2016comparison}).
A circle of 30 evenly spaced $\psi$ values
is attached to each of these pairs, giving a total of 8010 points distributed throughout $SO(3)$ in a way shown in Yershova~et~al.~\cite{yershova2010generating} to produce cells analogous to cubes.
We map the generated image sets $\imageSpace^{\object}_{\soThreeSet_3}$ for the 13 \ac{CAD} objects to a $32$-dimensional latent space using $\latentMap_{MDS}$.

\begin{figure*}[h]
  \begin{center}
    \includegraphics[width=0.59\textwidth, scale=1]{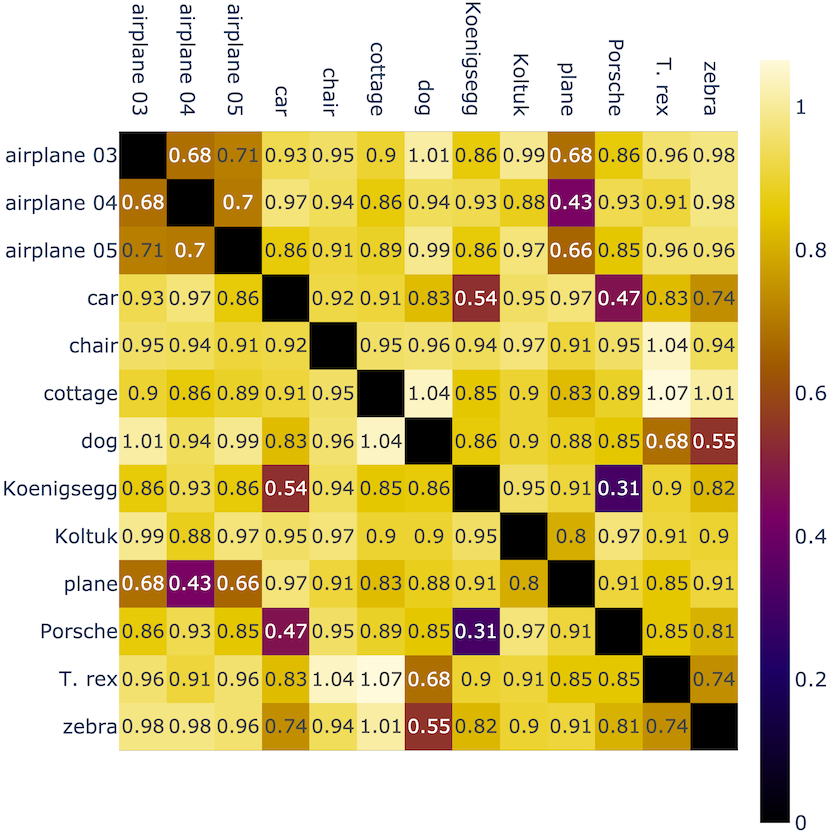}
    \includegraphics[width=0.39\textwidth, scale=1]{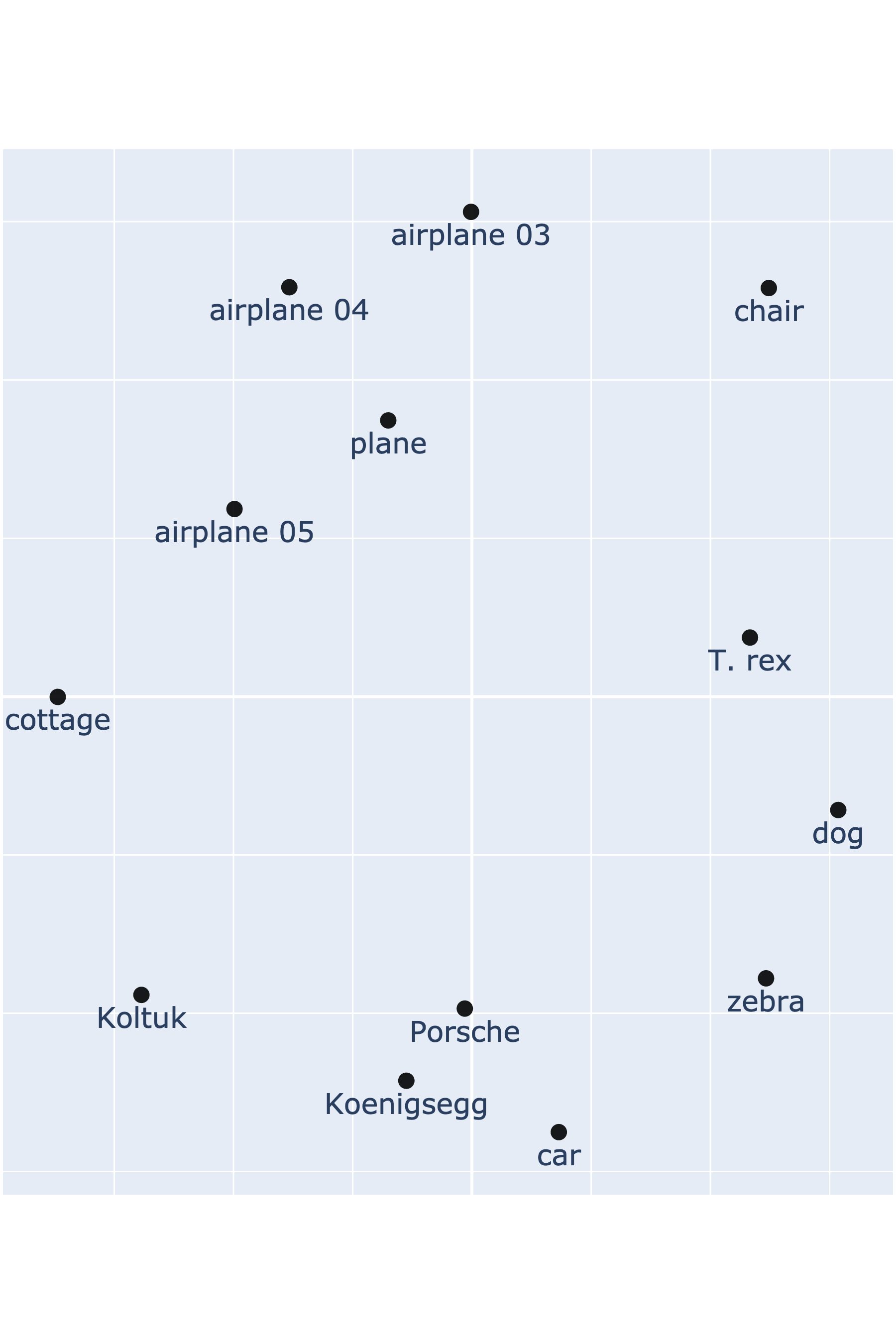}
    \caption{
      \emph{Left}: Latent space shape distances on the full-$SO(3)$ set.
      \emph{Right}: Two-dimensional MDS positions produced by the distances.
      }
  \label{fig:so3ShapeDistances}
  \end{center}
\end{figure*}


The shape distances $d_{\soThreeSet_3}$ from Eqn.~\ref{eq:shapeDistance} and their two-dimensional proximity plot are shown in Fig.~\ref{fig:so3ShapeDistances}.
For this set we rely on the pre-registration defined by the default positioning of the objects and do not perform further registration.
The $\soThreeSet_3$ set again forms car-type, animal-type, and airplane-type clusters as in sections~\ref{ssec:so2Data} and~\ref{ssec:t2Data}.
Notably, the airplanes form a tighter cluster that includes {\it airplane 03}, and the {\it T. rex} is closer to the {\it dog} and {\it zebra} than before.
With a fuller representation of all aspects of the objects, the parts of the {\it Koltuk} that are least similar to the cars are included, and it is no longer close to their cluster.

\subsection{Extension to Illumination Manifolds}
\label{ssec:changingLightSource}

The data in this section uses the lighting position set $\lightSet$ defined in section~\ref{ssec:data}, formed by placing the light source at 500 evenly spaced points along a circle centered directly above the origin.
We map the generated image sets $\imageSpace^{\object}_{\lightSet}$ for the 13 \ac{CAD} objects to an 8-dimensional latent space using $\latentMap_{MDS}$. 

\begin{figure*}
  \begin{center}
    \includegraphics[width=0.59\textwidth, scale=1]{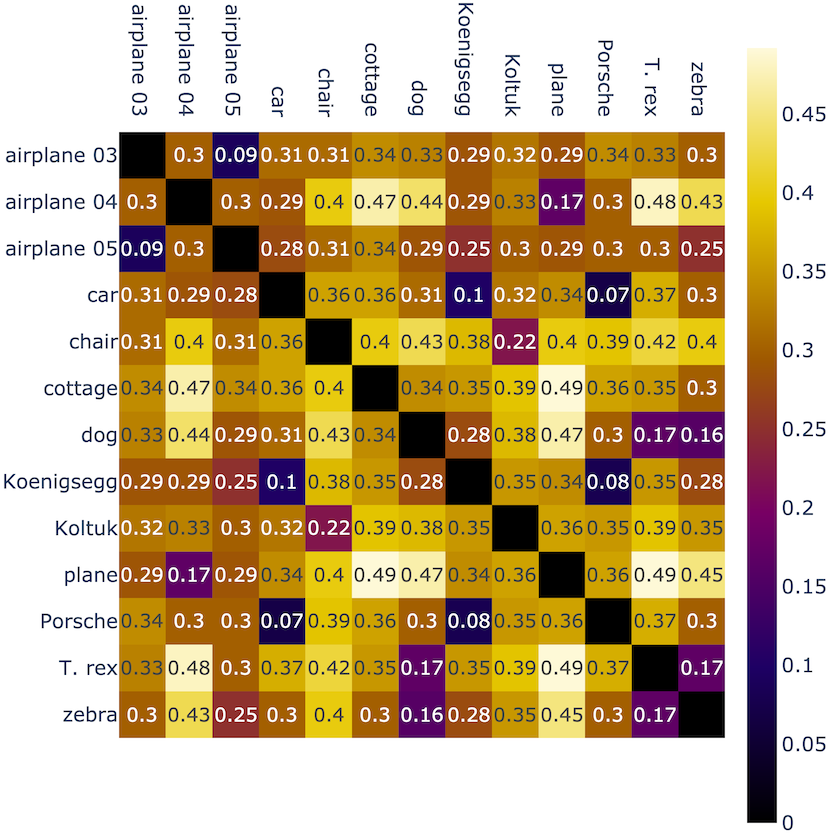}
    \includegraphics[width=0.39\textwidth, scale=1]{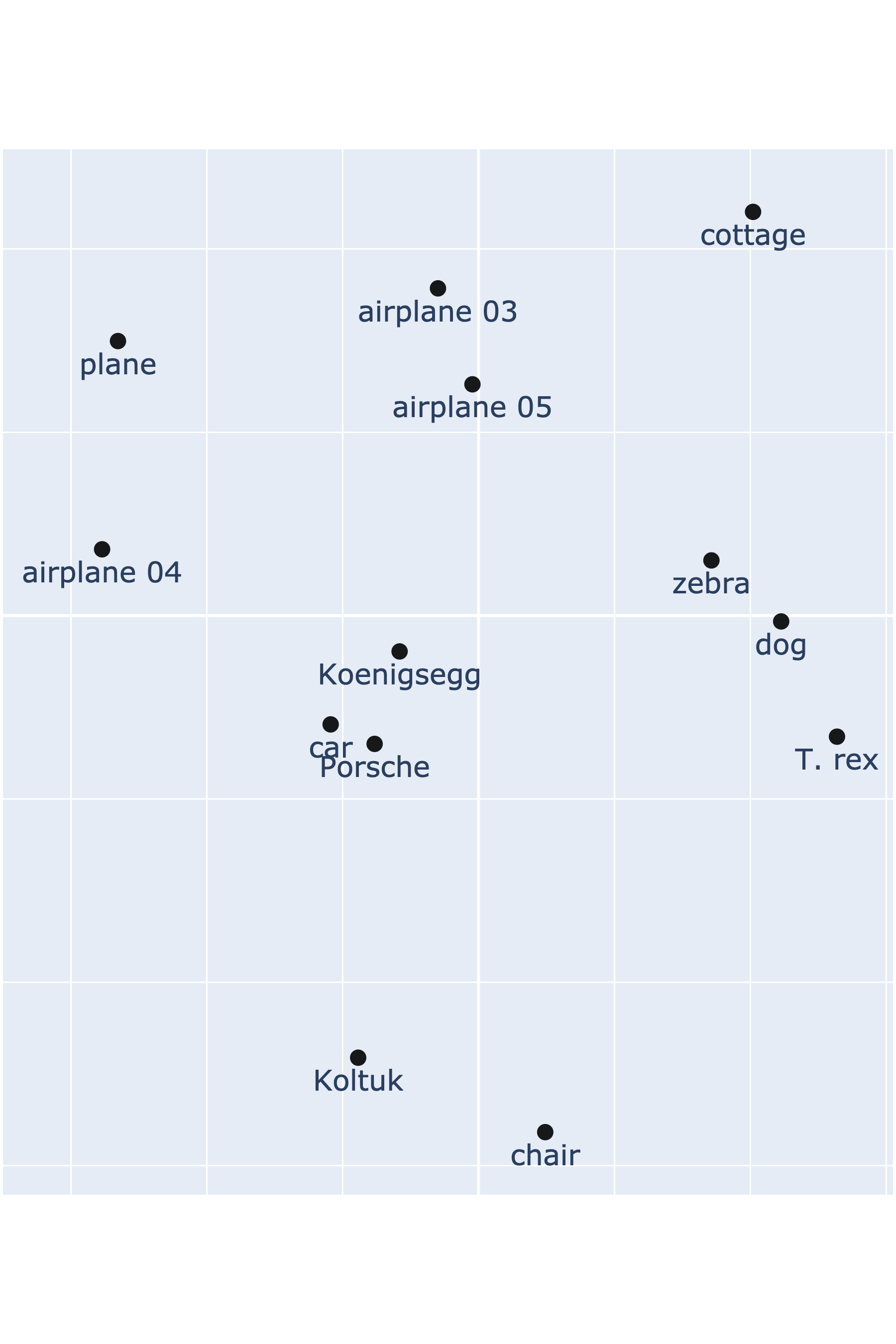}
    \caption{
      \emph{Left}: Latent space shape distances on the illumination set.
      \emph{Right}: Two-dimensional MDS positions produced by the distances.
      }
  \label{fig:lightPosShapeDistances}
  \end{center}
\end{figure*}

The shape distances $d_{\lightSet}$ from Eqn.~\ref{eq:shapeDistance} and their two-dimensional proximity plot are shown in Fig.~\ref{fig:lightPosShapeDistances}.
We continue to see distinct car-type, animal-type, and airplane-type clusters. The set $\lightSet$ also places the seat-type objects {\it chair} and {\it Koltuk} close to each other in shape space.

\begin{figure*}
    \centering
    \begin{tabular}{|ccc|}
        \hline
        \adjustbox{margin=1pt, trim={0} {0} {0} {0}, clip}{\includegraphics[width=0.25\textwidth]{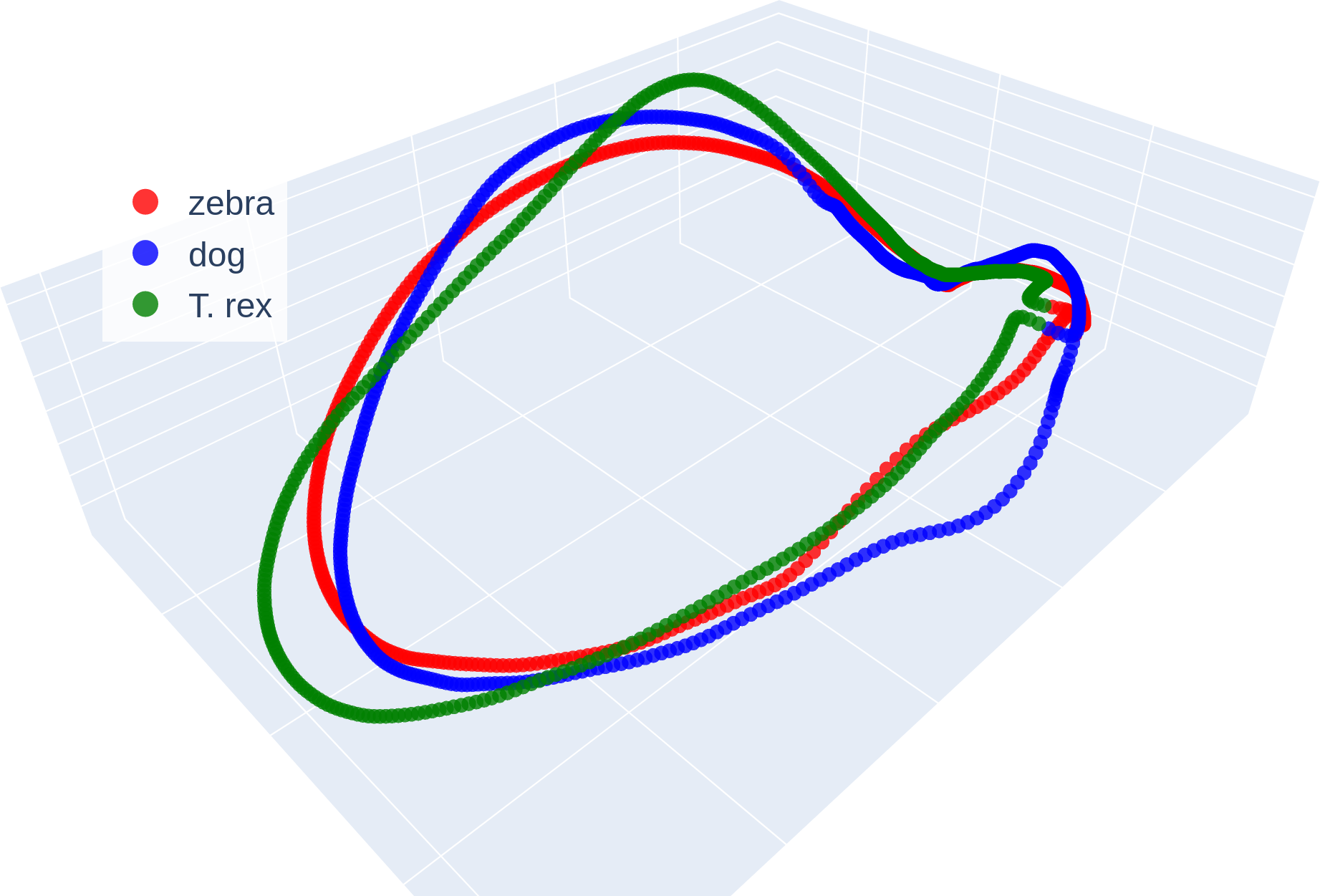}} &
        \adjustbox{margin=1pt, trim={0} {0} {0} {0}, clip}{\includegraphics[width=0.25\textwidth]{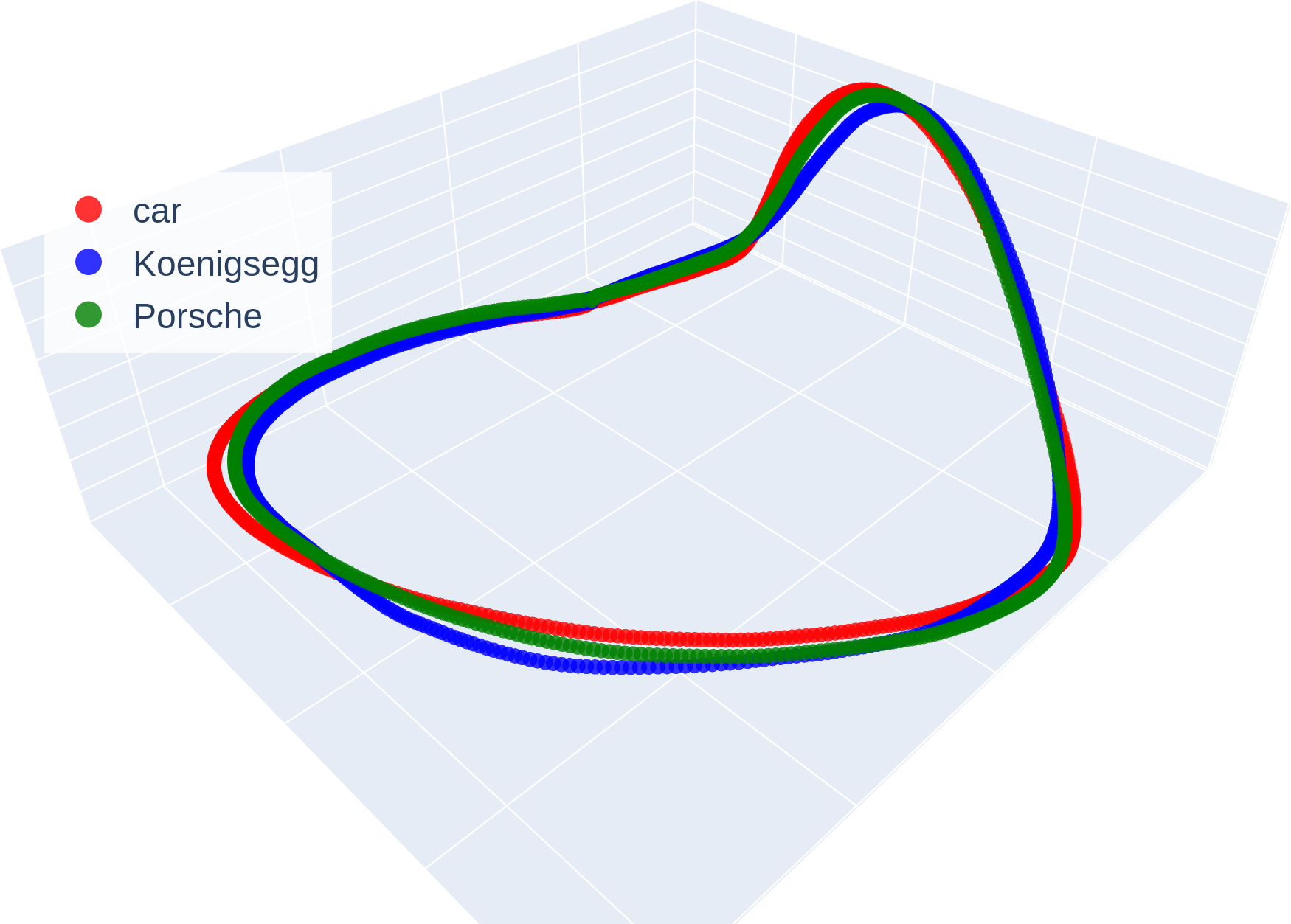}} &
        \adjustbox{margin=1pt, trim={0} {0} {0} {0}, clip}{\includegraphics[width=0.2\textwidth]{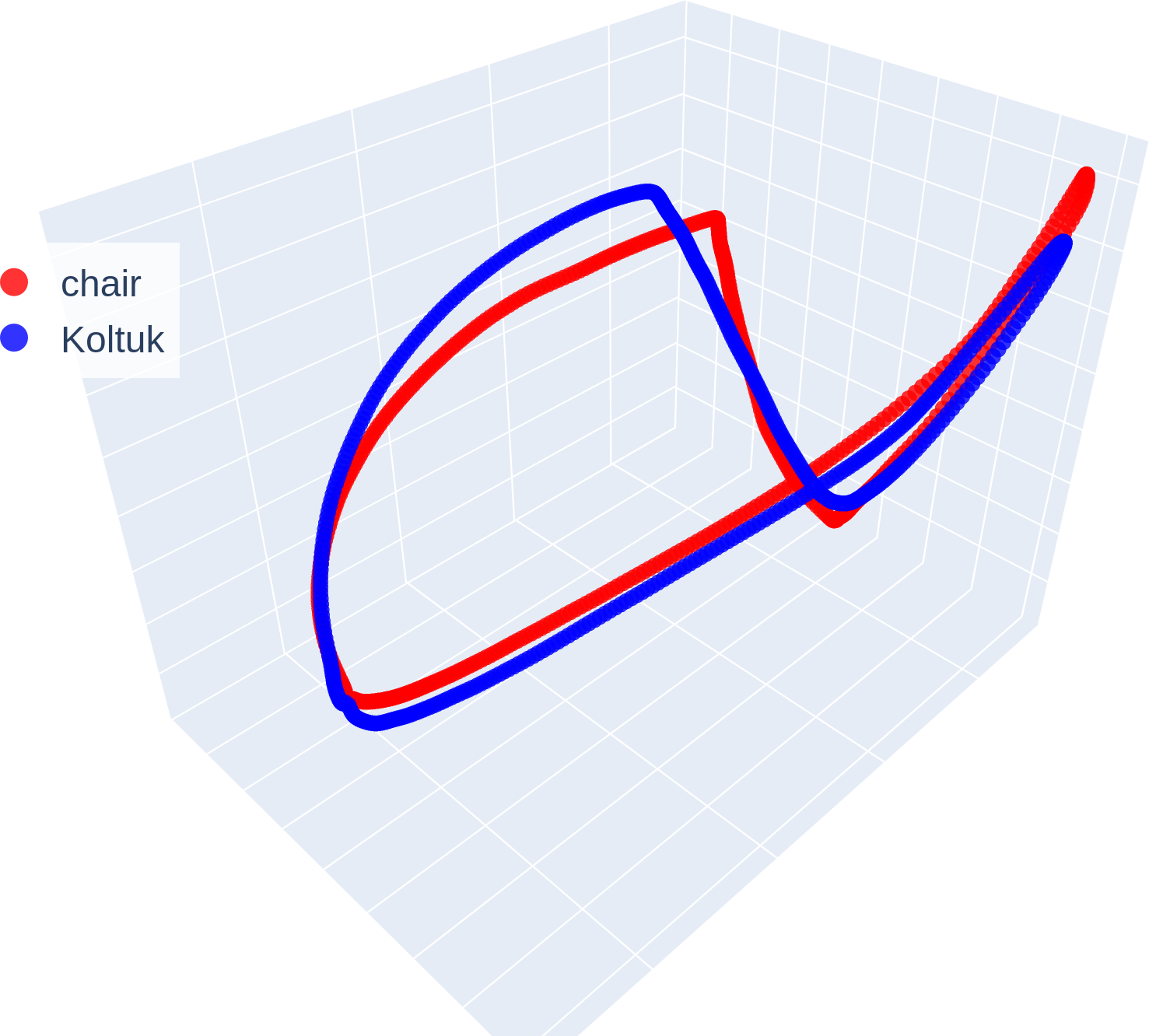}} 
        \vspace{-0pt} \\
        Animal-type & Car-type & Seat-type \\
        \hline
        \adjustbox{margin=1pt, trim={0} {0} {0} {0}, clip}{\includegraphics[width=0.25\textwidth]{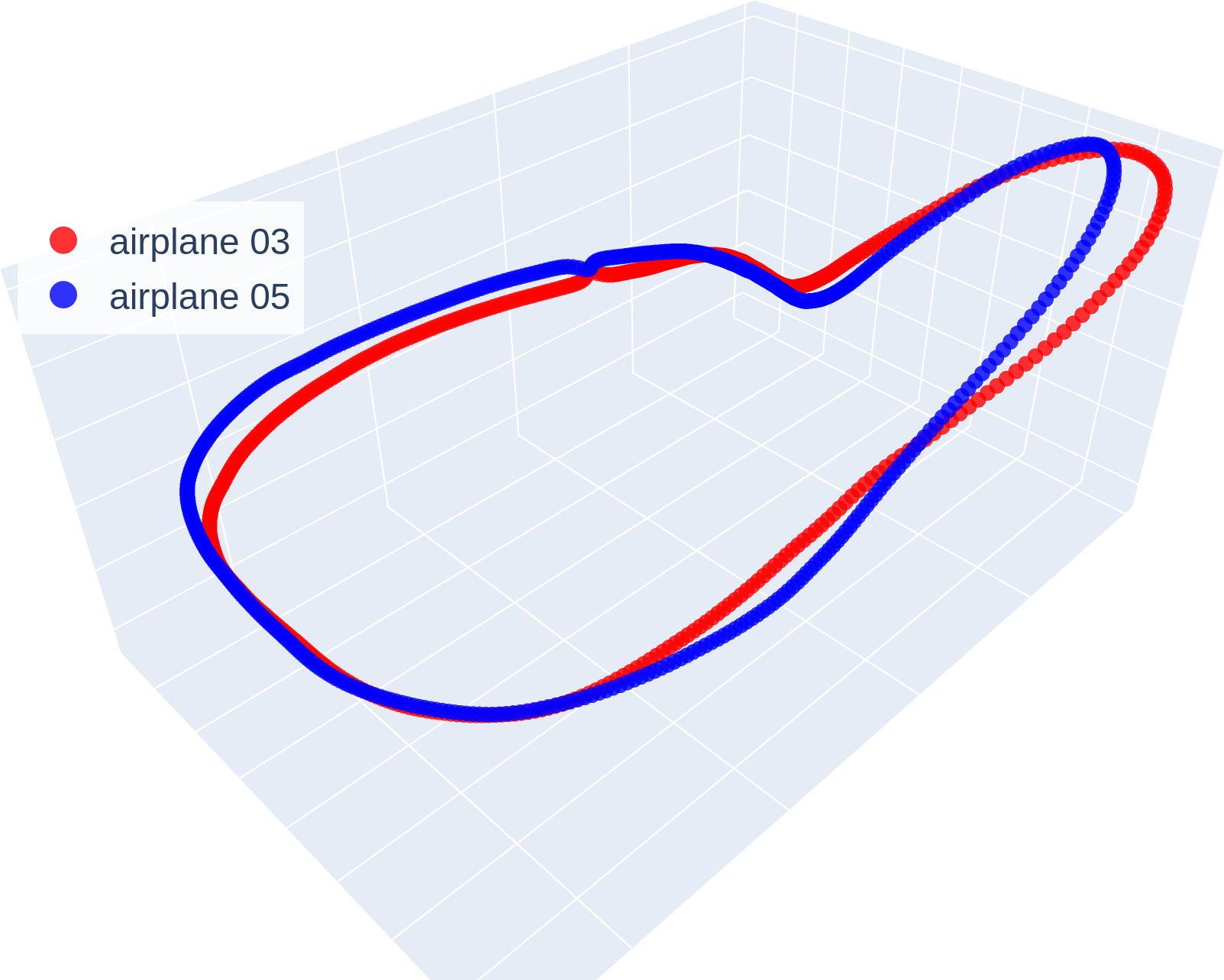}} &
        \adjustbox{margin=1pt, trim={0} {0} {0} {0}, clip}{\includegraphics[width=0.25\textwidth]{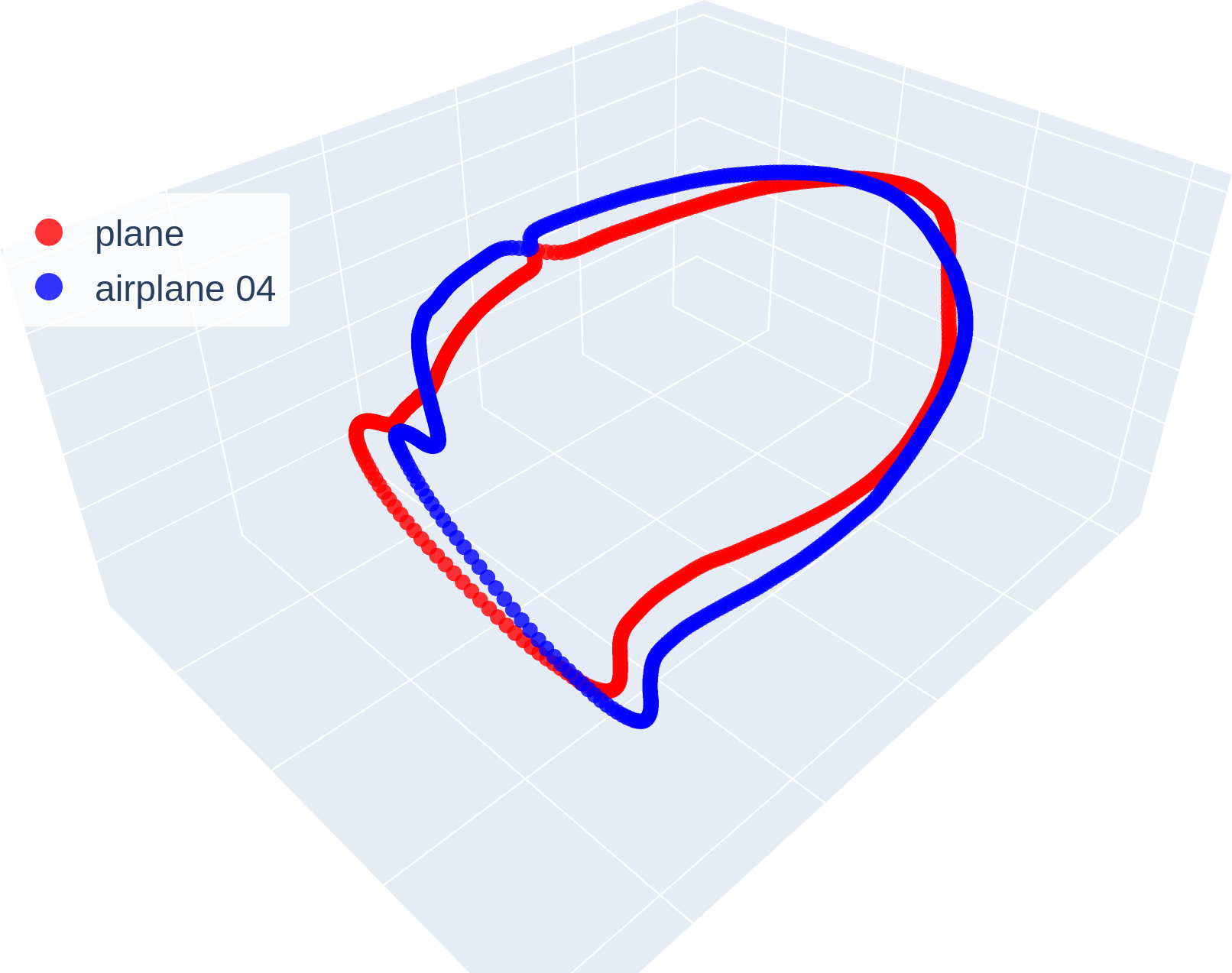}} &
        \adjustbox{margin=1pt, trim={0} {0} {0} {0}, clip}{\includegraphics[width=0.25\textwidth]{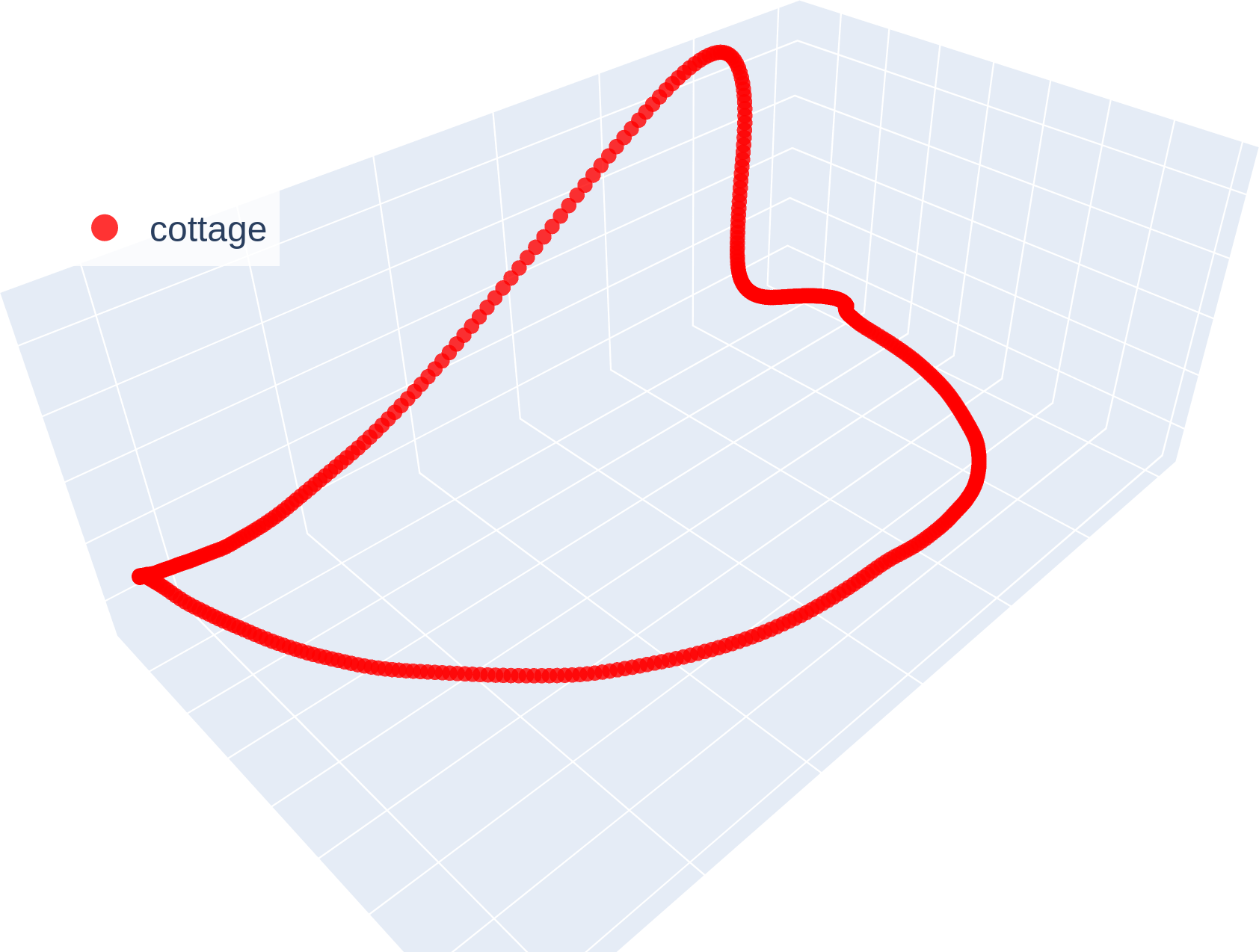}} 
        \vspace{-0pt} \\
        Airplane-type & Airplane-type & {\it cottage} \\
        \hline
    \end{tabular}
    \caption{
        Illumination MDS latent spaces for groups of similar objects.
    }
    \label{fig:light_clusters}
\end{figure*}

Fig.~\ref{fig:light_clusters} visualizes the latent manifolds for each cluster, demonstrating that objects within the same semantic category have similar latent space representations under these illumination conditions.
Interestingly, the illumination manifolds for some objects have saddle shapes similar to those seen for the pose manifolds. 
Further studies are needed to distinguish the shapes of illumination manifolds across objects. 


\section{Discussion and Future Directions}
\label{sec:discussion}

Our analysis thus far has focused on understanding the shape of an individual object manifold. However, several important questions arise when considering the practical applications and theoretical properties of manifold learning methods. In this section, we investigate two key aspects: how the presence of multiple objects affects the learned manifold structure of an individual object, and the stability of \ac{MDS} embeddings under different initializations and permutations. These investigations provide insights for future research on manifold shape analysis.

\subsection{Joint Multi-Object Manifold Learning}
\label{ssec:jointObjectMaps}


\begin{figure*}[h]
  \centering
  \begin{tabular}{cccc}
    \includegraphics[width=0.23\textwidth]{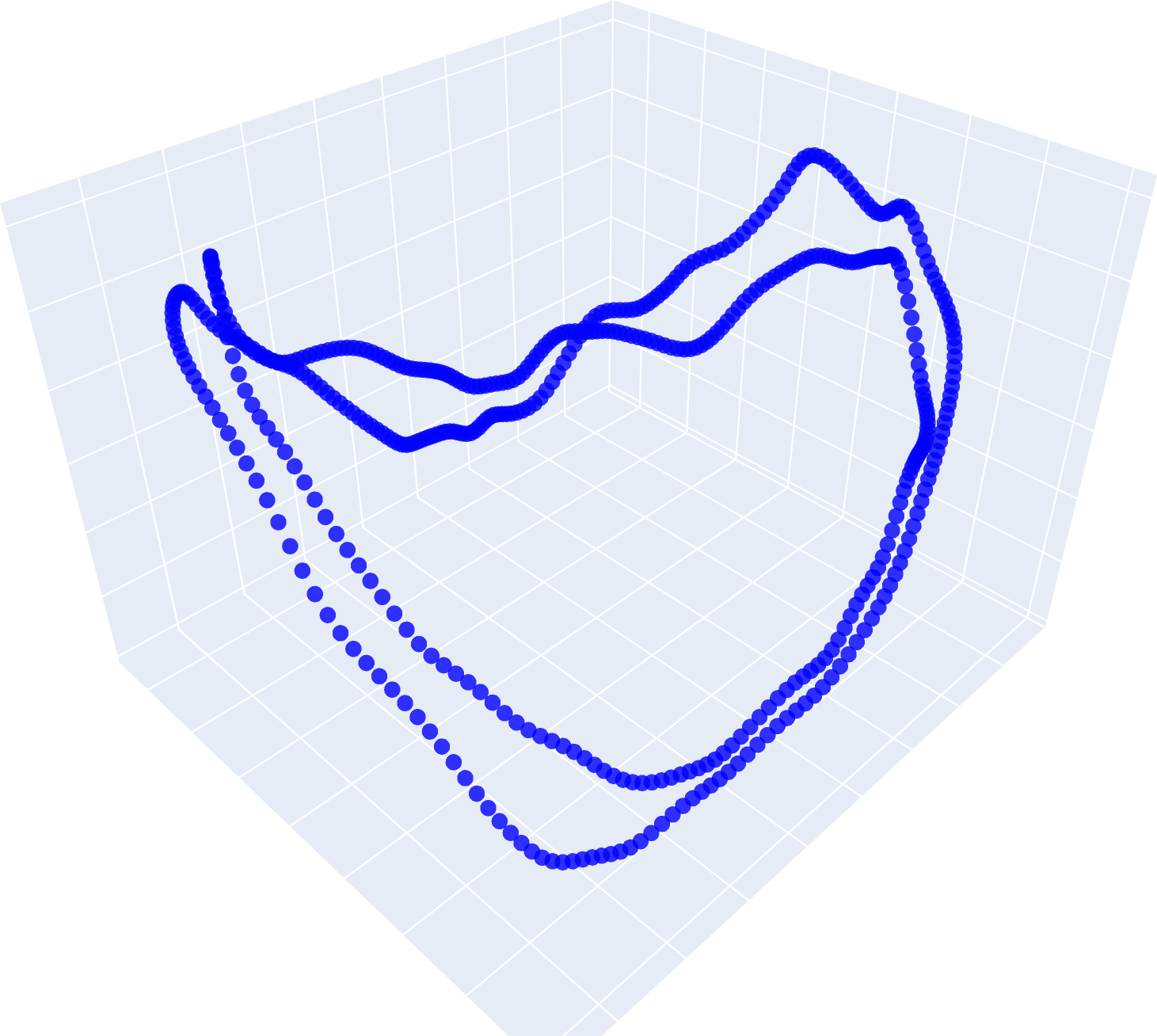}
    & \includegraphics[width=0.23\textwidth]{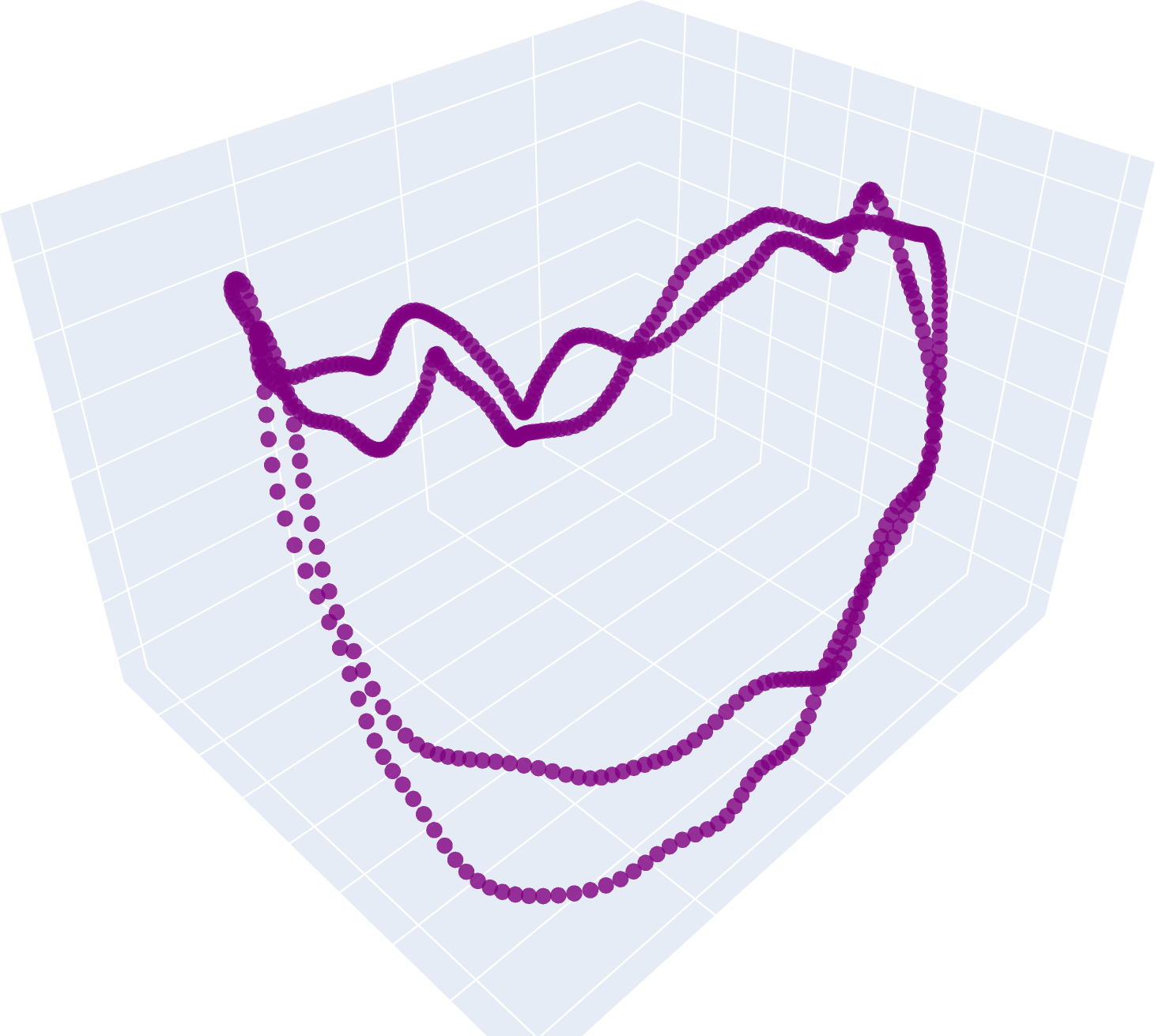}
    & \includegraphics[width=0.23\textwidth]{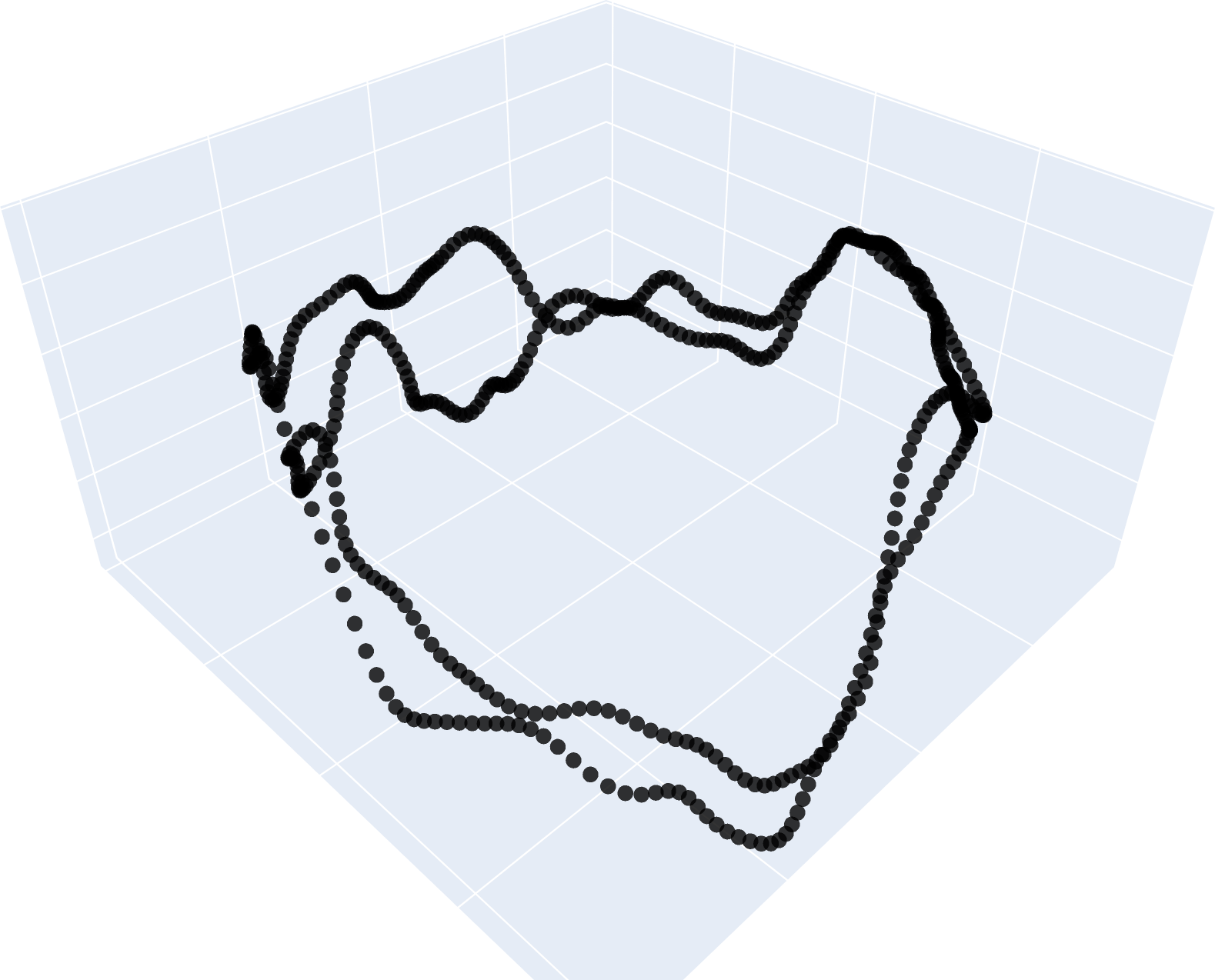} 
    & \includegraphics[width=0.23\textwidth]{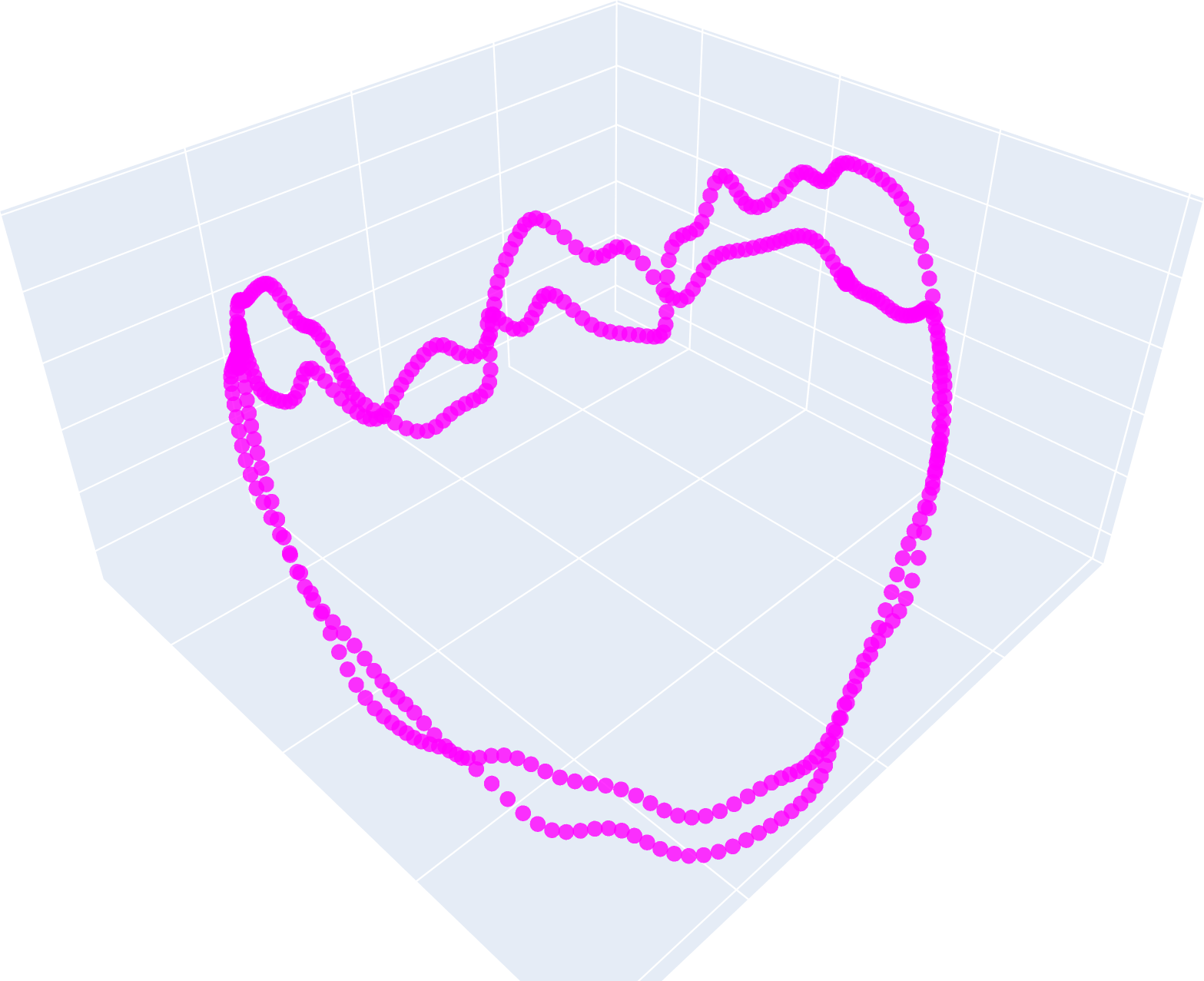}
    \\
    car 
    & car + Koenigsegg 
    & car + zebra 
    & car + Koenigsegg + zebra
  \end{tabular}
  \caption{
    Latent spaces for the {\it car} using individual MDS and joint embeddings with other objects included.
    }
  \label{fig:multi-mds}
\end{figure*}


We are interested in creating a map $\latentMap:\real^{\imageDimension\times\imageDimension}\to\real^\latentDimension$ that can read arbitrary points $\imagePoint$ that could belong to any number of objects, with the goal of identifying or classifying the object in the image.
A training set for such a map would include many different objects. We want to know how the shape of the latent pose manifold for a single object might be affected by the presence of other objects in the training set.
Here, we perform a brief investigation by viewing the impact of including multiple objects with varying degrees of similarity when creating an MDS map $\latentMap_{MDS}$.
Specifically, we compare the pose manifolds on the $SO(2)$ set $\soThreeSet_1$ from section~\ref{ssec:so2Data} for the object $\object_0$ = {\it car} generated by MDS when only $\object_0$ is included in the image set, versus when the similar object $\object_1$ = {\it Koenigsegg} and/or the dissimilar $\object_2$ = {\it zebra} are also included.
This creates the maps 
$\latentMap_{MDS,\soThreeSet_1}^{\object_0}$, 
$\latentMap_{MDS,\soThreeSet_1}^{\object_0+\object_1}$, 
$\latentMap_{MDS,\soThreeSet_1}^{\object_0+\object_2}$, and 
$\latentMap_{MDS,\soThreeSet_1}^{\object_0+\object_1+\object_2}$.
Separating the subsets of the embeddings corresponding to the {\it car} gives different latent pose manifolds. 
We compare these visually in Fig.~\ref{fig:multi-mds}, and see that the shape is generally preserved.
We leave further study of this approach for future work.

\subsection{Statstical Analysis of Image Manifolds}
\label{ssec:statAnalysis}

How can one exploit the shapes of these manifolds to accomplish computer vision tasks? 
An essential requirement for all of these tasks is that the map $\latentMap$ be invertible.
This allows us to perform any desired tasks in latent space, and evaluate the results in the image space.

\begin{figure} 
    \begin{center}
        \includegraphics[width=0.47\textwidth, scale=1]{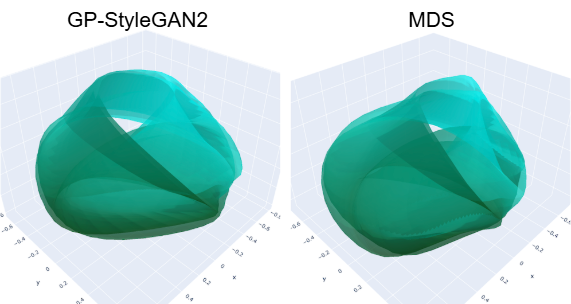}
    \end{center}
    \caption{
        Latent spaces for the {\it Koltuk} on a $\torus^2$ set using GP-StyleGAN2 and MDS.
        }
    \label{fig:gpMdsCompare}
\end{figure}

We have performed our analysis using MDS, which does not allow the kind of invertibility we need.
However, one can use a method such as GP-StyleGAN2~\cite{liang2024learning} to generate similar latent spaces, while having the additional functionality that comes with an invertible map.
Fig.~\ref{fig:gpMdsCompare} compares the latent spaces of MDS and GP-StyleGAN2. The result indicates that both methods produce visually similar manifolds.
This suggests that the shape analysis framework presented in this paper can be applied using GP-StyleGAN2, enabling new approaches to the design of computer vision applications.
For example, GP-StyleGAN2 has been used to create realistic image interpolations, perform image denoising, and compute mean poses from sample image sets.
The introduction of latent space shape distances might be used to improve object classification methods.


\subsection{MDS Invariance}
\label{ssec:mdsInvariance}


\begin{figure*}[h]
  \begin{center}
    \includegraphics[width=0.45\textwidth, scale=1]{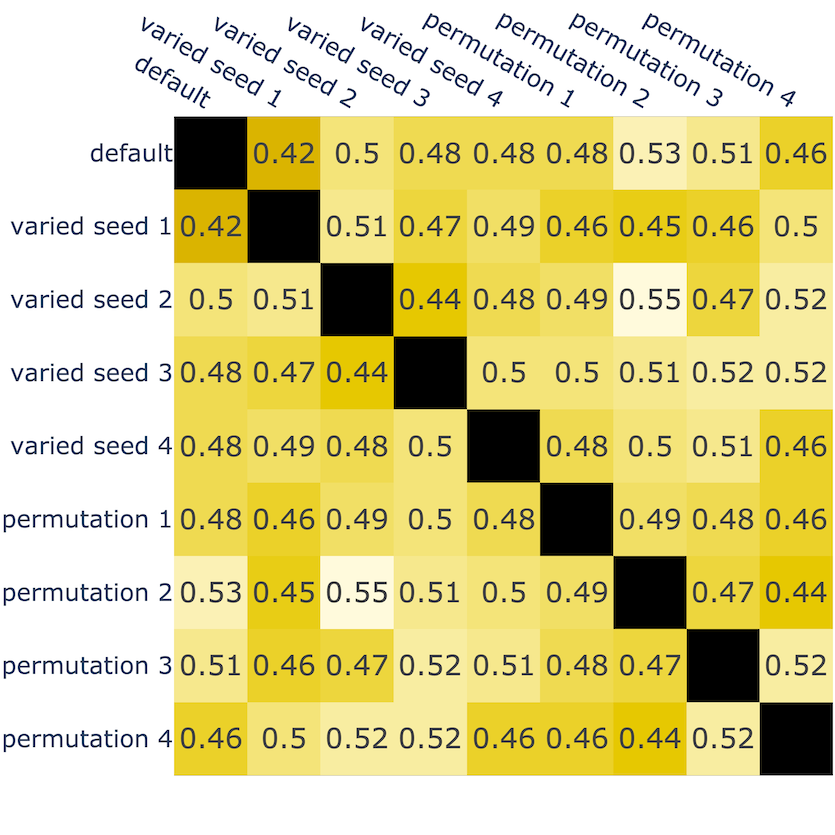}
    \includegraphics[width=0.40\textwidth, scale=1]{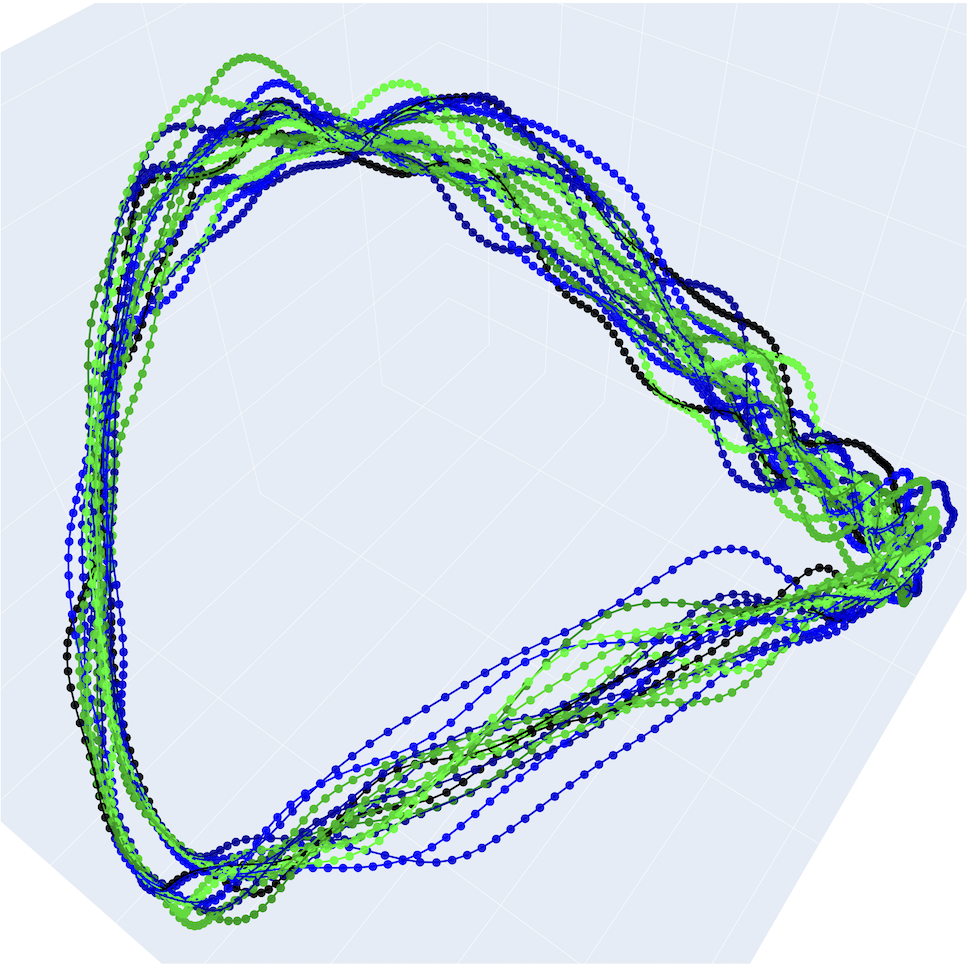}
    \includegraphics[width=0.13\textwidth, scale=1]{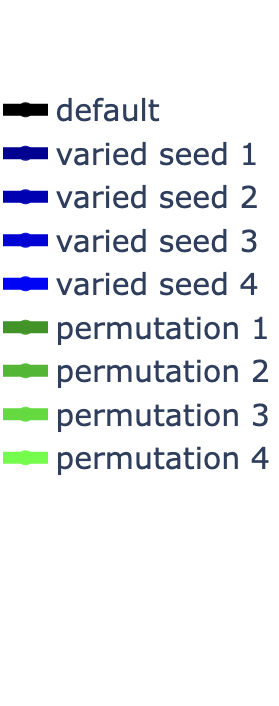}
    \caption{
      MDS embeddings of image distances from the {\it car} $SO(2)$ pose set with varied initializations and distance matrix permutations. 
      \emph{Left}: Latent space shape distances.
      \emph{Right}: First three PCA axes after alignment to the default embedding.
      }
  \label{fig:mdsConsistencyPermutations}
  \end{center}
\end{figure*}


Ideally, a map $\latentMap^{\object}$ would be uniquely determined by the distances between the input points. If viewed as a point cloud, changing the order of the points does not change the geometry of the set and so should not change the embedding positions using a geometry-preserving map.
However, this is not the case in practice.
In fact, the vagaries of algorithmic search allow not just a single unique latent manifold but instead a large set of possible latent manifolds which are similar but distinct.
We demonstrate this with a simple experiment.
Here we use MDS to embed a single set of distances but with varying details that should theoretically be irrelevant. 
We use the image space distances for the {\it car} on the one-dimensional set $\soThreeSet_1$ from section \ref{ssec:so2Data}.
We first set a random seed for the initialization and perform a default embedding to $8$ dimensions. Next we set four different random seeds for initialization and perform embeddings for each. Then we perform four random permutations of the distance matrix, equivalent to changing the point order, embed each of these using the original random seed, and perform the inverse permutations to return the latent point sets to the original order.

Fig.~\ref{fig:mdsConsistencyPermutations} shows the shape distances between these nine latent manifolds. 
The pairwise distances are generally close to $0.5$.
For reference, $\soThreeSet_1$ shape distances between similar objects (Fig.~\ref{fig:so2ShapeDistances}) are often around $0.5$ but can be as low as $0.16$, while distances between less-similar objects are often around $0.8$ and can be above $1.0$.
Fig.~\ref{fig:mdsConsistencyPermutations} also shows the first three PCA axes of all sets aligned to the default set.
Visually, the first three axes of these embeddings are all very similar. However, further investigation (not detailed here) shows that there are more significant differences on the further axes.
Importantly, we note that despite the differences between these latent spaces, the MDS stress values associated with each of them are approximately the same, meaning each is an equally good embedding of the image points according to this criterion.

This suggests the interpretation of the set of possible latent space manifolds for an object as an equivalence class. Then the shape distance between two objects on a pose set would be the infimum distance over all embeddings achieving near-minimum stress.
However, determining whether this infimum has been reached could be very difficult, both methodologically and in terms of computation time.
The process used in this paper manually pre-aligns the objects and fixes the seed used for initializations. This greatly reduces the impact of these variables, to an extent we believe is sufficient to give a reasonable characterization of the ideal results.
Other possible approaches might be to create a standardization rule for pre/post embedding permutation, or to compute embeddings for a large number of random permutations and use their mean to represent the latent space.

\section{Conclusion}
\label{sec:conclusion}

This paper provides a novel geometric perspective on image manifolds, offering both theoretical insights and practical implications. By applying \ac{MDS} to study pose and illumination manifolds, we have shown that these manifolds have inherent nonlinearity and distinct geometric properties that vary across classes of objects. Our systematic analysis demonstrates that global geometry-preserving dimensionality reduction techniques can effectively capture the intrinsic structure of these manifolds, offering clear advantages over unfolding-based methods.

The integration of Kendall's shape analysis has been particularly impactful, providing a robust framework for characterizing and comparing manifold geometries while maintaining invariance to latent space transformations. This approach shows that objects within the same class often have pose/illumination latent manifolds with similar shapes, suggesting an underlying relationship between object geometry and latent manifold structure.

These findings highlight opportunities to advance computer vision. By deforming latent manifolds and transferring knowledge between objects, we open the door to more efficient learning systems. At the same time, our shape-based clustering framework shows promise in automating object categorization using only manifold geometry. Additionally, our analysis of manifold nonlinearity provides valuable insights for developing geometry-driven deep generative models, paving the way for more advanced methods in image understanding and processing.




\section*{Acknowledgements}
This research was supported in part by the grants DARPA-PA-21-04-05—FP-052, NSF IIS 1955154, and NSF DMS – 2413748.
We also thank the producers of Clara.io for making their data public. 

\section*{Data Availability Statement}
The data used in this paper has been fully simulated using 3D models available at https://clara.io.

\backmatter
\printbibliography

@article{goodfellow2020generative,
  title={Generative adversarial networks},
  author={Goodfellow, Ian and Pouget-Abadie, Jean and Mirza, Mehdi and Xu, Bing and Warde-Farley, David and Ozair, Sherjil and Courville, Aaron and Bengio, Yoshua},
  journal={Communications of the ACM},
  volume={63},
  number={11},
  pages={139--144},
  year={2020},
  publisher={ACM New York, NY, USA}
}

@article{kingma2013auto,
  title={Auto-encoding variational bayes},
  author={Kingma, Diederik P. and Welling, Max},
  journal={arXiv preprint arXiv:1312.6114},
  year={2013}
}

@article{grenander-etal:2000,
    author = "Ulf Grenander and Anuj Srivastava and Michael Miller", 
    title = "Asymptotic Performance Analysis of Bayesian Object Recognition", 
    journal = "IEEE Transactions on Information Theory", 
    volume = "46", 
    number = "4", 
    pages = "1658-66", 
    year = {2000},
    }

@article{roweis-saul:2000,
author = {Sam T. Roweis  and Lawrence K. Saul },
title = {Nonlinear Dimensionality Reduction by Locally Linear Embedding},
journal = {Science},
volume = {290},
number = {5500},
pages = {2323-2326},
year = {2000},
}

@article{donoho-grimes:2003,
author = {David L. Donoho  and Carrie Grimes },
title = {Hessian eigenmaps: Locally linear embedding techniques for high-dimensional data},
journal = {Proceedings of the National Academy of Sciences},
volume = {100},
number = {10},
pages = {5591-5596},
year = {2003},
}

@inproceedings{shao-etal-arxivL2017,
  title={The riemannian geometry of deep generative models},
  author={Shao, Hang and Kumar, Abhishek and Thomas Fletcher, P},
  booktitle={Proceedings of the IEEE Conference on Computer Vision and Pattern Recognition Workshops},
  pages={315--323},
  year={2018}
}

@InProceedings{kuhnel:2021,
author="K{\"u}hnel, Line
and Fletcher, Tom
and Joshi, Sarang
and Sommer, Stefan",
title="Latent Space Geometric Statistics",
booktitle="Pattern Recognition. ICPR International Workshops and Challenges",
year={2021},
publisher="Springer International Publishing",
address="Cham",
pages="163--178",
}

@article{bengio-review:2013,
    author = "Yoshua Bengio and Aaron Courville  and Pascal Vincent",   title = "Representation Learning: A Review and New Perspectives", journal = "IEEE Trans. Pattern Anal. Mach. Intell.",
    volume = "35",
    volume = "8",
    pages = "1798-1828", 
    year = "August 2013",
    }

@inproceedings{shukla-etal:2018,
author = {Shukla, Ankita and Uppal, Shagun and Bhagat, Sarthak and Anand, Saket and Turaga, Pavan},
title = {Geometry of Deep Generative Models for Disentangled Representations},
year = {2020},
publisher = {Association for Computing Machinery},
address = {New York, NY, USA},
series = {ICVGIP 2018}
}

@article{yershova2010generating,
  title={Generating uniform incremental grids on SO (3) using the Hopf fibration},
  author={Yershova, Anna and Jain, Swati and Lavalle, Steven M. and Mitchell, Julie C.},
  journal={The International journal of robotics research},
  volume={29},
  number={7},
  pages={801--812},
  year={2010},
  publisher={SAGE Publications Sage UK: London, England}
}

@article{hardin2016comparison,
  title={A Comparison of Popular Point Configurations on $\mathbb{S}^2$},
  author={Hardin, Doug P. and Michaels, TJ and Saff, Edward B.},
  journal={arXiv preprint arXiv:1607.04590},
  year={2016}
}

@article{swinbank2006fibonacci,
  title={Fibonacci grids: A novel approach to global modelling},
  author={Swinbank, Richard and James Purser, R},
  journal={Quarterly Journal of the Royal Meteorological Society: A journal of the atmospheric sciences, applied meteorology and physical oceanography},
  volume={132},
  number={619},
  pages={1769--1793},
  year={2006},
  publisher={Wiley Online Library}
}

@misc{ClaraIO,
  title = {Clara.io},
  howpublished = {\url{https://clara.io}},
}

@software{meshio,
  author = {Schl\"omer, Nico},
  doi = {10.5281/zenodo.1173115},
  title = {{meshio: Tools for mesh files}},
  url = {https://github.com/nschloe/meshio},
}

@inproceedings{liang2024learning,
  title={Learning Geometry of Pose Image Manifolds in Latent Spaces Using Geometry-Preserving GANs},
  author={Liang, Shenyuan and Beaudett, Benjamin and Turaga, Pavan and Anand, Saket and Srivastava, Anuj},
  booktitle={International Conference on Pattern Recognition},
  pages={56--72},
  year={2024},
  organization={Springer}
}

@article{chen2020learning,
  title={Learning flat latent manifolds with vaes},
  author={Chen, Nutan and Klushyn, Alexej and Ferroni, Francesco and Bayer, Justin and Van Der Smagt, Patrick},
  journal={arXiv preprint arXiv:2002.04881},
  year={2020}
}

@inproceedings{lee2022regularized,
  title={Regularized autoencoders for isometric representation learning},
  author={Lee, Yonghyeon and Yoon, Sangwoong and Son, MinJun and Park, Frank C.},
  booktitle={International Conference on Learning Representations},
  year={2022}
}

@article{nazari2023geometric,
  title={Geometric Autoencoders--What You See is What You Decode},
  author={Nazari, Philipp and Damrich, Sebastian and Hamprecht, Fred A.},
  journal={arXiv preprint arXiv:2306.17638},
  year={2023}
}

@inproceedings{limgraph,
  title={Graph Geometry-Preserving Autoencoders},
  author={Lim, Jungbin and Kim, Jihwan and Lee, Yonghyeon and Jang, Cheongjae and Park, Frank C},
  booktitle={Forty-first International Conference on Machine Learning}
}

@inproceedings{singh2021structure,
  title={Structure-Preserving Deep Autoencoder-based Dimensionality Reduction for Data Visualization},
  author={Singh, Ayushman and Nag, Kaustuv},
  booktitle={2021 IEEE/ACIS 22nd International Conference on Software Engineering, Artificial Intelligence, Networking and Parallel/Distributed Computing (SNPD)},
  pages={43--48},
  year={2021},
  organization={IEEE}
}

@article{wold1987principal,
  title={Principal component analysis},
  author={Wold, Svante and Esbensen, Kim and Geladi, Paul},
  journal={Chemometrics and intelligent laboratory systems},
  volume={2},
  number={1-3},
  pages={37--52},
  year={1987},
  publisher={Elsevier}
}

@article{belkin2003laplacian,
  title={Laplacian eigenmaps for dimensionality reduction and data representation},
  author={Belkin, Mikhail and Niyogi, Partha},
  journal={Neural computation},
  volume={15},
  number={6},
  pages={1373--1396},
  year={2003},
  publisher={MIT Press}
}

@article{van2008visualizing,
  title={Visualizing data using t-SNE.},
  author={Van der Maaten, Laurens and Hinton, Geoffrey},
  journal={Journal of machine learning research},
  volume={9},
  number={11},
  year={2008}
}

@article{balasubramanian2002isomap,
  title={The isomap algorithm and topological stability},
  author={Balasubramanian, Mukund, and Schwartz, Eric},
  journal={Science},
  volume={295},
  number={5552},
  pages={7--7},
  year={2002},
  publisher={American Association for the Advancement of Science}
}

@book{pythonRef09, 
 author = {Van Rossum, Guido and Drake, Fred L.}, 
 title = {Python 3 Reference Manual}, 
 year = {2009}, 
 isbn = {1441412697}, 
 publisher = {CreateSpace}, 
 address = {Scotts Valley, CA} 
}

@article{scikitLearn,
 title={Scikit-learn: Machine Learning in {P}ython},
 author={Pedregosa, Fabian and Varoquaux, G. and Gramfort, A. and Michel, V. and Thirion, B. and Grisel, O. and Blondel, M. and Prettenhofer, P. and Weiss, R. and Dubourg, V. and Vanderplas, J. and Passos, A. and Cournapeau, D. and Brucher, M. and Perrot, M. and Duchesnay, E.},
 year={2011},
 journal={Journal of Machine Learning Research},
 volume={12},
 pages={2825--2830},
}

@ARTICLE{2020SciPy-NMeth,
  author  = {Virtanen, Pauli and Gommers, Ralf and Oliphant, Travis E. and
            Haberland, Matt and Reddy, Tyler and Cournapeau, David and
            Burovski, Evgeni and Peterson, Pearu and Weckesser, Warren and
            Bright, Jonathan and {van der Walt}, St{\'e}fan J. and
            Brett, Matthew and Wilson, Joshua and Millman, K. Jarrod and
            Mayorov, Nikolay and Nelson, Andrew R. J. and Jones, Eric and
            Kern, Robert and Larson, Eric and Carey, C J and
            Polat, {\.I}lhan and Feng, Yu and Moore, Eric W. and
            {VanderPlas}, Jake and Laxalde, Denis and Perktold, Josef and
            Cimrman, Robert and Henriksen, Ian and Quintero, E. A. and
            Harris, Charles R. and Archibald, Anne M. and
            Ribeiro, Ant{\^o}nio H. and Pedregosa, Fabian and
            {van Mulbregt}, Paul and {SciPy 1.0 Contributors}},
  title   = {{{SciPy} 1.0: Fundamental Algorithms for Scientific
            Computing in Python}},
  journal = {Nature Methods},
  year    = {2020},
  volume  = {17},
  pages   = {261--272},
  adsurl  = {https://rdcu.be/b08Wh},
  doi     = {10.1038/s41592-019-0686-2},
}

@online{plotly,
  author = {Plotly Technologies Inc.},
  title = {Collaborative data science},
  publisher = {Plotly Technologies Inc.},
  address = {Montreal, QC}, year = {2015},
  url = {https://plot.ly}
}

@article{torgerson1952multidimensional,
  title={Multidimensional scaling: I. Theory and method},
  author={Torgerson, Warren S.},
  journal={Psychometrika},
  volume={17},
  number={4},
  pages={401--419},
  year={1952},
  publisher={Springer}
}

@article{mead1992review,
  title={Review of the development of multidimensional scaling methods},
  author={Mead, Al},
  journal={Journal of the Royal Statistical Society: Series D (The Statistician)},
  volume={41},
  number={1},
  pages={27--39},
  year={1992},
  publisher={Wiley Online Library}
}

@article{pearson1901liii,
  title={LIII. On lines and planes of closest fit to systems of points in space},
  author={Pearson, Karl},
  journal={The London, Edinburgh, and Dublin philosophical magazine and journal of science},
  volume={2},
  number={11},
  pages={559--572},
  year={1901},
  publisher={Taylor \& Francis}
}

@article{lee2005acquiring,
  title={Acquiring linear subspaces for face recognition under variable lighting},
  author={Lee, Kuang-Chih and Ho, Jeffrey and Kriegman, David J.},
  journal={IEEE Transactions on pattern analysis and machine intelligence},
  volume={27},
  number={5},
  pages={684--698},
  year={2005},
  publisher={IEEE}
}

@article{peyre2009manifold,
  title={Manifold models for signals and images},
  author={Peyr{\'e}, Gabriel},
  journal={Computer vision and image understanding},
  volume={113},
  number={2},
  pages={249--260},
  year={2009},
  publisher={Elsevier}
}

@inproceedings{turaga2008statistical,
  title={Statistical analysis on Stiefel and Grassmann manifolds with applications in computer vision},
  author={Turaga, Pavan and Veeraraghavan, Ashok and Chellappa, Rama},
  booktitle={2008 IEEE conference on computer vision and pattern recognition},
  pages={1--8},
  year={2008},
  organization={IEEE}
}

@article{belhumeur1997eigenfaces,
  title={Eigenfaces vs. fisherfaces: Recognition using class specific linear projection},
  author={Belhumeur, Peter N. and Hespanha, Joao P. and Kriegman, David J.},
  journal={IEEE Transactions on pattern analysis and machine intelligence},
  volume={19},
  number={7},
  pages={711--720},
  year={1997},
  publisher={IEEE}
}

@article{tenenbaum2000global,
  title={A global geometric framework for nonlinear dimensionality reduction},
  author={Tenenbaum, Joshua B. and Silva, Vin de and Langford, John C.},
  journal={science},
  volume={290},
  number={5500},
  pages={2319--2323},
  year={2000},
  publisher={American Association for the Advancement of Science}
}

@article{bal2024statistical,
  title={Statistical Analysis of Complex Shape Graphs},
  author={Bal, Aditi Basu and Guo, Xiaoyang and Needham, Tom and Srivastava, Anuj},
  journal={IEEE Transactions on Pattern Analysis and Machine Intelligence},
  year={2024},
  publisher={IEEE}
}

@inproceedings{liang2024shape,
  title={Shape-graph matching network (SGM-net): Registration for statistical shape analysis},
  author={Liang, Shenyuan and Srivastava, Anuj and Segundo, Mauricio Pamplona and Sarkar, Sudeep and Aakur, Sathyanarayanan N.},
  booktitle={2024 IEEE International Symposium on Biomedical Imaging (ISBI)},
  pages={1--4},
  year={2024},
  organization={IEEE}
}

@book{mardia-dryden-book,
    author = "Dryden, Ian and Mardia, Kanti",
    title = "Statistical Shape Analysis",
    year = {1998},
    publisher = "John Wiley \& Son",
}

@book{small-shapes,
    author = "Small, Christopher G.",
    title = "The Statistical Theory of Shape",
    publisher = "Springer",
    year = {1996},
}

@book{kendall-barden-carne,
    title = "Shape and shape theory",
    author = "Kendall, David G. and Barden, Dennis and Carne, Thomas K. and Le, Hung",
    publisher = "Wiley",
    year = {1999},
}

@article{kendall1977diffusion,
  title={The diffusion of shape},
  author={Kendall, David G.},
  journal={Advances in applied probability},
  volume={9},
  number={3},
  pages={428--430},
  year={1977},
  publisher={Cambridge University Press}
}

@book{FDA,
  title={Functional and shape data analysis},
  author={Srivastava, Anuj, and Klassen, Eric},
  volume={1},
  year={2016},
  publisher={Springer}
}

@article{younes-distance,
    title = "Optimal matching between shapes via elastic deformations",
    author = "Younes, Laurent",
    journal = "Journal of Image and Vision Computing",
    volume = "17",
    number = "5/6",
    pages = "381-389",
    year = {1999},
}

@ARTICLE{younes-michor-mumford-shah:08,
  author = {Younes, Laurent and Michor, Peter W. and Shah, Jayant and Mumford, David},
  year = 2008,
  title = {A Metric on Shape Space with Explicit Geodesics},
  journal = {Matematica E Applicazioni},
  volume = 19,
  number = 1,
  pages = {25--57}
}

@article{su2020shape,
  title={Shape analysis of surfaces using general elastic metrics},
  author={Su, Zhe and Bauer, Martin and Preston, Stephen C. and Laga, Hamid and Klassen, Eric},
  journal={Journal of Mathematical Imaging and Vision},
  volume={62},
  number={8},
  pages={1087--1106},
  year={2020},
  publisher={Springer}
}

@article{jermyn2017elastic,
  title={Elastic shape analysis of three-dimensional objects},
  author={Jermyn, Ian H. and Kurtek, Sebastian and Laga, Hamid and Srivastava, Anuj},
  journal={Synthesis Lectures on Computer Vision},
  volume={12},
  number={1},
  pages={1--185},
  year={2017},
  publisher={Morgan \& Claypool Publishers}
}

@article{guo2022statistical,
  title={Statistical shape analysis of brain arterial networks (BAN)},
  author={Guo, Xiaoyang and Basu Bal, Aditi and Needham, Tom and Srivastava, Anuj},
  journal={The Annals of Applied Statistics},
  volume={16},
  number={2},
  pages={1130--1150},
  year={2022},
  publisher={Institute of Mathematical Statistics}
}

@article{beaudett2024reducing,
  title={Reducing Shape-Graph Complexity with Application to Classification of Retinal Blood Vessels and Neurons},
  author={Beaudett, Benjamin and Srivastava, Anuj},
  journal={arXiv preprint arXiv:2409.09168},
  year={2024}
}

@article{duque2022geometry,
  title={Geometry regularized autoencoders},
  author={Duque, Andres F. and Morin, Sacha and Wolf, Guy and Moon, Kevin R.},
  journal={IEEE transactions on pattern analysis and machine intelligence},
  volume={45},
  number={6},
  pages={7381--7394},
  year={2022},
  publisher={IEEE}
}

@inproceedings{pai2019dimal,
  title={Dimal: Deep isometric manifold learning using sparse geodesic sampling},
  author={Pai, Gautam and Talmon, Ronen and Bronstein, Alex and Kimmel, Ron},
  booktitle={2019 IEEE Winter Conference on Applications of Computer Vision (WACV)},
  pages={819--828},
  year={2019},
  organization={IEEE}
}

@article{gong6recovering,
  title={Recovering manifold representations via unsupervised meta-learning},
  author={Gong, Yunye and Yao, Jiachen and Lian, Ruyi and Lin, Xiao and Chen, Chao and Divakaran, Ajay and Yao, Yi},
  journal={Frontiers in Computer Science},
  volume={6},
  pages={1255517},
  year={2025},
  publisher={Frontiers Media SA}
}

@inproceedings{shan2005appearance,
  title={Appearance manifold of facial expression},
  author={Shan, Caifeng and Gong, Shaogang and McOwan, Peter W.},
  booktitle={International Workshop on Human-Computer Interaction},
  pages={221--230},
  year={2005},
  organization={Springer}
}

@inproceedings{einecke2007walking,
  title={Walking Appearance Manifolds without Falling Off},
  author={Einecke, Nils and Eggert, Julian and Hellbach, Sven and K{\"o}rner, Edgar},
  booktitle={International Conference on Neural Information Processing},
  pages={653--662},
  year={2007},
  organization={Springer}
}

@inproceedings{rahimi2005learning,
  title={Learning appearance manifolds from video},
  author={Rahimi, Ali and Darrell, Trevor and Recht, Ben},
  booktitle={2005 IEEE Computer Society Conference on Computer Vision and Pattern Recognition (CVPR'05)},
  volume={1},
  pages={868--875},
  year={2005},
  organization={IEEE}
}

@inproceedings{khrulkov2018geometry,
  title={Geometry score: A method for comparing generative adversarial networks},
  author={Khrulkov, Valentin and Oseledets, Ivan},
  booktitle={International conference on machine learning},
  pages={2621--2629},
  year={2018},
  organization={PMLR}
}


\end{document}